\documentclass[10pt]{article} 
\usepackage[accepted]{tmlr}

\usepackage{amsmath,amsfonts,bm}

\def\eqref#1{equation~\ref{#1}}

\def\1{\bm{1}}

\DeclareMathAlphabet{\mathsfit}{\encodingdefault}{\sfdefault}{m}{sl}
\SetMathAlphabet{\mathsfit}{bold}{\encodingdefault}{\sfdefault}{bx}{n}

\usepackage{microtype}
\usepackage{graphicx}
\usepackage{caption}
\usepackage{subcaption}
\usepackage{enumitem}
\usepackage{bbm}
\usepackage{float}
\usepackage{booktabs} 

\usepackage{hyperref}
\usepackage{url}
\usepackage{amsmath}
\usepackage{amssymb}
\usepackage{mathtools}
\usepackage{amsthm}

\usepackage[capitalize,noabbrev]{cleveref}

\theoremstyle{plain}

\theoremstyle{definition}

\theoremstyle{remark}

\usepackage[normalem]{ulem}
\newcommand\pcref[1]{(\cref{#1})}

\newcommand{\kkt}[1]{}

\title{Pretrained deep models outperform GBDTs \\ in Learning-To-Rank under label scarcity}

\author{\name Charlie Hou\thanks{Work partially done while Charlie Hou was an intern at Amazon.} \email charlieh@andrew.cmu.edu \\
      \addr Department of Electrical and Computer Engineering \\
      Carnegie Mellon University
      \AND
      \name Kiran K. Thekumparampil \email kkt@amazon.com \\
      \addr Amazon, Palo Alto
      \AND
      \name Michael Shavlovsky \email shavlov@amazon.com \\
      \addr Amazon, Palo Alto
      \AND 
      \name Giulia Fanti \email gfanti@andrew.cmu.edu \\
      \addr Department of Electrical and Computer Engineering \\
      Carnegie Mellon University
      \AND
      Yesh Dattatreya \email ydatta@amazon.com \\
      \addr Amazon, Palo Alto
      \AND
      Sujay Sanghavi \email sanghavi@mail.utexas.edu \\
      \addr Department of Computer Science \\
      University of Texas at Austin}

\def\month{11}  
\def\year{2024} 
\def\openreview{\url{https://openreview.net/forum?id=093Q9VxaWt}} 

\begin{document}

\maketitle

\begin{abstract}

On tabular data, a significant body of literature has shown that current deep learning (DL) models perform at best similarly to Gradient Boosted Decision Trees (GBDTs), while significantly underperforming them on outlier data \cite{gorishniy2021revisiting,rubachev2022revisiting,mcelfresh2023neural}. However, these works often study problem settings which may not fully capture the complexities of real-world scenarios. We identify a natural tabular data setting where DL models can outperform GBDTs: tabular Learning-to-Rank (LTR) under label scarcity. Tabular LTR applications, including search and recommendation, often have an abundance of unlabeled data, and \emph{scarce} labeled data. We show that DL rankers can utilize unsupervised pretraining to exploit this unlabeled data. In extensive experiments over both public and proprietary datasets, we show that pretrained DL rankers consistently outperform GBDT rankers on ranking metrics---sometimes by as much as $38\%$---both overall and on outliers. 

\end{abstract}

\section{Introduction}
The learning-to-rank (LTR) problem aims to train a model to rank a set of items according to their relevance or user preference \citep{liu2009learning}.  
An LTR model is typically trained on a dataset of queries and associated {\em query groups} (i.e., a set of potentially relevant \emph{documents} or \emph{items} per query), as well as an associated (generally incomplete) ground truth ranking of the items in the query group. 
The model is trained to output the optimal ranking of documents or items in a query group, given a query.
LTR is a core ML technology in many real world applications---most notably in search contexts including Bing web search \citep{mslr}, Amazon product search \citep{yang2022toward}, and Netflix movie recommendations \citep{lamkhede2021recommendations}. More recently, LTR has also found use in Retrieval Augmented Generation (RAG), where a retrieval system is paired with an LLM (large language model) to ground the latter's responses to truth and reduce hallucination \cite{lewis2020retrieval,glass2022re2g,pan2022end}.

Traditionally, LTR models took {\em tabular features}---numerical or categorical features---of queries and documents as input features \citep{burges2005learning, cao2007learning, chapelle2011yahoo, mslr, lucchese2016post}. 
As this continues to be the case for many important applications, \textit{tabular LTR} is a core problem in machine learning \citep{bower2021individually, yang2022toward, yu2020ptranking, kveton2022value, pan2022end, xia2008listwise, chen2009ranking}.
Today, deep models are largely outperformed by gradient boosted decision trees (GBDTs) \citep{friedman2001greedy} in ML problems with tabular features \citep{jeffares2023tangos, qin2021neural}. In contrast,  deep models are state-of-the-art by a significant margin in domains like text \citep{devlin2018bert} and images \citep{he2016deep}.

Recent breakthroughs in modeling non-tabular data like text and images
have been driven by first training a deep neural network to learn from unlabeled data (unsupervised pretraining, or pretraining) \citep{devlin2018bert, chen2020simple}, followed by supervised training (finetuning). 
Models that are pretrained in this way can perform significantly better than models that were only trained on existing labeled data.
This remarkable success 
stems from three factors: (1) limited access to labeled data, (2) large, available sources of unlabeled text and image data, and (3) pretraining methods that can exploit unlabeled data.

\begin{figure*}[t]
    \centering
    \begin{subfigure}[b]{0.32\textwidth}
        \centering
        \includegraphics[width=\textwidth]{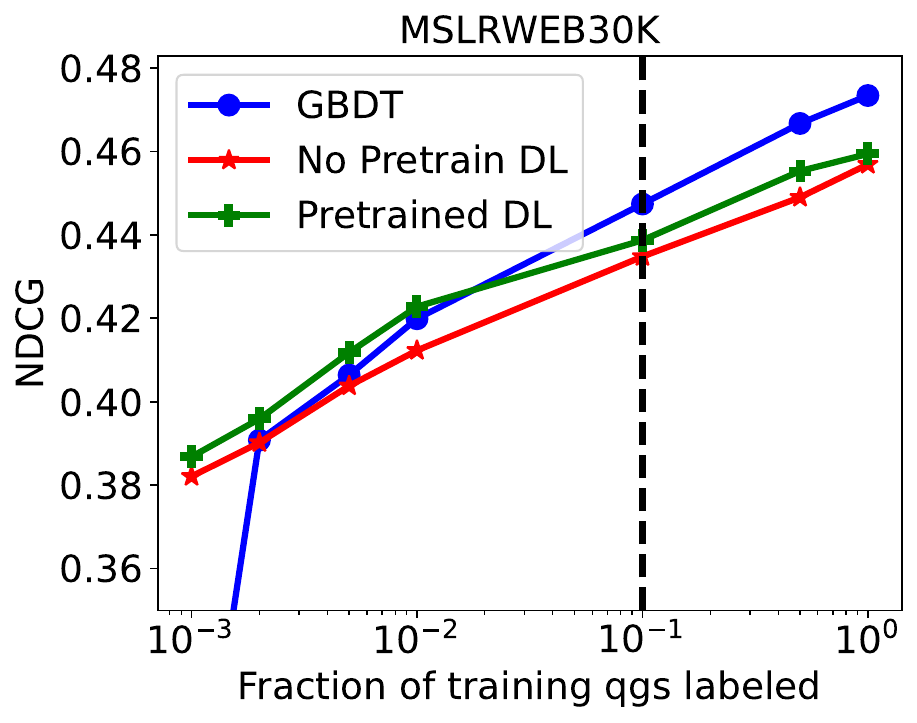}
        \label{fig:fraction mslr}
    \end{subfigure}
    \begin{subfigure}[b]{0.32\textwidth}
        \centering
        \includegraphics[width=\textwidth]{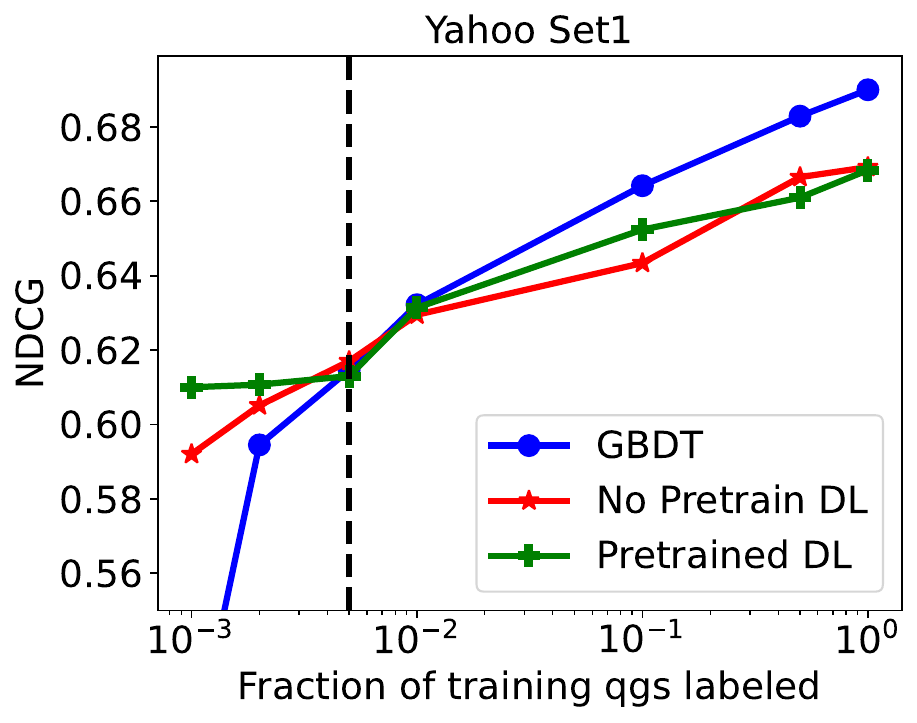}
        \label{fig:fraction yahoo}
    \end{subfigure}
    \begin{subfigure}[b]{0.32\textwidth}
      \centering
      \includegraphics[width=\textwidth]{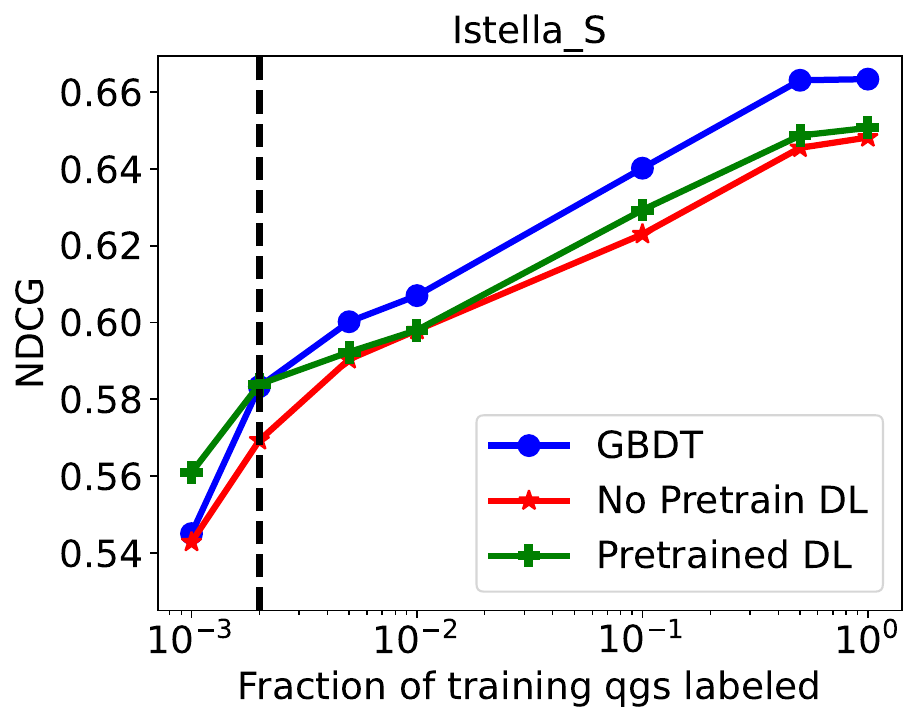}
      \label{fig:fraction istella}
  \end{subfigure}
       \caption{\textbf{Main result (Simulated crowdsourcing of labels):} We compare NDCG ($\uparrow$) 
       as we sweep the percentage of labeled training query groups (QGs); labeled QGs have relevance scores for every element of the QG.
       For small enough fractions of labeled query groups, pre-trained DL rankers (sometimes significantly) outperform both GBDTs (representing the best of supervised GBDTs and semi-supervised GBDTs) and non-pretrained DL rankers (representing the best of tabular ResNet \citep{gorishniy2021revisiting}, DeepFM \citep{guo2017deepfm}, and DCNv2 \citep{wang2021dcn}). To the left of the black dotted line, pretrained rankers perform the best (statistically significant at $p=0.05$ level).  Points are averages over 3 trials.}
       \label{fig:fractions}
       \vspace{-0.15in}
  \end{figure*}

A natural question is whether deep models can similarly use unsupervised pretraining to outperform GBDTs on the tabular LTR problem. 
To this end, many papers have studied  unsupervised pretraining techniques---both using deep learning and GBDTs---for tabular data \cite{yoon2020vime, ucar2021subtab, bahri2021scarf, verma2021towards, majmundar2022met, duh2008learning, pseudo2013simple}. 
However, none of these methods have shown convincingly that deep learning (with or without unsupervised pretraining) consistently outperforms GBDTs (semi-supervised or supervised); at best, deep methods appear to achieve roughly the same performance as GBDTs \citep{jeffares2023tangos, rubachev2022revisiting, gorishniy2021revisiting, shwartz2022tabular, grinsztajn2022tree}. 
Furthermore, recent work showed that GBDTs are significantly more robust against outliers in the dataset than deep learning models \cite{mcelfresh2023neural}. We note that most of these works study idealized problem settings which may fail to capture complexities of real-world scenarios such as label scarcity and noisy data.
Prior work in other domains show that pretrained deep models can outperform purely supervised models \cite{chen2020simple} and former models are more robust to noise \cite{hendrycks2019using}. Therefore, we hypothesize that pretrained deep models can outperform GBDTs in tabular settings that more accurately reflect real-world challenges.

Through extensive empirical evaluation, we confirm this hypothesis and show that deep models which use unsupervised pretraining consistently outperform GBDTs---sometimes by a large margin---\textbf{when labels are scarce}. By scarce, we mean that only a small number of query groups contain labels. 
This setting is more standard in most real-world deployments of LTR, rather than the idealized settings studied in prior work. For instance, scarcity can arise when organizations: (a) crowdsource labels for a limited number of  query groups due to cost constraints, and/or (b) use rare user feedback (e.g., clicks, purchases) as implicit labels. Our experiments consider both sources of scarcity. They are run on the three standard public datasets in the ranking literature: MSLRWEB30K \citep{mslr}, Yahoo \citep{chapelle2011yahoo}, Istella\_S \citep{lucchese2016post}, as well as an  industry-scale proprietary ranking dataset.


\textbf{Contributions:} \textit{(1) We demonstrate for the first time that unsupervised pretraining can produce deep models that outperform GBDTs in ranking.} Specifically, pretrained deep rankers can achieve  up to 38\% higher NDCG (normalized discounted cumulative gain) score \citep{burges2010ranknet} than GBDTs when labels are scarce \pcref{fig:fractions}. For instance, \cref{fig:fractions} shows on three datasets that as the labels get scarcer, GBDTs start to substantially underperform pretrained deep rankers (experimental setup in \cref{sec: sparse}).

\textit{(2) We prescribe empirically-justified LTR-specific pretraining strategies}, including a new ranking-specific pretraining loss, SimCLR-Rank, 
which is a modification of the widely used SimCLR \citep{chen2020simple} loss. In experiments, SimCLR-Rank achieves similar or better performance than most existing pretraining methods at an order-of-magnitude speedup. This allows SimCLR-Rank to scale to large real-world datasets such as our proprietary industry-scale ranking dataset.

\noindent \textit{(3) We demonstrate that when labels are scarce, pretrained deep rankers can perform significantly better than GBDTs on \textbf{outlier} query groups, boosting NDCG  by as much as 80\% (\cref{fig:outlier fractions}}, \cref{app:public-relevance}). This is perhaps surprising, given prior results in the label-rich regime showing that GBDTs are significantly more robust to data with irregularities than deep rankers \citep{mcelfresh2023neural}.

The main novelty of this work lies in the identification of label-scarce LTR as a practical real-world tabular learning problem and the results which challenge the prior understanding of the optimal modeling choices for tabular learning. We demonstrate that pretraining strongly help deep models to consistently outperform GBDTs in this label-scarce problem. Of note, these improvements over GBDTs happen even when the labeled data is plentiful in absolute numbers; up to 25\% labeled query groups in public datasets and 3 million+ labeled query groups in the large industry-scale ranking dataset. 
Prior to our work, the ML and ranking literature had not yet identified a widely applicable setting where (pretrained) deep models could consistently outperform GBDTs \citep{qin2021neural,mcelfresh2023neural}.
Due to the prevalence of label scarcity in practical settings, our results suggest a positive change in how practitioners may approach real-world tabular learning problems, especially in search and recommendation systems.

\noindent

\textbf{Related work:} We focus on the traditional LTR setting where the features are all numeric (tabular data). 
In this setting, gradient boosted decision trees (GBDTs) \citep{friedman2001greedy} have historically been  the de-facto models, though there is currently great interest in applying deep neural networks to tabular data. 
\citet{borisov2022deep} categorize existing techniques for using deep neural networks over tabular data into four types: (1) \textbf{Encoding-based methods} such as VIME \citep{yoon2020vime}, SCARF \citep{bahri2021scarf}, IGTD \citep{zhu2021converting}, and SuperTML \citep{sun2019supertml}; (2) \textbf{Novel hybrid architectures} such as DeepFM \citep{guo2017deepfm}, xDeepFM \citep{lian2018xdeepfm}, and many others \cite{cheng2016wide,frosst2017distilling,ke2018tabnn,ke2019deepgbm,popov2019neural,luo2020network,liu2020dnn2lr,ivanov2021boost,luo2021sdtr};  (3) \textbf{Transformer-based architectures} including SAINT \citep{somepalli2021saint}, TabNet \citep{tabnet}, TabTransformer \citep{huang2020tabtransformer}, and ARM-Net \cite{cai2021arm}); (4) \textbf{Regularized DNNs} \citep{shavitt2018regularization, kadra2021well}.

Within these categories, proposed architectures
include factorization machines \citep{rendle2010factorization}, wide \& deep architectures \citep{cheng2016wide}, and similar architectures designed for recommendation systems \citep{guo2017deepfm, wang2021dcn, lian2018xdeepfm, naumov2019deep, qu2016product}. While our paper is architecture agnostic, we perform comparisons of pretrained models against non-pretrained DeepFM \cite{guo2017deepfm} and DCNv2 \cite{wang2021dcn} models to show that pretraining is needed to outperform GBDTs consistently.
Self-supervised learning (SSL) or unsupervised pretraining has improved overall performance and robustness to noise
\citep{hendrycks2019using} in settings where there is a significant source of unlabeled data like text \citep{devlin2018bert, howard2018universal, kitaev2018multilingual, liu2017adversarial, song2015unsupervised} and images \citep{chen2020simple, chen2021exploring, grill2020bootstrap, ge2021robust, xie2020self}. 

Inspired by the success of SSL in images and text, many unsupervised pretraining tasks have been proposed for tabular data. We describe some of the most well-known methods. SubTab \citep{ucar2021subtab} proposes to train an autoencoder to reconstruct a table using a subset of the columns. VIME-self \citep{yoon2020vime} is a similar method. SCARF and DACL+ \citep{bahri2021scarf, verma2021towards} propose augmentations for the SimCLR pretraining task \citep{chen2020simple} that work for the tabular setting. RegCLR \citep{wang2022regclr} is a method for extracting tables from images, which is not directly applicable to our setting. 

A smaller class of approaches has studied regularized DNNs, finding that they require limited changes to perform well on tabular data \cite{shavitt2018regularization,kadra2021well}. However, the most competitive of these methods require extensive hyperparameter optimization over a 14-dimensional grid \cite{kadra2021well}.

While the body of work studying neural network architectures for tabular data is extensive, deep models have yet to outperform GBDTs convincingly \citep{qin2021neural, joachims2006training, ai2019learning,bruch2019revisiting,ai2018learning, pang2020setrank, mcelfresh2023neural}.
In this work, we consider encoding-based methods, which alter the representations over which a deep model can learn to rank; we do not focus on introducing new algorithms in this work, but rather evaluate mostly existing methods applied to tabular data. 
We comprehensively evaluate seven representative DL methods: SubTab, SCARF, DACL+, VIME-self, SimCLR-Rank, SAINT, SimSiam \citep{chen2021exploring}, and self-supervised approaches for GBDTs in \cref{sec:all pretrain comparisons}. Our results show that multiple pretrained deep rankers can consistently outperform GBDTs in the label-scarce tabular LTR setting, with SimCLR-Rank and SimSiam \citep{chen2021exploring}  generally performing the best.
We provide a detailed supplemental discussion of related works in the Appendix \ref{sec:relatedwork}.


\section{Learning-To-Rank problem and its metrics}
In the LTR problem, samples are query groups (QGs) consisting of a query (e.g.~shopping query) and $L$ potentially relevant items (e.g. products) for this query. Although an LTR dataset contains multiple QGs, for the sake of simplicity, the notation in this section only handles a single QG. The $i$-th relevant item in this QG is represented by a feature vector $\boldsymbol{x}_{i} \in \mathbb{R}^d$, which captures information about this query-item pair, subsuming any query related features. 
If and only if the QG is labeled, such as the ones used in supervised training or testing, we also have an associated scalar relevance label $y_{i}$, which could be a binary, ordinal, or real-valued measurement of the relevance of this item for this query \citep{qin2021neural}. The objective is to learn a function that ranks these $L$ items such that the items with highest true relevance are ranked at the top.

Most LTR algorithms formulate the problem as learning a scoring function $f_{\theta}: \mathbb{R}^d \to \mathbb{R}$ that maps the feature vector associated with each item to a score, and then ranking the items by sorting the scores in descending order.
To measure the quality of a ranking induced by our scoring function $f_\theta$ on a labeled test QG, a commonly-used (per-QG) metric is NDCG (normalized cumulative discounted gain) ($\uparrow$):
\begin{align}
\text{NDCG}(\pi_{f_\theta}, \{y_{i}\}_{i=1}^{L}) = \frac{\text{DCG}(\pi_{f_\theta}, \{y_{i}\}_{i=1}^{L}) }{ \text{DCG}(\pi^*, \{y_{i}\}_{i=1}^{L})} \in [0,1]\,,
\end{align}
where $\pi_{f_\theta}:[L]\to [L]$ ($[L] \triangleq \{1,\dots,L\}$) is a ranking of the $L$ elements induced by the scoring function $f_\theta$ on $\{\boldsymbol{x}_{i}\}_{i=1}^{L}$ (i.e.~$\pi(i)$ is the rank of the $i$-th item), while $\pi^*$ is the ideal ranking induced by the true relevance labels 
$\{y_{i}\}_{i=1}^{L}$, and discounted cumulative gain ($\text{DCG}$) is defined as 
$
\text{DCG}(\pi, \{y_{i}\}_{i=1}^{L}) = \sum_{i=1}^L \frac{2^{y_{i}} - 1}{\text{log}_2(1 + \pi(i))}.
$
Here, ($\uparrow$) indicates that larger metric values are preferred. Note that for summarizing this metric over the full test set, we simply take the average across all the QGs in the set.
Typically, a truncated version of NDCG is used that only considers the top-$k$ ranked items, denoted as NDCG@$k$. In the rest of our paper, we will refer to NDCG@5 as NDCG; this will be our evaluation metric.

We consider a natural label-scarce LTR setting with large amounts of unlabeled data and a limited supply of labels for a subset of this data.
The labels that do exist can arise from multiple sources, such as crowdsourced workers paid to label QGs, or through implicit user feedback (e.g., clicks or purchases); we model both in our  evaluation.
Unless specified otherwise, we use all the query groups for unsupervised pretraining, but we use only the labeled query groups for supervised training.

\subsection{Outlier-NDCG for outlier performance evaluation}
\label{sec: outlierndcg}

Motivated by a recent observation that GBDTs outperform DL models in datasets with irregularities \cite{mcelfresh2023neural}, we also study their performance on outliers.
In interactive ML systems like search, performing well on outlier queries is particularly valuable as it
empowers users to search for more outlier queries, which in turn allows the modeler to collect more data and improve the model.
To this end, we evaluate the outlier performance by computing the average NDCG on outlier queries groups and refer to it as \emph{Outlier-NDCG} for brevity.
In practice, some outlier queries may already be known, and the modeler can define the outlier datasets accordingly, like in our proprietary dataset. For example, since industry data pipelines often have missing data/features one could identify samples with missing features as outliers.
When outliers are unknown, 
detecting them (especially in higher-dimensional data) is challenging. 
Hence, we create a heuristic algorithm by assuming outliers are rare values that are separated from the bulk of the data. Using it we systematically selected outlier query groups for the public datasets that we use. More details about the outliers are provided in \cref{app:outlierselection}.
Separately measuring the performance on the outliers is justified by our later observation that performance on the full dataset and the outliers can significantly differ (\cref{tab:tau 4.5,tab:amazon}).

\section{Design: Unsupervised pretraining for LTR}
\label{sec: pretraining strategies}


Qualitatively, the ranking relevance of tabular records to a query is typically determined by a combination of fields. For example, if a person searches for a product (say a pair of sandals), we can think of the fields in the LTR problem as representing properties of the results, such as material, color, and price. Qualitatively, we expect that if a pair of records are the same except for small differences between fields (e.g., missing a field), then the two products should generally have similar rankings in the LTR problem. 
In this section, we discuss how existing methods fail to capture this qualitative intuition. 


\textbf{Semi-supervised learning for GBDTs:} While to the best of our knowledge there are no methods to perform unsupervised pretraining in GBDTs, one can use semi-supervised methods like consistency regularization \citep{jeong2020consistency, sajjadi2016regularization, miyato2018virtual, oliver2018realistic, berthelot2019mixmatch}, pseudo-labeling/self-training \citep{rizve2021defense, pseudo2013simple, shi2018transductive, yarowsky1995unsupervised, mcclosky2006effective}, and PCA \citep{duh2008learning} to address limited size of labeled data. In consistency regularization, one increases the size of the training set by using data augmentations. In pseudo-labeling/self-training, a model trained on the available labeled data is used to label unlabeled data. Then, the final model is trained on the resulting ``labeled'' dataset. In PCA, one projects the features along directions of maximum variability in the union of labeled and unlabeled datasets. 
Since prior work has found that consistency regularization decreases the performance of GBDTs \citep{rubachev2022revisiting}, we evaluate only pseudo-labeling and PCA. 
None of these methods explicitly ensures that semantically similar records have similar representations in the learned models.

\textbf{Pretraining methods for tabular data:} To address this gap, there have been many unsupervised pretraining methods proposed for tabular data, which can be applied to the tabular LTR setting. In this paper, we will evaluate (1) SCARF \citep{bahri2021scarf} and (2) DACL+ \citep{verma2021towards} which are contrastive learning methods \citep{chen2020simple}, (3) VIME-self \citep{yoon2020vime} and (4) SubTab \citep{ucar2021subtab} which are autoencoder based methods. The four most commonly-evaluated baselines in the pretraining literature for tabular data are \citet{ucar2021subtab, hajiramezanali2022stab, levin2022transfer, majmundar2022met}.
Our results suggest that many of these methods outperform GBDTs in the label-scarce regime (\Cref{sec:all pretrain comparisons}), in accordance with the main theme of this paper. 
However, the consistently best-performing methods in our experiments were more basic, domain-agnostic pretraining methods, which can be viewed as more primitive building blocks of some of the above methods. 
Surprisingly, they use \emph{less} domain knowledge of tabular data than these more sophisticated methods.

Specifically, in the image setting, SimCLR \citep{chen2020simple} and SimSiam \citep{chen2021exploring} are two widely-adopted unsupervised pretraining methods. They use the idea that two variations of the same data point should have similar representations. This  idea generalizes across domains, including tabular LTR.

\textit{SimCLR \citep{chen2020simple}:} In a batch of $B$ query-groups, the feature vector $\boldsymbol{x}_{q,i}$ of $q$-th query group and its $i$-th item is stochastically augmented twice to obtain $\boldsymbol{x}_{q,i}^{(0)}$ and $\boldsymbol{x}_{q,i}^{(1)}$, which are called a \emph{positive pair}. In this paper, we use combinations of the following two augmentation strategies for all domain-agnostic pretraining methods: 
\textit{(a) Zeroing:} Zero out features selected randomly and i.i.d. with fixed probability.
\textit{(b) Gaussian noise:} Add Gaussian noise of a fixed scale to all features.
Further details are given in \cref{app:exp-details}.


Next, a base encoder $h(\cdot)$ and projection head $g(\cdot)$ map $\boldsymbol{x}_{q,i}^{(a)}$ to $\boldsymbol{z}_{q,i}^{(a)} = g(h(\boldsymbol{x}_{q,i}^{(a)}))$ for $a=0,1$. Then by minimizing the InfoNCE loss \citep{oord2018representation} in this projected representation space, we push the positive pair, $(\boldsymbol{z}_{q,i}^{(0)}, \boldsymbol{z}_{q,i}^{(1)})$, closer while pushing them farther from all other augmented data points in the batch, which are called their {\em negatives}. Formally, for $\boldsymbol{x}_{q,i}$, we minimize $\ell_{q,i}^{(0)} + \ell_{q,i}^{(1)}$, where:
\begin{align}
\label{eq:simclrloss}
    &\ell_{q,i}^{(a)} = - \text{cos}(\boldsymbol{z}_{q,i}^{(a)}, \boldsymbol{z}_{q,i}^{(\bar{a})})/\tau + \text{log} \sum_{q'=1}^B \sum_{i'=1}^{L_{q'}} \sum_{a'=0}^1 \big[\mathbf{1}\{(q',i',a')\neq(q,i,a)\} \cdot \exp(\text{cos}(\boldsymbol{z}_{q,i}^{(a)}, \boldsymbol{z}_{q',i'}^{(a')})/\tau)\big] 
\end{align}
$\bar{a}=1-a$, $\tau$ is the temperature parameter, $L_{q'}$ is the number of items in $q'$-th query group, and $\text{cos}(\boldsymbol{z},\widetilde{\boldsymbol{z}}) = \langle \boldsymbol{z}, \widetilde{\boldsymbol{z}} \rangle/\|\boldsymbol{z}\|\|\widetilde{\boldsymbol{z}}\|$ denotes the cosine similarity. After pretraining, the encoder $h$ is used in downstream applications. 
SimCLR achieves superior performance in many domains \citep{chen2021exploring,wang2022importance,wang2022regclr}. It is known that contrasting a pair of data samples which are hard to distinguish from each other, called {\em hard negatives} can help an encoder learn good representations \citep{robinson2020contrastive, oh2016deep, schroff2015facenet, harwood2017smart, wu2017sampling, ge2018deep, suh2019stochastic}. SimCLR simply contrasts all pairs of data samples in the batch (including from other query groups) with the assumption that a large enough batch is likely to contain a hard negative. 



\textit{SimSiam \citep{chen2021exploring}:} Similar to SimCLR, for each data point $\boldsymbol{x}_{q,i}$, this method produces stochastically-augmented \emph{positive pairs} and their \emph{projected} representations. However, we pass these representations further through a predictor $\text{pred}(\cdot)$, to get $\boldsymbol{p}_{q,i}^{(a)} = \text{pred}(\boldsymbol{z}_{q,i}^{(a)})$ for $a=0,1$. Finally, we maximize the similarity between the projected and predicted representations: $\text{cos}(\boldsymbol{p}_{q,i}^{(0)}, \text{sg}(\boldsymbol{z}_{q,i}^{(1)})) + \text{cos}(\boldsymbol{p}_{q,i}^{(1)}, \text{sg}(\boldsymbol{z}_{q,i}^{(0)}))$, where {\em sg} is a stop-gradient. Unlike SimCLR, there are no \emph{negatives}, i.e., the loss function does not push the representation of an augmented sample away from that of other samples' augmentations. The asymmetry in the loss due to the stop gradient and the predictor prevents representation collapse to make negatives unnecessary \cite{tian2021understanding, zhang2021does, wang2022importance}. If the number of items in each query group is $L$, then the time/space complexity for SimSiam is only $O(BL)$ per batch, whereas it is $O(B^2L^2)$ for SimCLR. Therefore, SimSiam is more efficient and it can scale to larger batchsizes and data. 


\begin{figure}[t]
    \centering
    \begin{subfigure}[b]{0.8\textwidth}
        \centering
        \includegraphics[width=\textwidth]{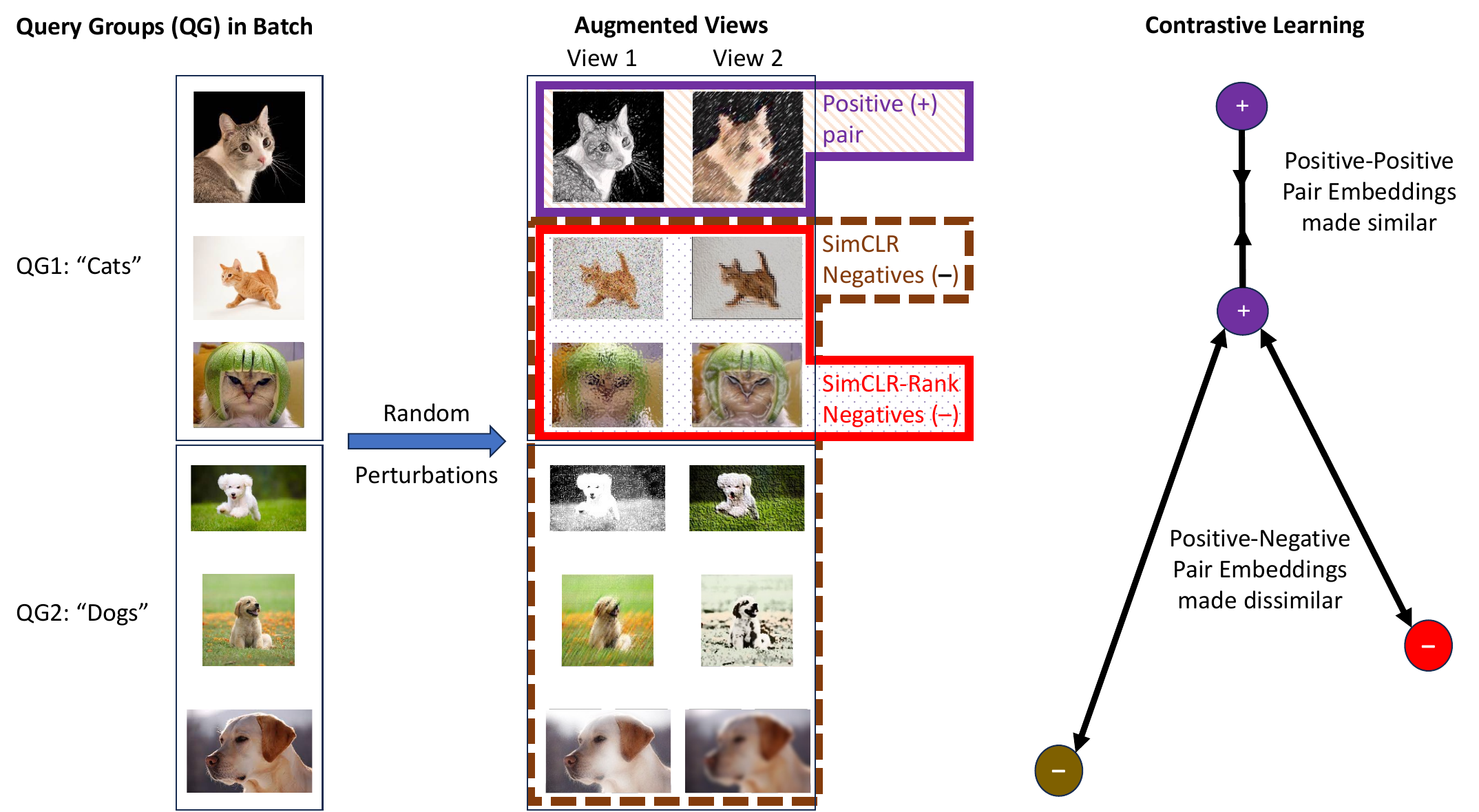}
    \end{subfigure}
       \caption{In SimCLR and its variants, \emph{positive pairs}, or augmented versions of the same data sample, are trained to have similar embeddings. Positive-negative pairs, or augmented samples originating from two different data points or classes, are trained to have distant embeddings. In vanilla SimCLR, each positive pair is contrasted with \emph{all other items in the batch}, denoted by the data points contained in the brown dashed line above. In SimCLR-Rank, each positive pair is contrasted with \emph{only the items in the same Query Group (QG)}, denoted by the data points inside the red solid line above. 
       }
       \label{fig:simclr vs simclrrank}
       \vspace{-0.2in}
\end{figure}

\textbf{Pretraining method for LTR:} Motivated by (a) SimCLR's high complexity and (b) the efficacy of contrasting with hard negatives, we additionally propose SimCLR-Rank, an LTR-specific alternative. 
Recall that the SimCLR loss for an item in \eqref{eq:simclrloss} uses all the other items in the batch as negatives.
Good pretraining strategies usually exploit the structure of the data, e.g.~text sequences motivates masked token prediction \cite{devlin2018bert}. Studying the query group structure of LTR data, we notice that high-quality hard negatives are freely available as \textit{the other items from the same query group}. These items are retrieved and deemed potentially similarly relevant for this query by an upstream retrieval model, making these items harder negatives than other items in the batch.
So, we propose SimCLR-Rank, which modifies SimCLR to contrast only with the other items in the same query group, as illustrated in \cref{fig:simclr vs simclrrank}. Formally, we modify the per-augmentation loss $\ell^{(a)}_{q,i}$ as:
\begin{align}
    \ell_{q,i}^{(a)} &= - \text{cos}(\boldsymbol{z}_{q,i}^{(a)}, \boldsymbol{z}_{q,i}^{(\bar{a})})/\tau + \text{log} \sum_{i'=1}^{L_{q}} \sum_{a'=0}^1 \big[ \mathbf{1}\{(i',a')\neq(i,a)\} \cdot \exp(\text{cos}(\boldsymbol{z}_{q,i}^{(a)}, \boldsymbol{z}_{q,i'}^{(a')})/\tau)\big] 
\end{align}
Empirically, SimCLR-Rank better separates embeddings within query groups relative to sampling negatives uniformly from each batch (\Cref{app:embedding-quality}).
The SimCLR-Rank loss has a time/space complexity of $O(B L^2)$, making it faster than SimCLR's $O(B^2 L^2)$ complexity, and enjoying SimSiam's linear complexity  in $B$. 
Next, we empirically study these methods.


\subsection{Evaluation of unsupervised pretraining techniques}
\label{sec:all pretrain comparisons}
In this subsection, we compare the discussed pretraining strategies in the following tabular LTR setting.

\begin{table*}[t]
    \centering
    \caption{{\bf Pretraining methods:} A comparison across unsupervised pretraining methods on NDCG, averaged over 3 trials. Here only 0.1\% of the query groups have labels for finetuning. \textbf{Bold} numbers denote the best in a column, and \underline{underlined} numbers are within the margin of error of the best. Given that SCARF/DACL+/SimCLR are too slow to scale to large LTR datasets (for example on the industrial-scale proprietary online shopping dataset), we find that SimCLR-Rank and SimSiam are the best pretraining methods for tabular LTR.}
    \begin{tabular}{l | c | c | c }
      \toprule
      {\bf Method} & {\bf MSLR} ($\uparrow$)  & {\bf Yahoo Set1} ($\uparrow$)  & {\bf Istella} ($\uparrow$)  \\
      \midrule
      Supervised GBDT & 0.2801 $\pm$ 0.0002 & 	0.5083 $\pm$ 0.0197 & 0.5450 $\pm$ 0.0000 \\
      Semi-supervised GBDT & 0.2839 $\pm$ 0.0004 & 0.5061 $\pm$ 0.0267 & 0.4656 $\pm$ 0.0310 \\
      SimCLR-Rank + GBDT & 0.3165 $\pm$ 0.0050 & 0.5504 $\pm$ 0.0135 & 0.5397 $\pm$ 0.0098 \\
      SimSiam + GBDT & 0.3158 $\pm$ 0.0030 & 0.5620 $\pm$ 0.0167 & 0.5297 $\pm$ 0.0051 \\
      \midrule
      SCARF { \citep{bahri2021scarf}} & 0.3807	$\pm$ 0.0016 & 0.5884	$\pm$ 0.0129 & 0.5542 $\pm$	0.0024 \\
      DACL+ { \citep{verma2021towards}} & \underline{0.3833 $\pm$	0.0037} & 0.5887 $\pm$	0.0012 & \textbf{0.5626 $\pm$ 0.0024} \\
      VIME-self {\citep{yoon2020vime}} & 0.3834	$\pm$ 0.0011 & 0.5839	$\pm$ 0.0058 & 0.5514 $\pm$ 0.0025\\
      SubTab { \citep{ucar2021subtab}} & 0.3748 $\pm$ 0.0025 & 0.5814 $\pm$ 0.0062 & 0.5082 $\pm$ 0.0050 \\
      SAINT { \citep{somepalli2021saint}} & 0.3355 $\pm$ 0.0043 & 0.5890 $\pm$ 0.0075 & 0.5560 $\pm$ 0.0066 \\
      SimCLR { \citep{chen2020simple}}& 0.3827 $\pm$ 0.0027 & 0.5837 $\pm$ 0.0093 & \underline{0.5602 $\pm$ 0.0055} \\
      SimCLR-Sample & 0.3498 $\pm$ 0.0541 & 0.5803 $\pm$ 0.0070 & 0.5479 $\pm$ 0.0016 \\
      SimCLR-Rank & \textbf{0.3868 $\pm$ 0.0026}  & 0.5843 $\pm$ 0.0062  & \underline{0.5609 $\pm$ 0.0040}  \\
      SimSiam { \citep{chen2021exploring}} & 0.3790 $\pm$ 0.0028 & \textbf{0.6100 $\pm$	0.0072} &  0.5189	$\pm$ 0.0096\\
    \bottomrule
    \end{tabular}
    \label{tab:all pretrain comparisons}
  \end{table*}

\textbf{Dataset.} We use MSLRWEB30K (or MSLR for brevity) \citep{mslr}, Yahoo Set1 (or Set1 for brevity) \citep{chapelle2011yahoo}, and Istella\_S \citep{lucchese2016post} (a smaller version of Istella \citep{dato2016fast}). In the tabular LTR literature, these three datasets are generally considered the standard;  most recent papers evaluate (only) on these datasets \citep{qin2021neural, ai2019learning, pang2020setrank, yang2022toward, yu2020ptranking}. To simulate scarcity of labels, we let 0.1\% of the training query groups in each dataset keep their labels, while the rest of the query groups have no labels. Dataset statistics are provided in \cref{app:datasets}. 

\textbf{Methodology.}
For the neural pretraining approaches (which include SimCLR-Rank, SimSiam, SimCLR, SubTab, VIME-self, DACL+, SAINT, and SCARF), the base of the model is the tabular ResNet\footnote{We emphasize that we do not prescribe any specific deep learning architecture choice; the choice for the best DL model may depend on the specific LTR task. Our proposed workflow of pretraining deep models in LTR is agnostic to the specific choice of DL architecture. Prior work has proposed DL architectures designed for click-through rate (CTR)/LTR based on factorization machines \citep{rendle2010factorization}, such as DeepFM and DCNv2 \citep{guo2017deepfm, wang2021dcn}, which we compare to in \cref{sec: sparse} and \cref{sec: sparse} against our pretrained models. Note that it is unclear currently how to do unsupervised pretraining using these CTR/LTR specific architectures at the moment.} (check the footnote for our explanation on the architecture choice). We also evaluate ``SimCLR-Sample'', a neural pretraining baseline we create to demonstrate the value of negatives from the same query group. SimCLR-Sample is like SimCLR-Rank, but at the beginning of each epoch the items are first randomly permuted and grouped into fake query groups. Then we apply SimCLR-Rank. Thus SimCLR-Sample has the same time/space complexity as SimCLR-Rank.
We pretrain on all the available query groups and finetune/train on only the labeled query groups. 

For the GBDT-based approaches, we  use the implementation in \texttt{lightgbm} \cite{ke2017lightgbm} and  adopt a tuning strategy similar to \citet{qin2021neural}. We report ``Supervised GBDT'' as GBDT utilizing only the labeled data, and in ``Semi-supervised GBDT'' we report the best among pseudo-labeled \citep{pseudo2013simple} and PCA-enriched \citep{duh2008learning} GBDTs as well as the combination of the two techniques. We also evaluate GBDTs enriched with SimCLR-Rank and SimSiam embeddings as additional features, which we call ``SimCLR-Rank + GBDT'' and ``SimSiam + GBDT''.
We describe our experimental setup in detail in \cref{app:exp-details}. This includes hyperparameters, procedures for data augmentation, GBDT pseudo-labeling/PCA details, model training, and finetuning. 
For comparison to prior baselines on tabular data, our experimental setup is chosen to mimic the original setup in them.

\textbf{Results.} \cref{tab:all pretrain comparisons} charts the test NDCG for each of the pretraining methods we study. We highlight a few observations:
(1) The pretraining-based DL methods almost all outperform GBDTs (both supervised and semi-supervised) in this label-scarce setting. 
Surprisingly, the semi-supervised GBDT baselines tend to achieve lower NDCG than the supervised GBDT baseline without semi-supervised learning.
(2) When  GBDTs are allowed to use representations learned from our pretraining methods, their performance consistently improves, though they do not outperform pretraining with DL.
(3) SimCLR-Rank and SimSiam perform the best or the second best among these methods for most of the datasets. In particular, SimCLR-Rank outperforms SimCLR-Sample, demonstrating the value and efficiency of negatives from the same query group.
As expected, SimCLR-Rank is also 7-14x faster than SimCLR/SCARF/DACL+ (\cref{tab:timecomparison} in \cref{app:runtime-comparison}).  
Given these advantages in terms of test NDCG and speed, we select SimCLR-Rank and SimSiam as the best overall pretraining methods for tabular LTR, and use them for our continued evaluation.

\section{Empirical evaluation}
Next we compare the best DL pretraining methods for tabular LTR (SimCLR-Rank and SimSiam) against GBDT and non-pretrained DL models on (1) the public datasets under different label scarcity patterns to simulate real-world scenarios\footnote{Code is provided at \url{https://github.com/houcharlie/ltr-pretrain/}.}, and (2) a private large-scale ranking dataset which is naturally label-scarce.

\subsection{Public datasets results}
For the public datasets (MSLRWEB30K, Yahoo Set1, Istella\_S; details given in \cref{sec:all pretrain comparisons}), we simulate label scarcity occurring in two scenarios: (1) \emph{crowdsourcing} of labels for a limited number of query groups, and  (2) using scarce \emph{implicit binary user feedback} (e.g.~clicks) as labels.

\subsubsection{Simulated crowdsourcing of labels}
\label{sec: relevance score}

We simulate a setting where labels are crowdsourced for only some query groups due to limited resources.

\textbf{Dataset.} For each of the three public datasets, we vary the fraction of labeled query groups in the training set in $\{0.001, 0.002, 0.005, 0.1, 0.5, 1.0\}$. Note that within each labeled query group all items are labeled.

\textbf{Methodology.} We repeat the methodology from \cref{sec:all pretrain comparisons} for the GBDT and the SimCLR-Rank and SimSiam pretraining methods. For each dataset and labeling fraction, we report the test metric of the best GBDT ranker among semi-supervised GBDTs and supervised GBDTs under ``GBDT''.
Similarly, under ``no pretrain DL'' we report the best supervised DL model amongst tabular ResNet \citep{gorishniy2021revisiting}, DeepFM \citep{guo2017deepfm}, and DCNv2 \citep{wang2021dcn} under the name ``no pretrain DL''.
Finally, we report the best pretrained ranker amongst SimCLR-Rank and SimSiam under the name ``pretrained DL''. Pretrained DL rankers use tabular ResNet (see \cref{app:exp-details} for more details). During hyperparameter tuning, we tune for NDCG and report the resulting test NDCG and test Outlier-NDCG.

\textbf{Results.}  We compare pretrained rankers, non-pretrained DL rankers, and GBDTs on public datasets in \cref{fig:fractions}. The results for Outlier-NDCG are provided in \cref{fig:outlier fractions} in \cref{app:public-relevance}. We find that pretrained rankers outperform non-pretrained methods (including GBDTs) on NDCG and Outlier-NDCG across all public datasets, up to a dataset-dependent fraction of QGs labeled, as shown in \cref{fig:fractions} (for NDCG) and \cref{fig:outlier fractions} (for Outlier-NDCG). We also note that GBDTs have better NDCG than non-pretrained DL models in most of these regimes. This shows that there exist scenarios with a limited supply of labeled query groups, where self-supervised pretraining is the factor that allows deep learning models to outperform GBDTs. We provide a detailed comparison between semi-supervised and supervised GBDT models in \cref{app:gbdt comparison}.

\subsubsection{Simulated implicit binary user feedback}
\label{sec: sparse}
Here we simulate a scenario when users of an LTR system provide binary implicit feedback for the items, like the clicking of an advertisement on a webpage. This kind of feedback is usually infrequent in an LTR system, as it is the result of a user choosing to spend more time or resources on an item in a presented list. Therefore, most query groups will not have any labels, and even labeled query groups will only have a few positively labeled items. On the other hand, such labels are often cheaper to collect than crowdsourced labels, so they are used in many industry use cases.

\textbf{Dataset.} To simulate this scenario, we follow the methodology from \citet{yang2022toward}, to generate independent stochastic binary labels for each item from its true relevance label for training and validation sets of each of the public datasets. 
Note that we still use the true (latent) relevance labels in the test set for evaluation. 
This models a scenario where we observe binary labels that are noisy observations of a true latent label, but the task is to use these noisy labels to learn to rank according to the true labels. 
The details of this binarizing transformation are provided in \cref{app:binary-label}.
In this transformation, a parameter, $\tau_{\text{target}}$, implicitly controls how sparse the binary labels are (larger is more sparse). 
We select $\tau_{\text{target}} = 4.5$, to obtain 8.9\%, 5.3\%, and 25.2\% labeled query groups which contain at least one positive label in MSLR, Yahoo Set1, Istella\_S datasets, respectively. We present detailed data statistics in \cref{tab:tau 4.5 stats} of Appendix.

\begin{table*}[t]
  \centering
  \caption{\textbf{Main result (Simulated implicit binary user feedback):} 
  When $X$\% (MSLRWEB30K=8.9\%, Yahoo Set1=5.3\%, and in Istella\_S=25.2\%) of query groups are assigned a few binary labels, pretrained DL method outperform both GBDT and non-pretrained DL models. We generate labels with $\tau_{\text{target}}=4.5$ \pcref{sec: sparse}. ${}^\clubsuit$ means that pretrained models are significantly better in NDCG than non-pretrained models,  measured by a t-test with significance $p < 0.05$.}
    \begin{tabular}{c || c | c | c }
      \toprule
      Method&MSLRWEB30K ($\uparrow$)  & Yahoo Set1 ($\uparrow$)  & Istella ($\uparrow$)  \\
      \midrule
      & \multicolumn{3}{c}{NDCG ($\uparrow$)}  \\
      \midrule
      GBDT & 0.346 $\pm$ 0.001 & 0.617 $\pm$ 0.000 & 0.602 $\pm$ 0.000  \\
      No Pretrain DL & 0.333 $\pm$ 0.001  & 0.621 $\pm$ 0.000 & 0.608 $\pm$ 0.002 \\
      Pretrained DL & \textbf{0.356 $\pm$ 0.004}${}^\clubsuit$  & \textbf{0.622 $\pm$ 0.001} & \textbf{0.613 $\pm$ 0.002}${}^\clubsuit$ \\
      \midrule
      & \multicolumn{3}{c}{Outlier-NDCG ($\uparrow$)}  \\
      \midrule
      GBDT & 0.288 $\pm$ 0.000 & 0.540 $\pm$ 0.000 & 0.678 $\pm$ 0.000  \\
      No Pretrain DL & 0.238 $\pm$ 0.026  & 0.536 $\pm$ 0.003  & 0.687 $\pm$ 0.015 \\
      Pretrained DL & \textbf{0.289 $\pm$ 0.012}  & \textbf{0.543 $\pm$ 0.002}${}^\clubsuit$  & \textbf{0.699 $\pm$ 0.003}  \\
    \bottomrule
    \end{tabular}
    \label{tab:tau 4.5}
\end{table*}

\textbf{Methodology.} We reuse the methodology of \cref{sec: relevance score}, except we use a linear scoring head for pretrained models to improve stability and performance (see \cref{app:exp-details}). Similar to \cref{sec: relevance score}, we report the best of multiple models for each method and dataset (see \cref{sec: relevance score} for details).

\textbf{Results.} In \cref{tab:tau 4.5}, we present the test NDCG and Outlier-NDCG metrics of the models under this setting. We see that pretrained rankers significantly outperform GBDTs and non-pretrained DL rankers in this simulated binary user feedback setting with up to, even with 25.2\% labeled query groups (Istella\_S). 
For MSLRWEB30K and Yahoo Set1 datasets, the second best model is the GBDT, and the non-pretrained DL ranker has a much worse Outlier-NDCG. This aligns with recent observations that GBDTs perform better than DL models in supervised training with datasets containing irregularities \cite{mcelfresh2023neural}. However, pretraining helps DL rankers close this gap, and at times beat GBDTs in outlier performance.
We also present results for $\tau_{\text{target}} \in \{4.25, 5.1\}$ in \cref{sec: sparse appendix}. There, we find that making the labels sparser ($\tau_{\text{target}} = 5.1$) increases the relative improvement of pretrained DL rankers, while in a less sparse setting ($\tau_{\text{target}} = 4.25$) GBDTs dominate.
This suggests that pretrained DL models outperform GBDTs when learning from rare implicit binary user feedback, as is common in search and recommendation systems. 

\subsection{Large-scale real-world dataset results}
\label{sec:amazon}

In this section we test whether the results from the simulated label scarce settings above, especially \cref{sec: sparse}, translate to a large-scale real-world LTR problem.

\textbf{Dataset.} Our proprietary dataset is derived from the online shopping logs of a large online retailer, numbering in the tens of millions of query groups. In this dataset, query groups consist of items presented to a shopper if they enter a search query. 
We assign each item a purchase label: label is 1 if the shopper purchased the item, or 0 if the shopper did not purchase the item.  
Hence, this is a real-world instance of the implicit binary user feedback setting simulated in \cref{sec: sparse}.
Further, only about 10\% of the query groups have items with non-zero labels. 
See \cref{app:datasets} for more dataset details.


\textbf{Methodology.} We compare the performance of pretrained DL models--SimCLR-Rank and SimSiam--with GBDT and non-pretrained DL models. SimSiam and SimCLR-Rank are first pretrained on all the query groups before being finetuned on the labeled query groups with some non-zero purchase labels. The GBDT and the non-pretrained models are only trained on the labeled data. More details about these models and training strategies is given in \cref{sec:large_dataset_expt_details}.

\begin{table}
    \centering
      \caption{{\bf Main result (Real-world dataset):} $\%$ improvement of rankers pretrained by SimSiam and SimCLR-Rank over GBDTs on a large-scale proprietary online shopping dataset.  Unsupervised pretraining improves  performance (1)  overall (full dataset) (internally, 2\% is a big improvement), and (2) on outliers. 
      }
      \label{tab:amazon}
      \setlength\tabcolsep{3pt}
      \begin{tabular}{l | c c } 
        \toprule
        {\bf Method} & {\bf NDCG} ($\uparrow$) & {\bf Outlier-NDCG} ($\uparrow$) \\ 
        \midrule
        & \multicolumn{2}{c}{{\em $\Delta$\% change w.r.t.~GBDT}} \\ 
        \midrule
        GBDT  & {+0.00\%} & {+0.00\%} \\ 
        No Pretrain DL &{+3.82\% $\pm$ 0.14\%} & {-6.97\% $\pm$ 0.43\%} \\ 
        SimCLR-Rank & +4.00\%  $\pm$ 0.14\% & -4.85\% $\pm$ 0.44\%  \\
        SimSiam &\textbf{+5.64\%  $\pm$ 0.14\%} & \textbf{+2.75\% $\pm$ 0.43\%} \\
      \bottomrule
    \end{tabular}
    \vspace{-0.15in}
\end{table}

\textbf{Results.}  
Our results are summarized in \cref{tab:amazon} and all numbers are given as relative percentage improvements ($\Delta \%$) over the GBDT ranker. First, note that all the DL models outperform GBDT models in terms of NDCG. Amongst them, the pretrained model using SimSiam performs the best with a gain of over $5.5\%$. This is substantial, given that in large-scale industry datasets, even a 2\% improvement in NDCG is considered significant.
In terms of Outlier-NDCG, the story is very different. Here, the non-pretrained DL model and SimCLR-Rank performs up to -7\% worse than the GBDT. However, SimSiam pretraining allows the DL ranker to overcome this shortcoming and beat the outlier performance of the GBDT.
This verifies our hypothesis from \cref{sec: sparse} that pretraining can help DL models outperform GBDTs on large-scale real-world LTR problems, especially with implicit user feedback.
Lastly, we see a significant performance difference between SimSiam and SimCLR-Rank. We investigate this in an ablation  with public datasets in \cref{sec:ablations}.




\subsection{Ablations: Choices for pretrained models in LTR}
\label{sec:ablations}
Our experiments suggest that the best pretraining and finetuning strategies for the LTR problem are different from the known best practices for image or text tasks. We make a few observations. (1) SimSiam and SimCLR-Rank perform significantly differently depending on the dataset. (2) Combining models pretrained by SimCLR-Rank and SimSiam can produce a model that is competitive against SimCLR-Rank and SimSiam across all datasets, producing a unified approach that may be performant across many settings. (3) Full finetuning of the model performs much better than {\em linear probing}, which only finetunes a linear layer over frozen pretrained embeddings. In the remainder of this section, we elaborate on these findings in more detail. 

\textbf{(1) SimCLR-Rank and SimSiam perform substantially differently depending on the dataset.}
SimSiam and SimCLR-Rank perform significantly differently depending on the dataset, which is different from the observations of \citet{chen2021exploring}, who proposed SimSiam and found that it performs similarly to SimCLR derivatives. We next compare SimCLR-Rank with SimSiam at different levels of label scarcity.

\begin{figure*}[h]
  \centering
  \begin{subfigure}[b]{0.32\textwidth}
      \centering
      \includegraphics[width=\textwidth]{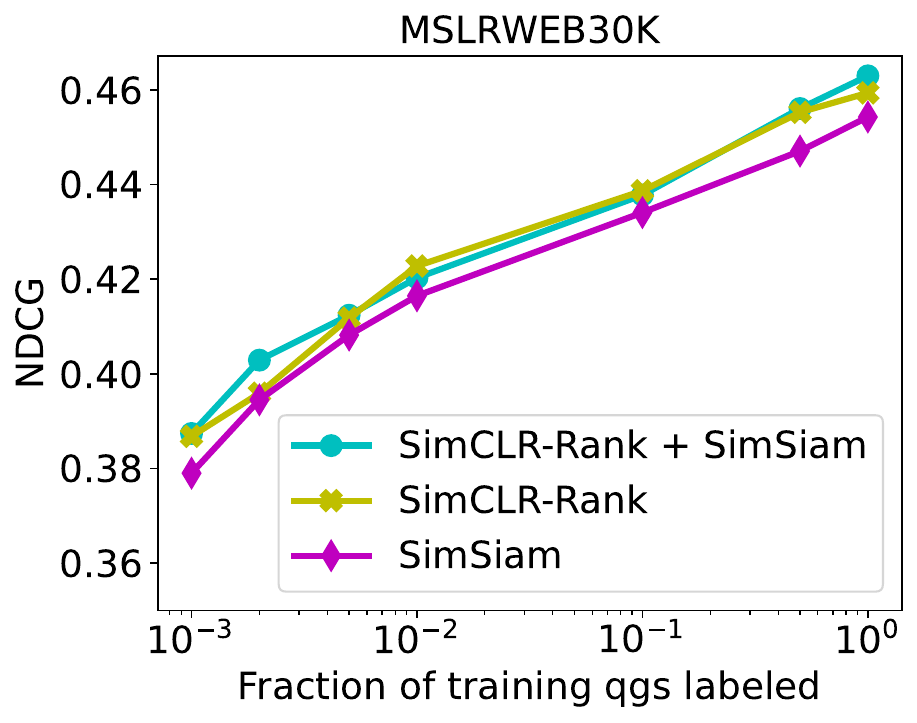}
      \label{fig:pretrain mslr}
  \end{subfigure}
  \begin{subfigure}[b]{0.32\textwidth}
      \centering
      \includegraphics[width=\textwidth]{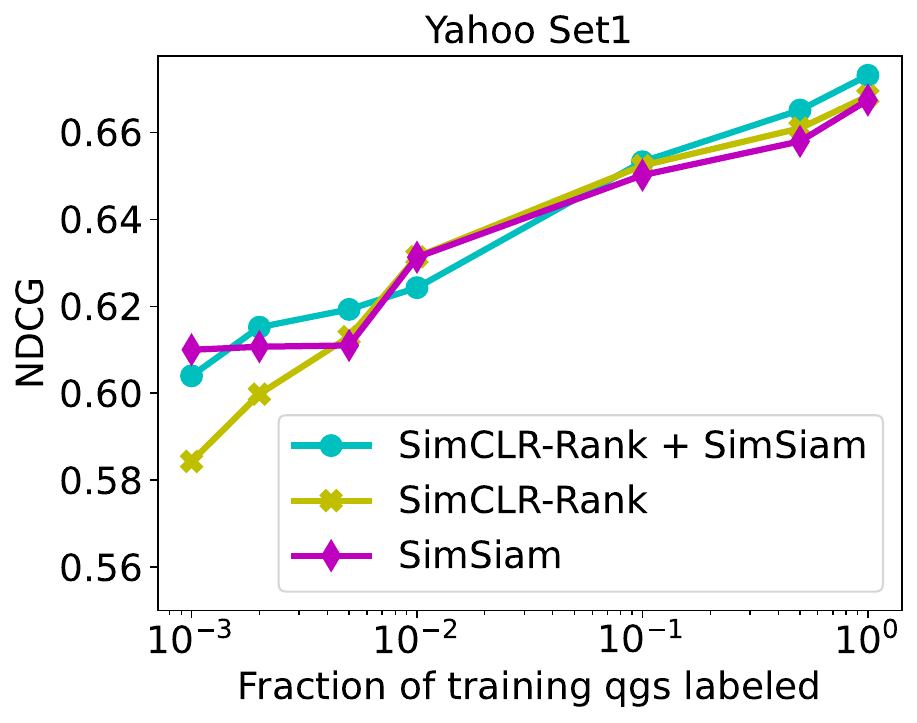}
      \label{fig:pretrain yahoo}
  \end{subfigure}
  \begin{subfigure}[b]{0.32\textwidth}
    \centering
    \includegraphics[width=\textwidth]{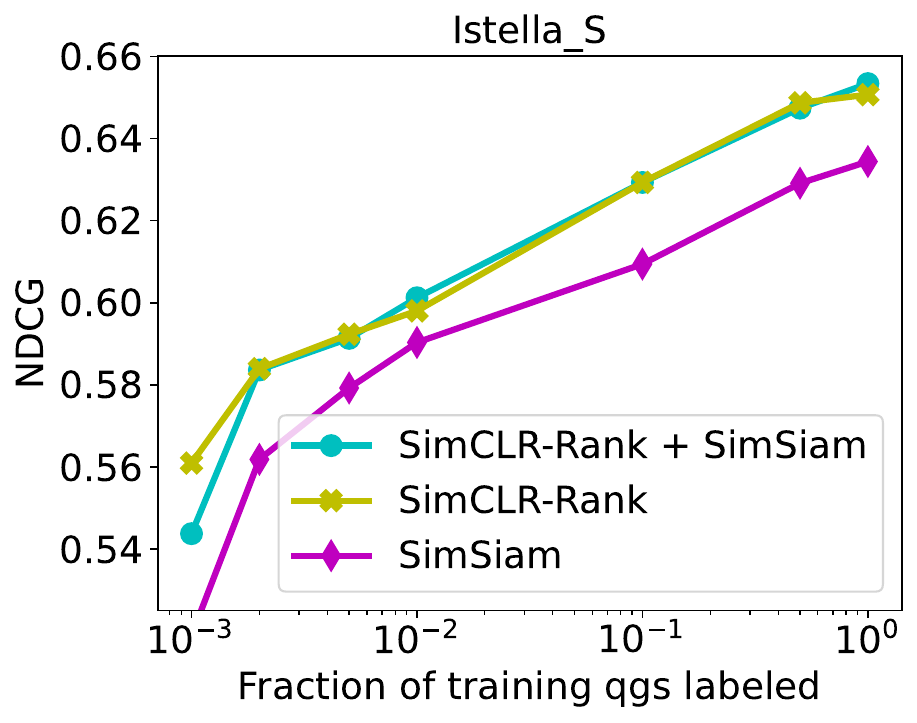}
    \label{fig:pretrain istella}
\end{subfigure}
     \caption{Comparison of SimCLR-Rank and SimSiam pretraining strategies on three public datasets with varying fraction of labeled query groups. Neither method consistently dominate the other in all the datasets. However, by combining SimCLR-Rank and SimSiam encoders, we can produce models that are competitive across all datasets. Data points are averages over 3 trials.  
     }
     \label{fig: pretrain comparison}
     \vspace{-0.1in}
\end{figure*}

\begin{figure}[h]
  \centering
  \begin{subfigure}[b]{0.32\textwidth}
      \centering
      \includegraphics[width=\textwidth]{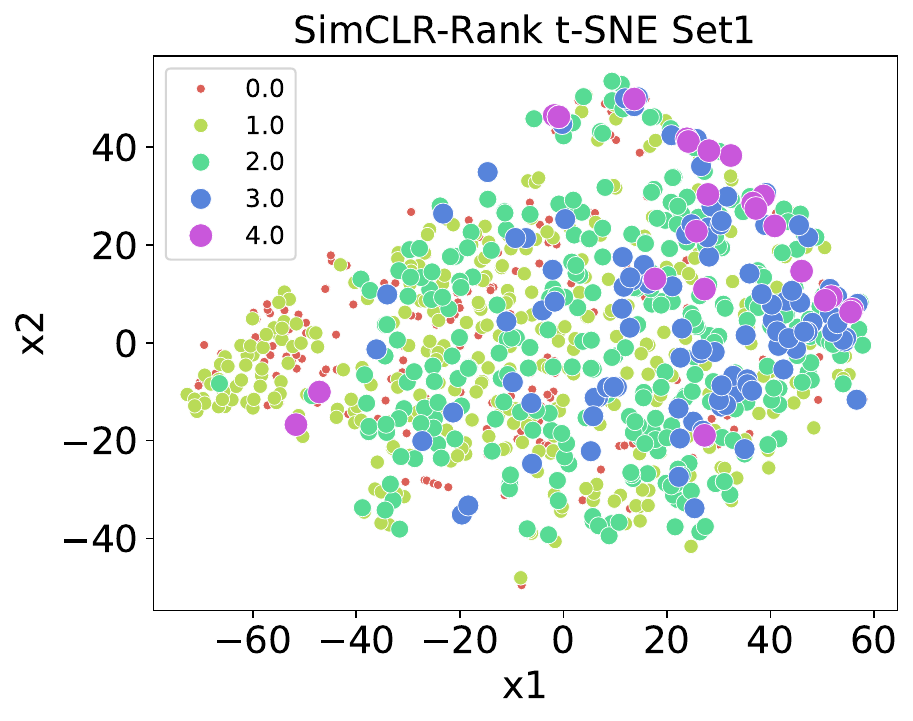}
      \label{fig:tsne set1 simclrrank}
  \end{subfigure}
  \begin{subfigure}[b]{0.32\textwidth}
      \centering
      \includegraphics[width=\textwidth]{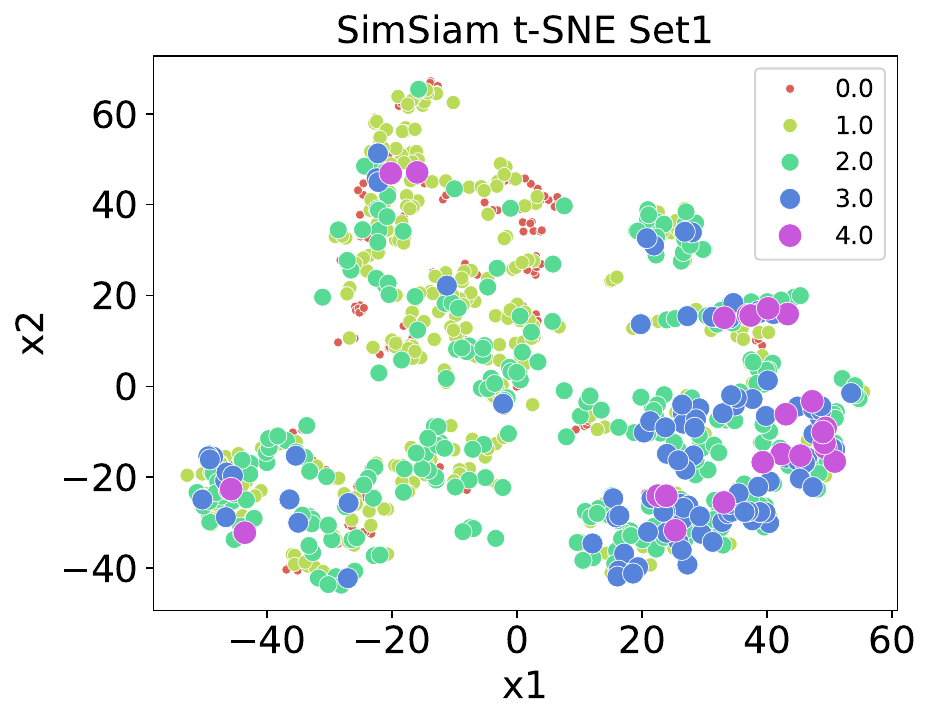}
      \label{fig:tsne set1 simsiam}
  \end{subfigure}
  \begin{subfigure}[b]{0.32\textwidth}
      \centering
      \includegraphics[width=\textwidth]{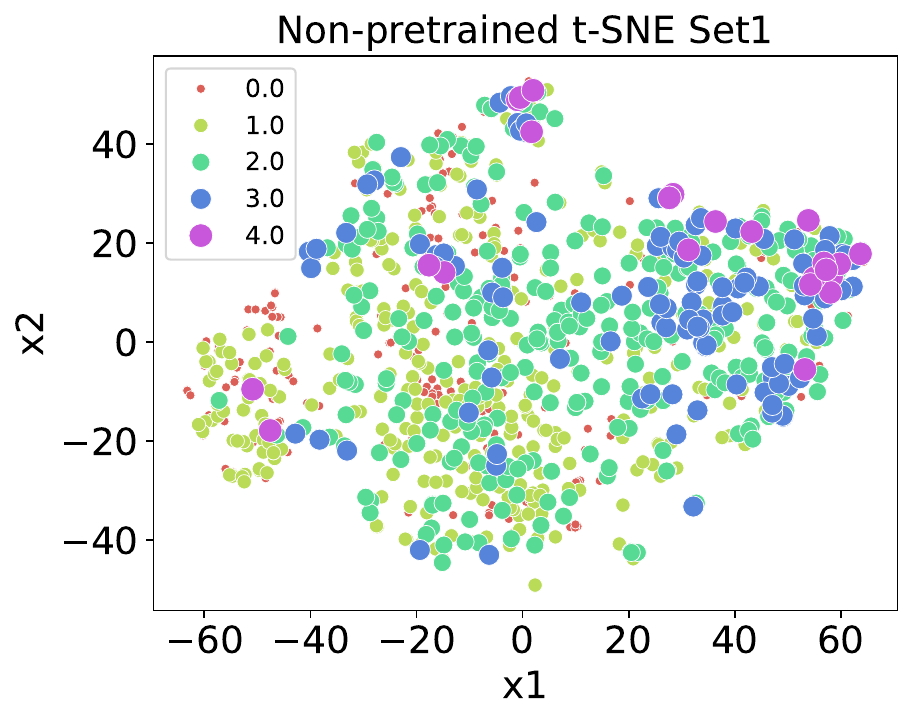}
      \label{fig:tsne set1 supervised}
  \end{subfigure}
     \caption{t-SNE plots of SimCLR-Rank, SimSiam, and a non-pretrained model trained on 0.1\% of the data for the Yahoo Set1 dataset. Both SimCLR-Rank and SimSiam use the zeros augmentation ($p=0.1$) to remove augmentation choice as a confounding variable. Relevance label is indicated by marker size/color. We see that SimCLR-Rank and SimSiam embeddings are qualitatively very different. Qualitatively, SimCLR-Rank and SimSiam embeddings separate different relevance labels better than the non-pretrained model. For this setting, SimSiam has the best downstream performance, and also gives the most clustered embeddings.} 
     \label{fig: tsne plots for set1 yahoo}
\end{figure}

\textit{Methodology.}  
The dataset and methodology follows that of \cref{sec: relevance score}, except we let (1) SimCLR-Rank use the Gaussian augmentation with scale 1, and (2) SimSiam use the zeroing augmentation with probability 0.1. These augmentations generally perform well for their respective methods.
Experiments illustrating the effect of each augmentation method are included in \cref{sec: appendix augmentations}.

\textit{Results.} In \cref{fig: pretrain comparison}, SimCLR-Rank performs better on MSLRWEB30K and Istella\_S, while SimSiam performs better on Yahoo Set1 when labels are scarce. 
\cref{fig: tsne plots for set1 yahoo} shows the t-SNE projections of sub-sampled item embeddings learned by SimCLR-Rank, SimSiam, and pre-final layer of a fully supervised model for Yahoo Set1. SimSiam exhibit qualitatively more clustered embeddings than SimCLR-Rank, particularly for high relevance labels. 
Similar t-SNE plots for other datasets are in \cref{app:simclr vs simsiam}.

\textbf{(2) Combining models pretrained by SimCLR-Rank and SimSiam produces a model that has the strengths of both.} To make steps towards a single recommendation for unsupervised pretraining in LTR, we unify SimCLR-Rank and SimSiam into a single method ``SimCLR-Rank + SimSiam''.

\textit{Methodology.} To produce a ``SimCLR-Rank + SimSiam'' model, we finetune a linear layer over the embeddings from a finetuned SimCLR-Rank concatenated with the embeddings from a finetuned SimCLR-Rank model. More details about how we combine the models is given in \cref{sec: appendix combine}.

\textit{Results.} In \cref{fig: pretrain comparison} we find that ``SimCLR-Rank + SimSiam'' performs equal to or better than SimCLR-Rank and SimSiam across many different percentages of labeled query groups. While the combined model does not uniformly dominate across all labeled query group percentages, it is a promising first step towards a single unified method for unsupervised pretraining in ranking.

\textbf{(3) Full finetuning outperforms linear probing.} 
Linear probing is a popular finetuning strategy in text and image domains, where it produces  good results \citep{chen2021exploring, chen2020simple, peters2019tune, kumar2022fine}. 
We next compare it with the full finetuning strategy in tabular LTR.

\textit{Methodology.} We use the methodology from \cref{sec:all pretrain comparisons} to compare 3 finetuning strategies: linear probing (LP), full-finetuning (FF), and multilayer probing (MP) for SimSiam and SimCLR-Rank. MP tunes a 3-layer MLP head on top of the frozen pretrained embeddings. Experimental details are in \cref{app:exp-details}.


\textit{Results.}  \Cref{tab:ff-or-lp} provides results on MSLRWEB30K dataset. We find that in terms of NDCG of both SimCLR-Rank and SimSiam, LP is the worst and FF is the best. Interestingly, MP comes as a close second in SimCLR-Rank, but it performs poorly on SimSiam. Our results on other datasets are similar (\cref{app:lp-vs-ff}). We thus recommend full fine-tuning as a stable and performant strategy.  
To explain the above observations, we we can look again at the t-SNE projections of the embeddings generated by the SimSiam and SimCLR-Rank encoders (pretrained using the zeros augmentation with $p=0.1$ to remove augmentation choice as a confounding factor, finetuned on 0.1\% of the data) and the pre-final layer of a model trained only on 0.1\% of the data on a randomly sampled 1000 samples from the full training set \pcref{fig: tsne plots for set1 yahoo}.

These plots offer the following explanations for the two phenomena. 
(1) The embeddings produced by the SimSiam/SimCLR-Rank are not sorted by relevance score in the projection space, unlike the fully supervised encoder. 
This suggests that a linear ranker cannot use the pretrained embeddings to predict the true relevance labels.
(2) MP works better for SimCLR-Rank than SimSiam because SimCLR-Rank's embeddings are more evenly spread out (less collapsed) than those of SimSiam. This allows an MLP trained on the SimCLR-Rank embeddings to distinguish the items with different relevance labels.

\begin{table}[]
\centering
\caption{{\bf Finetuning}: 
Comparison of fine   tuning strategies on MSLRWEB30K dataset (LP = Linear Probing, MP = Multilayer Probing, FF = Full Finetuning). 
Full finetuning consistently performs best or near-best, while
linear probing \textit{and} multilayer probing perform poorly.}
\begin{tabular}{ccc}
\toprule
\multicolumn{1}{c|}{\textbf{Method}} & \multicolumn{1}{c|}{\textbf{SimSiam}}                          & \textbf{SimCLR-Rank}                      \\ \hline
\multicolumn{3}{c}{\emph{NDCG} ($\uparrow$)}                                                                                                                         \\ \hline
\multicolumn{1}{c|}{LP}     & \multicolumn{1}{c|}{0.2679 $\pm$ 0.0007}                           & 0.3219 $\pm$ 0.0224                           \\
\multicolumn{1}{c|}{MP}     & \multicolumn{1}{c|}{0.2764 $\pm$ 0.0001}                           & 0.3890 $\pm$ 0.0011                           \\
\multicolumn{1}{c|}{FF}     & \multicolumn{1}{c|}{\textbf{0.3935 $\pm$ 0.0034}} &  \textbf{0.3959 $\pm$ 0.0022} \\ \hline
\multicolumn{3}{c}{\emph{Outlier-NDCG}  ($\uparrow$)}                                                                                                                 \\ \hline
\multicolumn{1}{c|}{LP}     & \multicolumn{1}{c|}{0.1803 $\pm$ 0.0033}                           & 0.2304 $\pm$ 0.0332                           \\
\multicolumn{1}{c|}{MP}     & \multicolumn{1}{c|}{0.1749 $\pm$ 0.0023}                           &   \textbf{0.2969 $\pm$ 0.0009} \\
\multicolumn{1}{c|}{FF}     & \multicolumn{1}{c|}{ \textbf{0.3149 $\pm$ 0.0119}} & 0.2892 $\pm$ 0.0025                           \\ 
\bottomrule
\end{tabular}
\label{tab:ff-or-lp}
\vspace{-0.15in}
\end{table}




\section{Conclusion}

We study the learning-to-rank (LTR) problem with tabular data under a  scarcity of labeled data---a
 common scenario in real-world practical LTR systems. 
Prior works on supervised learning with tabular data have shown that GBDTs outperform deep learning (DL) models, especially on datasets with outliers.
In this paper, we find that in the label-scarce setting,  DL pretraining methods can exploit available unlabeled data to obtain new state-of-the art performance.
Through experiments with public LTR datasets and a real-world large-scale online shopping dataset, we show that pretraining allows DL rankers to outperform GBDT rankers, especially on outlier data. 
Finally, we provide guidelines for pretraining on LTR datasets. 
There remain several open questions around pretraining for tabular and LTR data. It is still unknown whether we can achieve knowledge transfer across different tabular datasets, as is commonly done in image and text domains. It will also be useful to investigate LTR-specific pretraining methods which can further improve the ranking performance.

\section*{Acknowledgments}
GF and CH gratefully acknowledge the support of Google, Intel, the Sloan Foundation, and Bosch.

\bibliography{main}

\begin{thebibliography}{142}
\providecommand{\natexlab}[1]{#1}
\providecommand{\url}[1]{\texttt{#1}}
\expandafter\ifx\csname urlstyle\endcsname\relax
  \providecommand{\doi}[1]{doi: #1}\else
  \providecommand{\doi}{doi: \begingroup \urlstyle{rm}\Url}\fi

\bibitem[Ai et~al.(2018)Ai, Bi, Guo, and Croft]{ai2018learning}
Qingyao Ai, Keping Bi, Jiafeng Guo, and W~Bruce Croft.
\newblock Learning a deep listwise context model for ranking refinement.
\newblock In \emph{The 41st international ACM SIGIR conference on research \& development in information retrieval}, pp.\  135--144, 2018.

\bibitem[Ai et~al.(2019)Ai, Wang, Bruch, Golbandi, Bendersky, and Najork]{ai2019learning}
Qingyao Ai, Xuanhui Wang, Sebastian Bruch, Nadav Golbandi, Michael Bendersky, and Marc Najork.
\newblock Learning groupwise multivariate scoring functions using deep neural networks.
\newblock In \emph{Proceedings of the 2019 ACM SIGIR international conference on theory of information retrieval}, pp.\  85--92, 2019.

\bibitem[Arik \& Pfister(2021)Arik and Pfister]{tabnet}
Sercan Arik and Tomas Pfister.
\newblock Tabnet: Attentive interpretable tabular learning.
\newblock 2021.

\bibitem[Bahri et~al.(2021)Bahri, Jiang, Tay, and Metzler]{bahri2021scarf}
Dara Bahri, Heinrich Jiang, Yi~Tay, and Donald Metzler.
\newblock Scarf: Self-supervised contrastive learning using random feature corruption.
\newblock \emph{arXiv preprint arXiv:2106.15147}, 2021.

\bibitem[Berthelot et~al.(2019)Berthelot, Carlini, Goodfellow, Papernot, Oliver, and Raffel]{berthelot2019mixmatch}
David Berthelot, Nicholas Carlini, Ian Goodfellow, Nicolas Papernot, Avital Oliver, and Colin~A Raffel.
\newblock Mixmatch: A holistic approach to semi-supervised learning.
\newblock \emph{Advances in neural information processing systems}, 32, 2019.

\bibitem[Borisov et~al.(2022)Borisov, Leemann, Se{\ss}ler, Haug, Pawelczyk, and Kasneci]{borisov2022deep}
Vadim Borisov, Tobias Leemann, Kathrin Se{\ss}ler, Johannes Haug, Martin Pawelczyk, and Gjergji Kasneci.
\newblock Deep neural networks and tabular data: A survey.
\newblock \emph{IEEE transactions on neural networks and learning systems}, 2022.

\bibitem[Bower et~al.(2021)Bower, Eftekhari, Yurochkin, and Sun]{bower2021individually}
Amanda Bower, Hamid Eftekhari, Mikhail Yurochkin, and Yuekai Sun.
\newblock Individually fair ranking.
\newblock \emph{arXiv preprint arXiv:2103.11023}, 2021.

\bibitem[Brown et~al.(2020)Brown, Mann, Ryder, Subbiah, Kaplan, Dhariwal, Neelakantan, Shyam, Sastry, Askell, et~al.]{brown2020language}
Tom Brown, Benjamin Mann, Nick Ryder, Melanie Subbiah, Jared~D Kaplan, Prafulla Dhariwal, Arvind Neelakantan, Pranav Shyam, Girish Sastry, Amanda Askell, et~al.
\newblock Language models are few-shot learners.
\newblock \emph{Advances in neural information processing systems}, 33:\penalty0 1877--1901, 2020.

\bibitem[Bruch et~al.(2019)Bruch, Zoghi, Bendersky, and Najork]{bruch2019revisiting}
Sebastian Bruch, Masrour Zoghi, Michael Bendersky, and Marc Najork.
\newblock Revisiting approximate metric optimization in the age of deep neural networks.
\newblock In \emph{Proceedings of the 42nd international ACM SIGIR conference on research and development in information retrieval}, pp.\  1241--1244, 2019.

\bibitem[Burges et~al.(2005)Burges, Shaked, Renshaw, Lazier, Deeds, Hamilton, and Hullender]{burges2005learning}
Chris Burges, Tal Shaked, Erin Renshaw, Ari Lazier, Matt Deeds, Nicole Hamilton, and Greg Hullender.
\newblock Learning to rank using gradient descent.
\newblock In \emph{Proceedings of the 22nd international conference on Machine learning}, pp.\  89--96, 2005.

\bibitem[Burges(2010)]{burges2010ranknet}
Christopher~JC Burges.
\newblock From ranknet to lambdarank to lambdamart: An overview.
\newblock \emph{Learning}, 11\penalty0 (23-581):\penalty0 81, 2010.

\bibitem[Cai et~al.(2021)Cai, Zheng, Chen, Jagadish, Ooi, and Zhang]{cai2021arm}
Shaofeng Cai, Kaiping Zheng, Gang Chen, HV~Jagadish, Beng~Chin Ooi, and Meihui Zhang.
\newblock Arm-net: Adaptive relation modeling network for structured data.
\newblock In \emph{Proceedings of the 2021 International Conference on Management of Data}, pp.\  207--220, 2021.

\bibitem[Cao et~al.(2007)Cao, Qin, Liu, Tsai, and Li]{cao2007learning}
Zhe Cao, Tao Qin, Tie-Yan Liu, Ming-Feng Tsai, and Hang Li.
\newblock Learning to rank: from pairwise approach to listwise approach.
\newblock In \emph{Proceedings of the 24th international conference on Machine learning}, pp.\  129--136, 2007.

\bibitem[Chapelle \& Chang(2011)Chapelle and Chang]{chapelle2011yahoo}
Olivier Chapelle and Yi~Chang.
\newblock Yahoo! learning to rank challenge overview.
\newblock In \emph{Proceedings of the learning to rank challenge}, pp.\  1--24. PMLR, 2011.

\bibitem[Chen et~al.(2020)Chen, Kornblith, Norouzi, and Hinton]{chen2020simple}
Ting Chen, Simon Kornblith, Mohammad Norouzi, and Geoffrey Hinton.
\newblock A simple framework for contrastive learning of visual representations.
\newblock In \emph{International conference on machine learning}, pp.\  1597--1607. PMLR, 2020.

\bibitem[Chen et~al.(2009)Chen, Liu, Lan, Ma, and Li]{chen2009ranking}
Wei Chen, Tie-Yan Liu, Yanyan Lan, Zhi-Ming Ma, and Hang Li.
\newblock Ranking measures and loss functions in learning to rank.
\newblock \emph{Advances in Neural Information Processing Systems}, 22, 2009.

\bibitem[Chen \& He(2021)Chen and He]{chen2021exploring}
Xinlei Chen and Kaiming He.
\newblock Exploring simple siamese representation learning.
\newblock In \emph{Proceedings of the IEEE/CVF conference on computer vision and pattern recognition}, pp.\  15750--15758, 2021.

\bibitem[Cheng et~al.(2016)Cheng, Koc, Harmsen, Shaked, Chandra, Aradhye, Anderson, Corrado, Chai, Ispir, et~al.]{cheng2016wide}
Heng-Tze Cheng, Levent Koc, Jeremiah Harmsen, Tal Shaked, Tushar Chandra, Hrishi Aradhye, Glen Anderson, Greg Corrado, Wei Chai, Mustafa Ispir, et~al.
\newblock Wide \& deep learning for recommender systems.
\newblock In \emph{Proceedings of the 1st workshop on deep learning for recommender systems}, pp.\  7--10, 2016.

\bibitem[Darabi et~al.(2021)Darabi, Fazeli, Pazoki, Sankararaman, and Sarrafzadeh]{darabi2021contrastive}
Sajad Darabi, Shayan Fazeli, Ali Pazoki, Sriram Sankararaman, and Majid Sarrafzadeh.
\newblock Contrastive mixup: Self-and semi-supervised learning for tabular domain.
\newblock \emph{arXiv preprint arXiv:2108.12296}, 2021.

\bibitem[Dato et~al.(2016)Dato, Lucchese, Nardini, Orlando, Perego, Tonellotto, and Venturini]{dato2016fast}
Domenico Dato, Claudio Lucchese, Franco~Maria Nardini, Salvatore Orlando, Raffaele Perego, Nicola Tonellotto, and Rossano Venturini.
\newblock Fast ranking with additive ensembles of oblivious and non-oblivious regression trees.
\newblock \emph{ACM Transactions on Information Systems (TOIS)}, 35\penalty0 (2):\penalty0 1--31, 2016.

\bibitem[Devlin et~al.(2018)Devlin, Chang, Lee, and Toutanova]{devlin2018bert}
Jacob Devlin, Ming-Wei Chang, Kenton Lee, and Kristina Toutanova.
\newblock Bert: Pre-training of deep bidirectional transformers for language understanding.
\newblock \emph{arXiv preprint arXiv:1810.04805}, 2018.

\bibitem[Duh \& Kirchhoff(2008)Duh and Kirchhoff]{duh2008learning}
Kevin Duh and Katrin Kirchhoff.
\newblock Learning to rank with partially-labeled data.
\newblock In \emph{Proceedings of the 31st annual international ACM SIGIR conference on Research and development in information retrieval}, pp.\  251--258, 2008.

\bibitem[Friedman(2001)]{friedman2001greedy}
Jerome~H Friedman.
\newblock Greedy function approximation: a gradient boosting machine.
\newblock \emph{Annals of statistics}, pp.\  1189--1232, 2001.

\bibitem[Frosst \& Hinton(2017)Frosst and Hinton]{frosst2017distilling}
Nicholas Frosst and Geoffrey Hinton.
\newblock Distilling a neural network into a soft decision tree.
\newblock \emph{arXiv preprint arXiv:1711.09784}, 2017.

\bibitem[Ge et~al.(2021)Ge, Mishra, Li, Wang, and Jacobs]{ge2021robust}
Songwei Ge, Shlok Mishra, Chun-Liang Li, Haohan Wang, and David Jacobs.
\newblock Robust contrastive learning using negative samples with diminished semantics.
\newblock \emph{Advances in Neural Information Processing Systems}, 34:\penalty0 27356--27368, 2021.

\bibitem[Ge(2018)]{ge2018deep}
Weifeng Ge.
\newblock Deep metric learning with hierarchical triplet loss.
\newblock In \emph{Proceedings of the European conference on computer vision (ECCV)}, pp.\  269--285, 2018.

\bibitem[Glass et~al.(2022)Glass, Rossiello, Chowdhury, Naik, Cai, and Gliozzo]{glass2022re2g}
Michael Glass, Gaetano Rossiello, Md~Faisal~Mahbub Chowdhury, Ankita~Rajaram Naik, Pengshan Cai, and Alfio Gliozzo.
\newblock Re2g: Retrieve, rerank, generate.
\newblock \emph{arXiv preprint arXiv:2207.06300}, 2022.

\bibitem[Goren et~al.(2018)Goren, Kurland, Tennenholtz, and Raiber]{goren2018ranking}
Gregory Goren, Oren Kurland, Moshe Tennenholtz, and Fiana Raiber.
\newblock Ranking robustness under adversarial document manipulations.
\newblock In \emph{The 41st International ACM SIGIR Conference on Research \& Development in Information Retrieval}, pp.\  395--404, 2018.

\bibitem[Gorishniy et~al.(2021)Gorishniy, Rubachev, Khrulkov, and Babenko]{gorishniy2021revisiting}
Yury Gorishniy, Ivan Rubachev, Valentin Khrulkov, and Artem Babenko.
\newblock Revisiting deep learning models for tabular data.
\newblock \emph{Advances in Neural Information Processing Systems}, 34:\penalty0 18932--18943, 2021.

\bibitem[Grill et~al.(2020)Grill, Strub, Altch{\'e}, Tallec, Richemond, Buchatskaya, Doersch, Avila~Pires, Guo, Gheshlaghi~Azar, et~al.]{grill2020bootstrap}
Jean-Bastien Grill, Florian Strub, Florent Altch{\'e}, Corentin Tallec, Pierre Richemond, Elena Buchatskaya, Carl Doersch, Bernardo Avila~Pires, Zhaohan Guo, Mohammad Gheshlaghi~Azar, et~al.
\newblock Bootstrap your own latent-a new approach to self-supervised learning.
\newblock \emph{Advances in neural information processing systems}, 33:\penalty0 21271--21284, 2020.

\bibitem[Grinsztajn et~al.(2022)Grinsztajn, Oyallon, and Varoquaux]{grinsztajn2022tree}
L{\'e}o Grinsztajn, Edouard Oyallon, and Ga{\"e}l Varoquaux.
\newblock Why do tree-based models still outperform deep learning on typical tabular data?
\newblock \emph{Advances in Neural Information Processing Systems}, 35:\penalty0 507--520, 2022.

\bibitem[Guo et~al.(2017)Guo, Tang, Ye, Li, and He]{guo2017deepfm}
Huifeng Guo, Ruiming Tang, Yunming Ye, Zhenguo Li, and Xiuqiang He.
\newblock Deepfm: a factorization-machine based neural network for ctr prediction.
\newblock \emph{arXiv preprint arXiv:1703.04247}, 2017.

\bibitem[Hager et~al.(2023)Hager, Menten, and Rueckert]{Hager_2023_CVPR}
Paul Hager, Martin~J. Menten, and Daniel Rueckert.
\newblock Best of both worlds: Multimodal contrastive learning with tabular and imaging data.
\newblock In \emph{Proceedings of the IEEE/CVF Conference on Computer Vision and Pattern Recognition (CVPR)}, pp.\  23924--23935, June 2023.

\bibitem[Hajiramezanali et~al.(2022)Hajiramezanali, Diamant, Scalia, and Shen]{hajiramezanali2022stab}
Ehsan Hajiramezanali, Nathaniel~Lee Diamant, Gabriele Scalia, and Max~W Shen.
\newblock Stab: Self-supervised learning for tabular data.
\newblock In \emph{NeurIPS 2022 First Table Representation Workshop}, 2022.

\bibitem[Han et~al.(2020)Han, Wang, Bendersky, and Najork]{han2020learning}
Shuguang Han, Xuanhui Wang, Mike Bendersky, and Marc Najork.
\newblock Learning-to-rank with bert in tf-ranking.
\newblock \emph{arXiv preprint arXiv:2004.08476}, 2020.

\bibitem[Harwood et~al.(2017)Harwood, Kumar~BG, Carneiro, Reid, and Drummond]{harwood2017smart}
Ben Harwood, Vijay Kumar~BG, Gustavo Carneiro, Ian Reid, and Tom Drummond.
\newblock Smart mining for deep metric learning.
\newblock In \emph{Proceedings of the IEEE international conference on computer vision}, pp.\  2821--2829, 2017.

\bibitem[He et~al.(2016)He, Zhang, Ren, and Sun]{he2016deep}
Kaiming He, Xiangyu Zhang, Shaoqing Ren, and Jian Sun.
\newblock Deep residual learning for image recognition.
\newblock In \emph{Proceedings of the IEEE conference on computer vision and pattern recognition}, pp.\  770--778, 2016.

\bibitem[Hegselmann et~al.(2023)Hegselmann, Buendia, Lang, Agrawal, Jiang, and Sontag]{hegselmann2023tabllm}
Stefan Hegselmann, Alejandro Buendia, Hunter Lang, Monica Agrawal, Xiaoyi Jiang, and David Sontag.
\newblock Tabllm: Few-shot classification of tabular data with large language models.
\newblock In \emph{International Conference on Artificial Intelligence and Statistics}, pp.\  5549--5581. PMLR, 2023.

\bibitem[Hendrycks et~al.(2019)Hendrycks, Mazeika, Kadavath, and Song]{hendrycks2019using}
Dan Hendrycks, Mantas Mazeika, Saurav Kadavath, and Dawn Song.
\newblock Using self-supervised learning can improve model robustness and uncertainty.
\newblock \emph{Advances in neural information processing systems}, 32, 2019.

\bibitem[Holzm{\"u}ller et~al.(2024)Holzm{\"u}ller, Grinsztajn, and Steinwart]{holzmuller2024better}
David Holzm{\"u}ller, L{\'e}o Grinsztajn, and Ingo Steinwart.
\newblock Better by default: Strong pre-tuned mlps and boosted trees on tabular data.
\newblock \emph{arXiv preprint arXiv:2407.04491}, 2024.

\bibitem[Howard \& Ruder(2018)Howard and Ruder]{howard2018universal}
Jeremy Howard and Sebastian Ruder.
\newblock Universal language model fine-tuning for text classification.
\newblock \emph{arXiv preprint arXiv:1801.06146}, 2018.

\bibitem[Huang et~al.(2020)Huang, Khetan, Cvitkovic, and Karnin]{huang2020tabtransformer}
Xin Huang, Ashish Khetan, Milan Cvitkovic, and Zohar Karnin.
\newblock Tabtransformer: Tabular data modeling using contextual embeddings.
\newblock \emph{arXiv preprint arXiv:2012.06678}, 2020.

\bibitem[Ivanov \& Prokhorenkova(2021)Ivanov and Prokhorenkova]{ivanov2021boost}
Sergei Ivanov and Liudmila Prokhorenkova.
\newblock Boost then convolve: Gradient boosting meets graph neural networks.
\newblock \emph{arXiv preprint arXiv:2101.08543}, 2021.

\bibitem[Jeffares et~al.(2023)Jeffares, Liu, Crabb{\'e}, Imrie, and van~der Schaar]{jeffares2023tangos}
Alan Jeffares, Tennison Liu, Jonathan Crabb{\'e}, Fergus Imrie, and Mihaela van~der Schaar.
\newblock Tangos: Regularizing tabular neural networks through gradient orthogonalization and specialization.
\newblock \emph{arXiv preprint arXiv:2303.05506}, 2023.

\bibitem[Jeong \& Shin(2020)Jeong and Shin]{jeong2020consistency}
Jongheon Jeong and Jinwoo Shin.
\newblock Consistency regularization for certified robustness of smoothed classifiers.
\newblock \emph{Advances in Neural Information Processing Systems}, 33:\penalty0 10558--10570, 2020.

\bibitem[Joachims(2006)]{joachims2006training}
Thorsten Joachims.
\newblock Training linear svms in linear time.
\newblock In \emph{Proceedings of the 12th ACM SIGKDD international conference on Knowledge discovery and data mining}, pp.\  217--226, 2006.

\bibitem[Kadra et~al.(2021)Kadra, Lindauer, Hutter, and Grabocka]{kadra2021well}
Arlind Kadra, Marius Lindauer, Frank Hutter, and Josif Grabocka.
\newblock Well-tuned simple nets excel on tabular datasets.
\newblock \emph{Advances in neural information processing systems}, 34:\penalty0 23928--23941, 2021.

\bibitem[Ke et~al.(2017)Ke, Meng, Finley, Wang, Chen, Ma, Ye, and Liu]{ke2017lightgbm}
Guolin Ke, Qi~Meng, Thomas Finley, Taifeng Wang, Wei Chen, Weidong Ma, Qiwei Ye, and Tie-Yan Liu.
\newblock Lightgbm: A highly efficient gradient boosting decision tree.
\newblock \emph{Advances in neural information processing systems}, 30, 2017.

\bibitem[Ke et~al.(2018)Ke, Zhang, Xu, Bian, and Liu]{ke2018tabnn}
Guolin Ke, Jia Zhang, Zhenhui Xu, Jiang Bian, and Tie-Yan Liu.
\newblock Tabnn: A universal neural network solution for tabular data.
\newblock 2018.

\bibitem[Ke et~al.(2019)Ke, Xu, Zhang, Bian, and Liu]{ke2019deepgbm}
Guolin Ke, Zhenhui Xu, Jia Zhang, Jiang Bian, and Tie-Yan Liu.
\newblock Deepgbm: A deep learning framework distilled by gbdt for online prediction tasks.
\newblock In \emph{Proceedings of the 25th ACM SIGKDD International Conference on Knowledge Discovery \& Data Mining}, pp.\  384--394, 2019.

\bibitem[Kingma \& Ba(2014)Kingma and Ba]{kingma2014adam}
Diederik~P Kingma and Jimmy Ba.
\newblock Adam: A method for stochastic optimization.
\newblock \emph{arXiv preprint arXiv:1412.6980}, 2014.

\bibitem[Kitaev et~al.(2018)Kitaev, Cao, and Klein]{kitaev2018multilingual}
Nikita Kitaev, Steven Cao, and Dan Klein.
\newblock Multilingual constituency parsing with self-attention and pre-training.
\newblock \emph{arXiv preprint arXiv:1812.11760}, 2018.

\bibitem[Kojima et~al.(2022)Kojima, Gu, Reid, Matsuo, and Iwasawa]{kojima2022large}
Takeshi Kojima, Shixiang~Shane Gu, Machel Reid, Yutaka Matsuo, and Yusuke Iwasawa.
\newblock Large language models are zero-shot reasoners.
\newblock \emph{Advances in neural information processing systems}, 35:\penalty0 22199--22213, 2022.

\bibitem[Kumar et~al.(2022)Kumar, Raghunathan, Jones, Ma, and Liang]{kumar2022fine}
Ananya Kumar, Aditi Raghunathan, Robbie~Matthew Jones, Tengyu Ma, and Percy Liang.
\newblock Fine-tuning can distort pretrained features and underperform out-of-distribution.
\newblock In \emph{International Conference on Learning Representations}, 2022.

\bibitem[Kveton et~al.(2022)Kveton, Meshi, Zoghi, and Qin]{kveton2022value}
Branislav Kveton, Ofer Meshi, Masrour Zoghi, and Zhen Qin.
\newblock On the value of prior in online learning to rank.
\newblock In \emph{International Conference on Artificial Intelligence and Statistics}, pp.\  6880--6892. PMLR, 2022.

\bibitem[Lamkhede \& Kofler(2021)Lamkhede and Kofler]{lamkhede2021recommendations}
Sudarshan~Dnyaneshwar Lamkhede and Christoph Kofler.
\newblock Recommendations and results organization in netflix search.
\newblock In \emph{Proceedings of the 15th ACM Conference on Recommender Systems}, pp.\  577--579, 2021.

\bibitem[Lee \& Shin(2022)Lee and Shin]{lee2022renyicl}
Kyungmin Lee and Jinwoo Shin.
\newblock R{\'e}nyicl: Contrastive representation learning with skew r{\'e}nyi divergence.
\newblock \emph{Advances in Neural Information Processing Systems}, 35:\penalty0 6463--6477, 2022.

\bibitem[Levin et~al.(2022)Levin, Cherepanova, Schwarzschild, Bansal, Bruss, Goldstein, Wilson, and Goldblum]{levin2022transfer}
Roman Levin, Valeriia Cherepanova, Avi Schwarzschild, Arpit Bansal, C~Bayan Bruss, Tom Goldstein, Andrew~Gordon Wilson, and Micah Goldblum.
\newblock Transfer learning with deep tabular models.
\newblock \emph{arXiv preprint arXiv:2206.15306}, 2022.

\bibitem[Lewis et~al.(2020)Lewis, Perez, Piktus, Petroni, Karpukhin, Goyal, K{\"u}ttler, Lewis, Yih, Rockt{\"a}schel, et~al.]{lewis2020retrieval}
Patrick Lewis, Ethan Perez, Aleksandra Piktus, Fabio Petroni, Vladimir Karpukhin, Naman Goyal, Heinrich K{\"u}ttler, Mike Lewis, Wen-tau Yih, Tim Rockt{\"a}schel, et~al.
\newblock Retrieval-augmented generation for knowledge-intensive nlp tasks.
\newblock \emph{Advances in Neural Information Processing Systems}, 33:\penalty0 9459--9474, 2020.

\bibitem[Li et~al.(2023{\natexlab{a}})Li, Chen, Su, Ai, and Liu]{li2023towards}
Haitao Li, Jia Chen, Weihang Su, Qingyao Ai, and Yiqun Liu.
\newblock Towards better web search performance: pre-training, fine-tuning and learning to rank.
\newblock \emph{arXiv preprint arXiv:2303.04710}, 2023{\natexlab{a}}.

\bibitem[Li et~al.(2023{\natexlab{b}})Li, Xiong, Kong, Sun, Chen, Wang, and Yin]{mpgraf}
Yuchen Li, Haoyi Xiong, Linghe Kong, Zeyi Sun, Hongyang Chen, Shuaiqiang Wang, and Dawei Yin.
\newblock { MPGraf: a Modular and Pre-trained Graphformer for Learning to Rank at Web-scale }.
\newblock In \emph{2023 IEEE International Conference on Data Mining (ICDM)}, pp.\  339--348, Los Alamitos, CA, USA, December 2023{\natexlab{b}}. IEEE Computer Society.
\newblock \doi{10.1109/ICDM58522.2023.00043}.
\newblock URL \url{https://doi.ieeecomputersociety.org/10.1109/ICDM58522.2023.00043}.

\bibitem[Li et~al.(2023{\natexlab{c}})Li, Xiong, Kong, Wang, Wang, Chen, and Yin]{s2phere}
Yuchen Li, Haoyi Xiong, Linghe Kong, Qingzhong Wang, Shuaiqiang Wang, Guihai Chen, and Dawei Yin.
\newblock S2phere: Semi-supervised pre-training for web search over heterogeneous learning to rank data.
\newblock In \emph{Proceedings of the 29th ACM SIGKDD Conference on Knowledge Discovery and Data Mining}, KDD '23, pp.\  4437–4448, New York, NY, USA, 2023{\natexlab{c}}. Association for Computing Machinery.
\newblock ISBN 9798400701030.
\newblock \doi{10.1145/3580305.3599935}.
\newblock URL \url{https://doi.org/10.1145/3580305.3599935}.

\bibitem[Li et~al.(2023{\natexlab{d}})Li, Xiong, Wang, Kong, Liu, Li, Bian, Wang, Chen, Dou, and Yin]{pingpong}
Yuchen Li, Haoyi Xiong, Qingzhong Wang, Linghe Kong, Hao Liu, Haifang Li, Jiang Bian, Shuaiqiang Wang, Guihai Chen, Dejing Dou, and Dawei Yin.
\newblock Coltr: Semi-supervised learning to rank with co-training and over-parameterization for web search.
\newblock \emph{IEEE Transactions on Knowledge and Data Engineering}, 35\penalty0 (12):\penalty0 12542--12555, 2023{\natexlab{d}}.
\newblock \doi{10.1109/TKDE.2023.3270750}.

\bibitem[Li et~al.(2024)Li, Xiong, Kong, Bian, Wang, Chen, and Yin]{gs2p}
Yuchen Li, Haoyi Xiong, Linghe Kong, Jiang Bian, Shuaiqiang Wang, Guihai Chen, and Dawei Yin.
\newblock Gs2p: a generative pre-trained learning to rank model with over-parameterization for web-scale search.
\newblock \emph{Mach. Learn.}, 113\penalty0 (8):\penalty0 5331–5349, January 2024.
\newblock ISSN 0885-6125.
\newblock \doi{10.1007/s10994-023-06469-9}.
\newblock URL \url{https://doi.org/10.1007/s10994-023-06469-9}.

\bibitem[Lian et~al.(2018)Lian, Zhou, Zhang, Chen, Xie, and Sun]{lian2018xdeepfm}
Jianxun Lian, Xiaohuan Zhou, Fuzheng Zhang, Zhongxia Chen, Xing Xie, and Guangzhong Sun.
\newblock xdeepfm: Combining explicit and implicit feature interactions for recommender systems.
\newblock In \emph{Proceedings of the 24th ACM SIGKDD international conference on knowledge discovery \& data mining}, pp.\  1754--1763, 2018.

\bibitem[Lin et~al.(2023)Lin, Qu, Guo, Dai, Tang, Yu, and Zhang]{lin2023map}
Jianghao Lin, Yanru Qu, Wei Guo, Xinyi Dai, Ruiming Tang, Yong Yu, and Weinan Zhang.
\newblock Map: A model-agnostic pretraining framework for click-through rate prediction.
\newblock In \emph{Proceedings of the 29th ACM SIGKDD Conference on Knowledge Discovery and Data Mining}, pp.\  1384--1395, 2023.

\bibitem[Liu et~al.(2022)Liu, Yang, and Wu]{liu2022ptab}
Guang Liu, Jie Yang, and Ledell Wu.
\newblock Ptab: Using the pre-trained language model for modeling tabular data.
\newblock \emph{arXiv preprint arXiv:2209.08060}, 2022.

\bibitem[Liu et~al.(2023)Liu, Chen, Fritz, and King]{liu2023tabcontrast}
Hao Liu, Yixin Chen, Bradley Fritz, and Christopher King.
\newblock Tabcontrast: A local-global level method for tabular contrastive learning.
\newblock In \emph{NeurIPS 2023 Second Table Representation Learning Workshop}, 2023.

\bibitem[Liu et~al.(2017)Liu, Qiu, and Huang]{liu2017adversarial}
Pengfei Liu, Xipeng Qiu, and Xuanjing Huang.
\newblock Adversarial multi-task learning for text classification.
\newblock \emph{arXiv preprint arXiv:1704.05742}, 2017.

\bibitem[Liu(2009)]{liu2009learning}
Tie-Yan Liu.
\newblock Learning to rank for information retrieval.
\newblock \emph{Foundations and Trends{\textregistered} in Information Retrieval}, 3\penalty0 (3):\penalty0 225--331, 2009.

\bibitem[Liu et~al.(2020)Liu, Liu, Zhang, and Chen]{liu2020dnn2lr}
Zhaocheng Liu, Qiang Liu, Haoli Zhang, and Yuntian Chen.
\newblock Dnn2lr: Interpretation-inspired feature crossing for real-world tabular data.
\newblock \emph{arXiv preprint arXiv:2008.09775}, 2020.

\bibitem[Lucchese et~al.(2016)Lucchese, Nardini, Orlando, Perego, Silvestri, and Trani]{lucchese2016post}
Claudio Lucchese, Franco~Maria Nardini, Salvatore Orlando, Raffaele Perego, Fabrizio Silvestri, and Salvatore Trani.
\newblock Post-learning optimization of tree ensembles for efficient ranking.
\newblock In \emph{Proceedings of the 39th International ACM SIGIR conference on Research and Development in Information Retrieval}, pp.\  949--952, 2016.

\bibitem[Luo et~al.(2021)Luo, Cheng, Yu, and Yi]{luo2021sdtr}
Haoran Luo, Fan Cheng, Heng Yu, and Yuqi Yi.
\newblock Sdtr: Soft decision tree regressor for tabular data.
\newblock \emph{IEEE Access}, 9:\penalty0 55999--56011, 2021.

\bibitem[Luo et~al.(2020)Luo, Zhou, Tu, Chen, Dai, and Yang]{luo2020network}
Yuanfei Luo, Hao Zhou, Wei-Wei Tu, Yuqiang Chen, Wenyuan Dai, and Qiang Yang.
\newblock Network on network for tabular data classification in real-world applications.
\newblock In \emph{Proceedings of the 43rd International ACM SIGIR Conference on Research and Development in Information Retrieval}, pp.\  2317--2326, 2020.

\bibitem[Majmundar et~al.(2022)Majmundar, Goyal, Netrapalli, and Jain]{majmundar2022met}
Kushal Majmundar, Sachin Goyal, Praneeth Netrapalli, and Prateek Jain.
\newblock Met: Masked encoding for tabular data.
\newblock \emph{arXiv preprint arXiv:2206.08564}, 2022.

\bibitem[McClosky et~al.(2006)McClosky, Charniak, and Johnson]{mcclosky2006effective}
David McClosky, Eugene Charniak, and Mark Johnson.
\newblock Effective self-training for parsing.
\newblock In \emph{Proceedings of the Human Language Technology Conference of the NAACL, Main Conference}, pp.\  152--159, 2006.

\bibitem[McElfresh et~al.(2023)McElfresh, Khandagale, Valverde, Ramakrishnan, Goldblum, White, et~al.]{mcelfresh2023neural}
Duncan McElfresh, Sujay Khandagale, Jonathan Valverde, Ganesh Ramakrishnan, Micah Goldblum, Colin White, et~al.
\newblock When do neural nets outperform boosted trees on tabular data?
\newblock \emph{arXiv preprint arXiv:2305.02997}, 2023.

\bibitem[Mitra et~al.(2018)Mitra, Craswell, et~al.]{mitra2018introduction}
Bhaskar Mitra, Nick Craswell, et~al.
\newblock An introduction to neural information retrieval.
\newblock \emph{Foundations and Trends{\textregistered} in Information Retrieval}, 13\penalty0 (1):\penalty0 1--126, 2018.

\bibitem[Miyato et~al.(2018)Miyato, Maeda, Koyama, and Ishii]{miyato2018virtual}
Takeru Miyato, Shin-ichi Maeda, Masanori Koyama, and Shin Ishii.
\newblock Virtual adversarial training: a regularization method for supervised and semi-supervised learning.
\newblock \emph{IEEE transactions on pattern analysis and machine intelligence}, 41\penalty0 (8):\penalty0 1979--1993, 2018.

\bibitem[Nam et~al.(2023{\natexlab{a}})Nam, Song, Park, Tack, Yun, Kim, and Shin]{nam2023semi}
Jaehyun Nam, Woomin Song, Seong~Hyeon Park, Jihoon Tack, Sukmin Yun, Jaehyung Kim, and Jinwoo Shin.
\newblock Semi-supervised tabular classification via in-context learning of large language models.
\newblock In \emph{Workshop on Efficient Systems for Foundation Models@ ICML2023}, 2023{\natexlab{a}}.

\bibitem[Nam et~al.(2023{\natexlab{b}})Nam, Tack, Lee, Lee, and Shin]{nam2023stunt}
Jaehyun Nam, Jihoon Tack, Kyungmin Lee, Hankook Lee, and Jinwoo Shin.
\newblock Stunt: Few-shot tabular learning with self-generated tasks from unlabeled tables.
\newblock \emph{arXiv preprint arXiv:2303.00918}, 2023{\natexlab{b}}.

\bibitem[Naumov et~al.(2019)Naumov, Mudigere, Shi, Huang, Sundaraman, Park, Wang, Gupta, Wu, Azzolini, et~al.]{naumov2019deep}
Maxim Naumov, Dheevatsa Mudigere, Hao-Jun~Michael Shi, Jianyu Huang, Narayanan Sundaraman, Jongsoo Park, Xiaodong Wang, Udit Gupta, Carole-Jean Wu, Alisson~G Azzolini, et~al.
\newblock Deep learning recommendation model for personalization and recommendation systems.
\newblock \emph{arXiv preprint arXiv:1906.00091}, 2019.

\bibitem[Nogueira et~al.(2019)Nogueira, Yang, Cho, and Lin]{nogueira2019multi}
Rodrigo Nogueira, Wei Yang, Kyunghyun Cho, and Jimmy Lin.
\newblock Multi-stage document ranking with bert.
\newblock \emph{arXiv preprint arXiv:1910.14424}, 2019.

\bibitem[Oh~Song et~al.(2016)Oh~Song, Xiang, Jegelka, and Savarese]{oh2016deep}
Hyun Oh~Song, Yu~Xiang, Stefanie Jegelka, and Silvio Savarese.
\newblock Deep metric learning via lifted structured feature embedding.
\newblock In \emph{Proceedings of the IEEE conference on computer vision and pattern recognition}, pp.\  4004--4012, 2016.

\bibitem[Oliver et~al.(2018)Oliver, Odena, Raffel, Cubuk, and Goodfellow]{oliver2018realistic}
Avital Oliver, Augustus Odena, Colin~A Raffel, Ekin~Dogus Cubuk, and Ian Goodfellow.
\newblock Realistic evaluation of deep semi-supervised learning algorithms.
\newblock \emph{Advances in neural information processing systems}, 31, 2018.

\bibitem[Oord et~al.(2018)Oord, Li, and Vinyals]{oord2018representation}
Aaron van~den Oord, Yazhe Li, and Oriol Vinyals.
\newblock Representation learning with contrastive predictive coding.
\newblock \emph{arXiv preprint arXiv:1807.03748}, 2018.

\bibitem[Pan et~al.(2022)Pan, Canim, Glass, Gliozzo, and Hendler]{pan2022end}
Feifei Pan, Mustafa Canim, Michael Glass, Alfio Gliozzo, and James Hendler.
\newblock End-to-end table question answering via retrieval-augmented generation.
\newblock \emph{arXiv preprint arXiv:2203.16714}, 2022.

\bibitem[Pang et~al.(2020)Pang, Xu, Ai, Lan, Cheng, and Wen]{pang2020setrank}
Liang Pang, Jun Xu, Qingyao Ai, Yanyan Lan, Xueqi Cheng, and Jirong Wen.
\newblock Setrank: Learning a permutation-invariant ranking model for information retrieval.
\newblock In \emph{Proceedings of the 43rd international ACM SIGIR conference on research and development in information retrieval}, pp.\  499--508, 2020.

\bibitem[Penha et~al.(2022)Penha, C{\^a}mara, and Hauff]{penha2022evaluating}
Gustavo Penha, Arthur C{\^a}mara, and Claudia Hauff.
\newblock Evaluating the robustness of retrieval pipelines with query variation generators.
\newblock In \emph{European conference on information retrieval}, pp.\  397--412. Springer, 2022.

\bibitem[Peters et~al.(2019)Peters, Ruder, and Smith]{peters2019tune}
Matthew~E Peters, Sebastian Ruder, and Noah~A Smith.
\newblock To tune or not to tune? adapting pretrained representations to diverse tasks.
\newblock \emph{arXiv preprint arXiv:1903.05987}, 2019.

\bibitem[Popov et~al.(2019)Popov, Morozov, and Babenko]{popov2019neural}
Sergei Popov, Stanislav Morozov, and Artem Babenko.
\newblock Neural oblivious decision ensembles for deep learning on tabular data.
\newblock \emph{arXiv preprint arXiv:1909.06312}, 2019.

\bibitem[Pseudo-Label(2013)]{pseudo2013simple}
Dong-Hyun~Lee Pseudo-Label.
\newblock The simple and efficient semi-supervised learning method for deep neural networks.
\newblock In \emph{ICML 2013 Workshop: Challenges in Representation Learning}, pp.\  1--6, 2013.

\bibitem[Qin \& Liu(2013)Qin and Liu]{mslr}
Tao Qin and Tie{-}Yan Liu.
\newblock Introducing {LETOR} 4.0 datasets.
\newblock \emph{CoRR}, abs/1306.2597, 2013.
\newblock URL \url{http://arxiv.org/abs/1306.2597}.

\bibitem[Qin et~al.(2021)Qin, Yan, Zhuang, Tay, Pasumarthi, Wang, Bendersky, and Najork]{qin2021neural}
Zhen Qin, Le~Yan, Honglei Zhuang, Yi~Tay, Rama~Kumar Pasumarthi, Xuanhui Wang, Mike Bendersky, and Marc Najork.
\newblock Are neural rankers still outperformed by gradient boosted decision trees?
\newblock 2021.

\bibitem[Qu et~al.(2016)Qu, Cai, Ren, Zhang, Yu, Wen, and Wang]{qu2016product}
Yanru Qu, Han Cai, Kan Ren, Weinan Zhang, Yong Yu, Ying Wen, and Jun Wang.
\newblock Product-based neural networks for user response prediction.
\newblock In \emph{2016 IEEE 16th international conference on data mining (ICDM)}, pp.\  1149--1154. IEEE, 2016.

\bibitem[Ramazanli et~al.(2025)Ramazanli, Eghbalzadeh, Liu, Wang, Fu, Rangadurai, Park, Long, and Feng]{meta2025}
Ilqar Ramazanli, Hamid Eghbalzadeh, Xiaoyi Liu, Yang Wang, Jiaxiang Fu, Kaushik Rangadurai, Sem Park, Bo~Long, and Xue Feng.
\newblock Beyond self-consistency: Loss-balanced perturbation-based regularization improves industrial-scale ads ranking, 2025.
\newblock URL \url{https://arxiv.org/abs/2502.18478}.

\bibitem[Rendle(2010)]{rendle2010factorization}
Steffen Rendle.
\newblock Factorization machines.
\newblock In \emph{2010 IEEE International conference on data mining}, pp.\  995--1000. IEEE, 2010.

\bibitem[Rizve et~al.(2021)Rizve, Duarte, Rawat, and Shah]{rizve2021defense}
Mamshad~Nayeem Rizve, Kevin Duarte, Yogesh~S Rawat, and Mubarak Shah.
\newblock In defense of pseudo-labeling: An uncertainty-aware pseudo-label selection framework for semi-supervised learning.
\newblock \emph{arXiv preprint arXiv:2101.06329}, 2021.

\bibitem[Robinson et~al.(2020)Robinson, Chuang, Sra, and Jegelka]{robinson2020contrastive}
Joshua Robinson, Ching-Yao Chuang, Suvrit Sra, and Stefanie Jegelka.
\newblock Contrastive learning with hard negative samples.
\newblock \emph{arXiv preprint arXiv:2010.04592}, 2020.

\bibitem[Rubachev et~al.(2022)Rubachev, Alekberov, Gorishniy, and Babenko]{rubachev2022revisiting}
Ivan Rubachev, Artem Alekberov, Yury Gorishniy, and Artem Babenko.
\newblock Revisiting pretraining objectives for tabular deep learning.
\newblock \emph{arXiv preprint arXiv:2207.03208}, 2022.

\bibitem[Sajjadi et~al.(2016)Sajjadi, Javanmardi, and Tasdizen]{sajjadi2016regularization}
Mehdi Sajjadi, Mehran Javanmardi, and Tolga Tasdizen.
\newblock Regularization with stochastic transformations and perturbations for deep semi-supervised learning.
\newblock \emph{Advances in neural information processing systems}, 29, 2016.

\bibitem[Schroff et~al.(2015)Schroff, Kalenichenko, and Philbin]{schroff2015facenet}
Florian Schroff, Dmitry Kalenichenko, and James Philbin.
\newblock Facenet: A unified embedding for face recognition and clustering.
\newblock In \emph{Proceedings of the IEEE conference on computer vision and pattern recognition}, pp.\  815--823, 2015.

\bibitem[Schuh et~al.(2024)Schuh, Boldini, and Sieber]{schuh2024twinbooster}
Maximilian~G Schuh, Davide Boldini, and Stephan~A Sieber.
\newblock Twinbooster: Synergising large language models with barlow twins and gradient boosting for enhanced molecular property prediction.
\newblock \emph{arXiv preprint arXiv:2401.04478}, 2024.

\bibitem[Shavitt \& Segal(2018)Shavitt and Segal]{shavitt2018regularization}
Ira Shavitt and Eran Segal.
\newblock Regularization learning networks: deep learning for tabular datasets.
\newblock \emph{Advances in Neural Information Processing Systems}, 31, 2018.

\bibitem[Shi et~al.(2018)Shi, Gong, Ding, Tao, and Zheng]{shi2018transductive}
Weiwei Shi, Yihong Gong, Chris Ding, Zhiheng~MaXiaoyu Tao, and Nanning Zheng.
\newblock Transductive semi-supervised deep learning using min-max features.
\newblock In \emph{Proceedings of the European Conference on Computer Vision (ECCV)}, pp.\  299--315, 2018.

\bibitem[Shwartz-Ziv \& Armon(2022)Shwartz-Ziv and Armon]{shwartz2022tabular}
Ravid Shwartz-Ziv and Amitai Armon.
\newblock Tabular data: Deep learning is not all you need.
\newblock \emph{Information Fusion}, 81:\penalty0 84--90, 2022.

\bibitem[Somepalli et~al.(2021)Somepalli, Goldblum, Schwarzschild, Bruss, and Goldstein]{somepalli2021saint}
Gowthami Somepalli, Micah Goldblum, Avi Schwarzschild, C~Bayan Bruss, and Tom Goldstein.
\newblock Saint: Improved neural networks for tabular data via row attention and contrastive pre-training.
\newblock \emph{arXiv preprint arXiv:2106.01342}, 2021.

\bibitem[Song \& Roth(2015)Song and Roth]{song2015unsupervised}
Yangqiu Song and Dan Roth.
\newblock Unsupervised sparse vector densification for short text similarity.
\newblock In \emph{Proceedings of the 2015 conference of the North American chapter of the association for computational linguistics: human language technologies}, pp.\  1275--1280, 2015.

\bibitem[Suh et~al.(2019)Suh, Han, Kim, and Lee]{suh2019stochastic}
Yumin Suh, Bohyung Han, Wonsik Kim, and Kyoung~Mu Lee.
\newblock Stochastic class-based hard example mining for deep metric learning.
\newblock In \emph{Proceedings of the IEEE/CVF Conference on Computer Vision and Pattern Recognition}, pp.\  7251--7259, 2019.

\bibitem[Sui et~al.(2023)Sui, Wu, Cresswell, Wu, Stein, Huang, Zhang, Volkovs, et~al.]{sui2023self}
Yi~Sui, Tongzi Wu, Jesse Cresswell, Ga~Wu, George Stein, Xiaoshi Huang, Xiaochen Zhang, Maksims Volkovs, et~al.
\newblock Self-supervised representation learning from random data projectors.
\newblock \emph{arXiv preprint arXiv:2310.07756}, 2023.

\bibitem[Sun et~al.(2019)Sun, Yang, Zhang, Lin, Dong, Young, and Dong]{sun2019supertml}
Baohua Sun, Lin Yang, Wenhan Zhang, Michael Lin, Patrick Dong, Charles Young, and Jason Dong.
\newblock Supertml: Two-dimensional word embedding for the precognition on structured tabular data.
\newblock In \emph{Proceedings of the IEEE/CVF conference on computer vision and pattern recognition workshops}, pp.\  0--0, 2019.

\bibitem[Swezey et~al.(2021)Swezey, Grover, Charron, and Ermon]{swezey2021pirank}
Robin Swezey, Aditya Grover, Bruno Charron, and Stefano Ermon.
\newblock Pirank: Scalable learning to rank via differentiable sorting.
\newblock \emph{Advances in Neural Information Processing Systems}, 34:\penalty0 21644--21654, 2021.

\bibitem[Syed \& Mirza(2023)Syed and Mirza]{syed2023self}
Tahir Syed and Behroz Mirza.
\newblock Self-supervision for tabular data by learning to predict additive homoskedastic gaussian noise as pretext.
\newblock \emph{ACM Transactions on Knowledge Discovery from Data}, 2023.

\bibitem[Tian et~al.(2021)Tian, Chen, and Ganguli]{tian2021understanding}
Yuandong Tian, Xinlei Chen, and Surya Ganguli.
\newblock Understanding self-supervised learning dynamics without contrastive pairs.
\newblock In \emph{International Conference on Machine Learning}, pp.\  10268--10278. PMLR, 2021.

\bibitem[Touvron et~al.(2023)Touvron, Martin, Stone, Albert, Almahairi, Babaei, Bashlykov, Batra, Bhargava, Bhosale, et~al.]{touvron2023llama}
Hugo Touvron, Louis Martin, Kevin Stone, Peter Albert, Amjad Almahairi, Yasmine Babaei, Nikolay Bashlykov, Soumya Batra, Prajjwal Bhargava, Shruti Bhosale, et~al.
\newblock Llama 2: Open foundation and fine-tuned chat models.
\newblock \emph{arXiv preprint arXiv:2307.09288}, 2023.

\bibitem[Ucar et~al.(2021)Ucar, Hajiramezanali, and Edwards]{ucar2021subtab}
Talip Ucar, Ehsan Hajiramezanali, and Lindsay Edwards.
\newblock Subtab: Subsetting features of tabular data for self-supervised representation learning.
\newblock \emph{Advances in Neural Information Processing Systems}, 34:\penalty0 18853--18865, 2021.

\bibitem[Verma et~al.(2021)Verma, Luong, Kawaguchi, Pham, and Le]{verma2021towards}
Vikas Verma, Thang Luong, Kenji Kawaguchi, Hieu Pham, and Quoc Le.
\newblock Towards domain-agnostic contrastive learning.
\newblock In \emph{International Conference on Machine Learning}, pp.\  10530--10541. PMLR, 2021.

\bibitem[Voorhees(2005)]{voorhees2005trec}
Ellen~M Voorhees.
\newblock The trec robust retrieval track.
\newblock In \emph{ACM SIGIR Forum}, volume~39, pp.\  11--20. ACM New York, NY, USA, 2005.

\bibitem[Wang et~al.(2021)Wang, Shivanna, Cheng, Jain, Lin, Hong, and Chi]{wang2021dcn}
Ruoxi Wang, Rakesh Shivanna, Derek Cheng, Sagar Jain, Dong Lin, Lichan Hong, and Ed~Chi.
\newblock Dcn v2: Improved deep \& cross network and practical lessons for web-scale learning to rank systems.
\newblock In \emph{Proceedings of the web conference 2021}, pp.\  1785--1797, 2021.

\bibitem[Wang et~al.(2022{\natexlab{a}})Wang, Kim, and Ganapathi]{wang2022regclr}
Weiyao Wang, Byung-Hak Kim, and Varun Ganapathi.
\newblock Regclr: A self-supervised framework for tabular representation learning in the wild.
\newblock \emph{arXiv preprint arXiv:2211.01165}, 2022{\natexlab{a}}.

\bibitem[Wang et~al.(2022{\natexlab{b}})Wang, Fan, Tian, Kihara, and Chen]{wang2022importance}
Xiao Wang, Haoqi Fan, Yuandong Tian, Daisuke Kihara, and Xinlei Chen.
\newblock On the importance of asymmetry for siamese representation learning.
\newblock In \emph{Proceedings of the IEEE/CVF Conference on Computer Vision and Pattern Recognition}, pp.\  16570--16579, 2022{\natexlab{b}}.

\bibitem[Wang \& Sun(2022)Wang and Sun]{wang2022transtab}
Zifeng Wang and Jimeng Sun.
\newblock Transtab: Learning transferable tabular transformers across tables.
\newblock \emph{Advances in Neural Information Processing Systems}, 35:\penalty0 2902--2915, 2022.

\bibitem[Wu et~al.(2017)Wu, Manmatha, Smola, and Krahenbuhl]{wu2017sampling}
Chao-Yuan Wu, R~Manmatha, Alexander~J Smola, and Philipp Krahenbuhl.
\newblock Sampling matters in deep embedding learning.
\newblock In \emph{Proceedings of the IEEE international conference on computer vision}, pp.\  2840--2848, 2017.

\bibitem[Wu et~al.(2022{\natexlab{a}})Wu, Zhang, Guo, Chen, Fan, de~Rijke, and Cheng]{wu2022certified}
Chen Wu, Ruqing Zhang, Jiafeng Guo, Wei Chen, Yixing Fan, Maarten de~Rijke, and Xueqi Cheng.
\newblock Certified robustness to word substitution ranking attack for neural ranking models.
\newblock In \emph{Proceedings of the 31st ACM International Conference on Information \& Knowledge Management}, pp.\  2128--2137, 2022{\natexlab{a}}.

\bibitem[Wu et~al.(2022{\natexlab{b}})Wu, Zhang, Guo, Fan, and Cheng]{wu2022neural}
Chen Wu, Ruqing Zhang, Jiafeng Guo, Yixing Fan, and Xueqi Cheng.
\newblock Are neural ranking models robust?
\newblock \emph{ACM Transactions on Information Systems}, 41\penalty0 (2):\penalty0 1--36, 2022{\natexlab{b}}.

\bibitem[Wu et~al.(2023)Wu, Zhang, Guo, De~Rijke, Fan, and Cheng]{wu2023prada}
Chen Wu, Ruqing Zhang, Jiafeng Guo, Maarten De~Rijke, Yixing Fan, and Xueqi Cheng.
\newblock Prada: practical black-box adversarial attacks against neural ranking models.
\newblock \emph{ACM Transactions on Information Systems}, 41\penalty0 (4):\penalty0 1--27, 2023.

\bibitem[Xia et~al.(2008)Xia, Liu, Wang, Zhang, and Li]{xia2008listwise}
Fen Xia, Tie-Yan Liu, Jue Wang, Wensheng Zhang, and Hang Li.
\newblock Listwise approach to learning to rank: theory and algorithm.
\newblock In \emph{Proceedings of the 25th international conference on Machine learning}, pp.\  1192--1199, 2008.

\bibitem[Xie et~al.(2020)Xie, Luong, Hovy, and Le]{xie2020self}
Qizhe Xie, Minh-Thang Luong, Eduard Hovy, and Quoc~V Le.
\newblock Self-training with noisy student improves imagenet classification.
\newblock In \emph{Proceedings of the IEEE/CVF conference on computer vision and pattern recognition}, pp.\  10687--10698, 2020.

\bibitem[Xu et~al.(2024)Xu, Qiu, Bai, Zhang, Miao, Wang, Tang, Liu, and Li]{grace}
Enqiang Xu, Yiming Qiu, Junyang Bai, Ping Zhang, Dadong Miao, Songlin Wang, Guoyu Tang, Lin Liu, and MingMing Li.
\newblock Optimizing e-commerce search: Toward a generalizable and rank-consistent pre-ranking model.
\newblock In \emph{Proceedings of the 47th International ACM SIGIR Conference on Research and Development in Information Retrieval}, SIGIR '24, pp.\  2875–2879, New York, NY, USA, 2024. Association for Computing Machinery.
\newblock ISBN 9798400704314.
\newblock \doi{10.1145/3626772.3661343}.
\newblock URL \url{https://doi.org/10.1145/3626772.3661343}.

\bibitem[Yan et~al.(2024{\natexlab{a}})Yan, Chen, Wang, Chen, and Wu]{yan2024team}
Jiahuan Yan, Jintai Chen, Qianxing Wang, Danny~Z Chen, and Jian Wu.
\newblock Team up gbdts and dnns: Advancing efficient and effective tabular prediction with tree-hybrid mlps.
\newblock In \emph{Proceedings of the 30th ACM SIGKDD Conference on Knowledge Discovery and Data Mining}, pp.\  3679--3689, 2024{\natexlab{a}}.

\bibitem[Yan et~al.(2024{\natexlab{b}})Yan, Zheng, Xu, Zhu, Chen, Sun, Wu, and Chen]{yan2024making}
Jiahuan Yan, Bo~Zheng, Hongxia Xu, Yiheng Zhu, Danny~Z Chen, Jimeng Sun, Jian Wu, and Jintai Chen.
\newblock Making pre-trained language models great on tabular prediction.
\newblock \emph{arXiv preprint arXiv:2403.01841}, 2024{\natexlab{b}}.

\bibitem[Yang et~al.(2022)Yang, Sanghavi, Rahmanian, Bakus, and SVN]{yang2022toward}
Shuo Yang, Sujay Sanghavi, Holakou Rahmanian, Jan Bakus, and Vishwanathan SVN.
\newblock Toward understanding privileged features distillation in learning-to-rank.
\newblock \emph{Advances in Neural Information Processing Systems}, 35:\penalty0 26658--26670, 2022.

\bibitem[Yarowsky(1995)]{yarowsky1995unsupervised}
David Yarowsky.
\newblock Unsupervised word sense disambiguation rivaling supervised methods.
\newblock In \emph{33rd annual meeting of the association for computational linguistics}, pp.\  189--196, 1995.

\bibitem[Yates et~al.(2021)Yates, Nogueira, and Lin]{yates2021pretrained}
Andrew Yates, Rodrigo Nogueira, and Jimmy Lin.
\newblock Pretrained transformers for text ranking: Bert and beyond.
\newblock In \emph{Proceedings of the 14th ACM International Conference on web search and data mining}, pp.\  1154--1156, 2021.

\bibitem[Ye et~al.(2023)Ye, Lu, Wang, Li, Wu, Chen, and Zhao]{ye2023ct}
Chao Ye, Guoshan Lu, Haobo Wang, Liyao Li, Sai Wu, Gang Chen, and Junbo Zhao.
\newblock Ct-bert: learning better tabular representations through cross-table pre-training.
\newblock \emph{arXiv preprint arXiv:2307.04308}, 2023.

\bibitem[Yoon et~al.(2020)Yoon, Zhang, Jordon, and van~der Schaar]{yoon2020vime}
Jinsung Yoon, Yao Zhang, James Jordon, and Mihaela van~der Schaar.
\newblock Vime: Extending the success of self-and semi-supervised learning to tabular domain.
\newblock \emph{Advances in Neural Information Processing Systems}, 33:\penalty0 11033--11043, 2020.

\bibitem[Yu(2020)]{yu2020ptranking}
Hai-Tao Yu.
\newblock Pt-ranking: A benchmarking platform for neural learning-to-rank, 2020.

\bibitem[Zhang et~al.(2021)Zhang, Zhang, Zhang, Pham, Yoo, and Kweon]{zhang2021does}
Chaoning Zhang, Kang Zhang, Chenshuang Zhang, Trung~X Pham, Chang~D Yoo, and In~So Kweon.
\newblock How does simsiam avoid collapse without negative samples? a unified understanding with self-supervised contrastive learning.
\newblock In \emph{International Conference on Learning Representations}, 2021.

\bibitem[Zhang et~al.(2013)Zhang, Song, Wang, and Hou]{zhang2013bias}
Peng Zhang, Dawei Song, Jun Wang, and Yuexian Hou.
\newblock Bias-variance decomposition of ir evaluation.
\newblock In \emph{Proceedings of the 36th international ACM SIGIR conference on Research and development in information retrieval}, pp.\  1021--1024, 2013.

\bibitem[Zhu et~al.(2023)Zhu, Shi, Erickson, Li, Karypis, and Shoaran]{zhu2023xtab}
Bingzhao Zhu, Xingjian Shi, Nick Erickson, Mu~Li, George Karypis, and Mahsa Shoaran.
\newblock Xtab: Cross-table pretraining for tabular transformers.
\newblock \emph{arXiv preprint arXiv:2305.06090}, 2023.

\bibitem[Zhu et~al.(2021)Zhu, Brettin, Xia, Partin, Shukla, Yoo, Evrard, Doroshow, and Stevens]{zhu2021converting}
Yitan Zhu, Thomas Brettin, Fangfang Xia, Alexander Partin, Maulik Shukla, Hyunseung Yoo, Yvonne~A Evrard, James~H Doroshow, and Rick~L Stevens.
\newblock Converting tabular data into images for deep learning with convolutional neural networks.
\newblock \emph{Scientific reports}, 11\penalty0 (1):\penalty0 11325, 2021.

\bibitem[Zhuang et~al.(2023)Zhuang, Qin, Jagerman, Hui, Ma, Lu, Ni, Wang, and Bendersky]{zhuang2023rankt5}
Honglei Zhuang, Zhen Qin, Rolf Jagerman, Kai Hui, Ji~Ma, Jing Lu, Jianmo Ni, Xuanhui Wang, and Michael Bendersky.
\newblock Rankt5: Fine-tuning t5 for text ranking with ranking losses.
\newblock In \emph{Proceedings of the 46th International ACM SIGIR Conference on Research and Development in Information Retrieval}, pp.\  2308--2313, 2023.

\end{thebibliography}
\bibliographystyle{tmlr}

\appendix
\section{Appendix}

\subsection{Detailed discussion of related work}
\label{sec:relatedwork}
\noindent
\textbf{Learning-To-Rank.}  In our paper, we focus on the traditional LTR setting where the features are all numeric (tabular data).  However, there is a line of work in LTR where raw text is also an input.  In this case, one can leverage large language models in the ranking setting \citep{zhuang2023rankt5, han2020learning,yates2021pretrained,nogueira2019multi, mitra2018introduction}.

In tabular LTR problems, the dominant models currently used are gradient boosted decision trees (GBDTs) \citep{friedman2001greedy}, which are not deep learning models.  GBDT models, which perform well on tabular data, are adapted to the LTR setting via losses that are surrogates for ranking metrics like NDCG. Surrogate losses (including LambdaRank/RankNet \citep{burges2010ranknet} and PiRank \citep{swezey2021pirank}) are needed because many important ranking metrics (like NDCG) are non-differentiable.  The combination of tree-based learners and ranking losses has become the de-facto standard in ranking problems, and deep models have yet to outperform them convincingly \citep{qin2021neural, joachims2006training, ai2019learning,bruch2019revisiting,ai2018learning, pang2020setrank}.  

\noindent
\textbf{Deep tabular models.} Given the success of neural methods in many other domains, there have been many attempts to adapt deep models to the tabular domain. 

\citet{borisov2022deep} categorize existing techniques for using deep neural networks over tabular data into four types: (1) \textbf{Encoding-based methods} such as VIME \citep{yoon2020vime}, SCARF \citep{bahri2021scarf}, IGTD \citep{zhu2021converting}, and SuperTML \citep{sun2019supertml}; (2) \textbf{Novel hybrid architectures} such as DeepFM \citep{guo2017deepfm}, xDeepFM \citep{lian2018xdeepfm}, and many others \cite{cheng2016wide,frosst2017distilling,ke2018tabnn,ke2019deepgbm,popov2019neural,luo2020network,liu2020dnn2lr,ivanov2021boost,luo2021sdtr};  (3) \textbf{Transformer-based architectures} including SAINT \citep{somepalli2021saint}, TabNet \citep{tabnet}, FT-Transformer \cite{gorishniy2021revisiting}, TabTransformer \citep{huang2020tabtransformer}, and ARM-Net \cite{cai2021arm}); (4) \textbf{Regularized DNNs} \citep{shavitt2018regularization, kadra2021well}. For instance, 
TANGOS introduced special tabular-specific regularization to try to improve deep models' performance \citep{jeffares2023tangos}.   
We also note that a more recent work, Tree-hybrid simple MLP \citep{yan2024team}, proposes to combine MLPs and GBDTs together for the tabular prediction problem. 

\noindent
\textbf{Self-supervised learning.}  Self-supervised learning (SSL) or unsupervised pretraining has improved performance in settings where there is a significant source of unlabeled data like text \citep{devlin2018bert} and images \citep{chen2020simple}.  
In SSL, deep models are first pretrained on perturbed unlabeled data using self-supervised tasks to learn useful representations for the data. Then these models are finetuned for a downstream task with labeled data.  Finetuning often takes one of two forms: (1) linear probing (a popular finetuning strategy in text and images \citep{chen2021exploring,chen2020simple,peters2019tune}),where we freeze the pretrained model and only update the linear head during supervised finetuning, and (2) full finetuning, where we update the whole model during supervised finetuning \citep{devlin2018bert}.  Sometimes a mix of the two is used \citep{kumar2022fine}.  
The core idea behind prominent SSL approaches like SimSiam and SimCLR is to carefully perturb input training samples, and train a representation that is consistent for perturbations of the same sample. This provides robustness to natural perturbations and noise  in data \citep{hendrycks2019using}.

Inspired by the success of pretraining and self-supervised learning in images and text, several works show how to apply SSL to unlabeled tabular data.  One strategy is to corrupt tabular data and train a deep model to reconstruct it \citep{yoon2020vime, majmundar2022met, ucar2021subtab, nam2023stunt, lin2023map, syed2023self, hajiramezanali2022stab}.  Another approach is to use contrastive losses, which have been highly successful in the image domain \citep{chen2020simple}. These methods are also applicable to our case because tabular data, like image data, is often composed of fixed-dimensional real vectors \citep{verma2021towards, bahri2021scarf, lee2022renyicl, Hager_2023_CVPR, liu2023tabcontrast, darabi2021contrastive}.  Concurrent work has also investigated training tabular-specific LLMs for tabular tasks \citep{yan2024making, schuh2024twinbooster}. Concurrent to us, \citet{holzmuller2024better} also proposes pretraining (with a modified architecture) to improve tabular deep learning performance. \citet{rubachev2022revisiting} evaluate a variety of different pretraining methods for tabular learning across many different datasets, finding that there is not a clear state of the art.  


\noindent
\textbf{Robustness in LTR.} There has been prior work on studying worst-case behavior (robustness) of rankers \citep{voorhees2005trec,zhang2013bias,goren2018ranking,wu2022neural, wu2022certified,penha2022evaluating}.  Some previous metrics measure a model's robustness against adversarial attack \citep{goren2018ranking,wu2022certified,wu2023prada}.  Others measure the model's per-query performance variance on a dataset \citep{voorhees2005trec, zhang2013bias, wu2022neural}.  Our outlier metric, Outlier-NDCG, is a departure from previous work because it is not directly a measure of robustness and it is possible for a model to perform \textit{better} on outlier data.

\textbf{Transfer learning.} We summarize some related work on transfer learning in tabular domain.  One direction revolves around pretraining models on common columns across many datasets \citep{zhu2023xtab, wang2022transtab, sui2023self, ye2023ct}.  Another direction leverages LLMs (large language models) to do few-shot tabular learning \citep{hegselmann2023tabllm, liu2022ptab, nam2023semi}. Pretrained model features are often useful across datasets. For example, text features are transferable between text datasets \citep{devlin2018bert}, which has contributed to the recent success of foundation models \citep{devlin2018bert, touvron2023llama, kojima2022large, brown2020language}. Unfortunately, it is still unclear how to achieve transferability in tabular LTR, where previous work has leveraged common columns (or at least column documentation) to achieve transfer learning \citep{levin2022transfer, hegselmann2023tabllm}. However most LTR datasets do not have common columns or publicly-documented column meanings \citep{mslr,chapelle2011yahoo,lucchese2016post}. 
Developing a method to transfer features in tabular LTR is an important open problem.

\textbf{Pretrained models in learning to rank.} Concurrent to the submission/review of this paper, recent works have also validated the benefits of pretraining in LTR. S2phere \citep{s2phere} proposes an elaborate recipe combining pseudo-labeling, contrastive learning, and transformer-based masked autoencoding to pretrain an LTR model, and shows that it performs well in offline and online metrics. MPGraf \citep{mpgraf} proposes a pretraining method for LTR that combines both the query-document regression formulation of LTR and the graph-based formulation of LTR into a single model. GS2P \citep{gs2p} proposes to use an ensemble of differently-trained models to generate pseudo-labels for the final model to learn. 
Like us, GRACE \citep{grace} uses contrastive learning for pre-ranking (retrieval); unlike our method, they use this process to align item embeddings with pretrained embeddings during supervised training phase.
COLTR \citep{pingpong} proposes to increase the dimensionality of the features with Fourier features and to alternative pseudo-labeling between a listwise and pointwise ranker to produce better ranking results. \citet{li2023towards} finetune pretrained language models for the document ranking task. \citet{meta2025} study self-supervised learning for recommendation systems in concurrent work.

\subsection{Experimental Details}

\subsubsection{Dataset statistics}
\label{app:datasets}
In table \cref{tab: relevance stats}, we provide the details of our three public datasets and the proprietary real-world dataset. Note that for the public dataset experiments we sub-sample the pre-training dataset in different ways (as described in \cref{sec: pretraining strategies,sec: relevance score,sec: pretraining strategies,sec: sparse}) to create the training dataset and simulate scarcity of labeled samples. However, we use the same validation dataset without and with the labels for the self-supervised pretraining and supervised learning stages, respectively. Note that in all these datasets, the higher the label value the higher the relevance of the item.

\begin{table*}[h]
  \centering
  \caption{Dataset statistics on the MSLRWEB30K, Yahoo Set1, Istella\_S, and proprietary datasets.}
  \begin{tabular}{c | c c | c c | c | c | c }
    \toprule
    & \multicolumn{5}{c|}{{\bf Number of query groups}} & {\bf Number} & {\bf List} \\
    & \multicolumn{2}{c|}{\textit{\textbf{Pretraining}}} & \multicolumn{2}{c|}{\textit{\textbf{Supervised learning}}} & & {\bf of} & {\bf of} \\
    {\bf Dataset} & {\em Pretrain} & {\em SSL Val} & {\em Train} & {\em Val} & {\em Test} & {\bf features}  & {\bf labels} \\
    \midrule
    MSLRWEB30K & 18151 & 6072 & Subset Pretrain & SSL Val & 6072 & 136 & \{0,1,2,3,4\} \\
    Yahoo Set1 & 14477 &2147 & Subset Pretrain & SSL Val & 5089 & 700 & \{0,1,2,3,4\} \\
    Istella\_S & 19200 & 7202 & Subset Pretrain & SSL Val & 6523 & 220 & \{0,1,2,3,4\}  \\
    Proprietary dataset & 35.6M & 1.62M & 3.68M & 171k & 1.08M & 63 & \{0,1\} \\
  \bottomrule
  \end{tabular}
  \label{tab: relevance stats}
\end{table*}


\subsubsection{Outliers query group details}
\label{app:outlierselection}

Here we detail how we select the outlier query groups for the Outlier-NDCG metrics (\cref{sec: outlierndcg}). For our proprietary online shopping dataset, there are known outlier features that cause low quality predictions. These outliers arise from noise introduced in various stages in the data pipeline. We have mitigation strategies in place against these feature outliers, but they are often 
imperfect. Because we already know some of the outliers features, we deem any query group with items with outlier features as an outlier. 
These outliers comprise roughly 10\% of the test set, which is still a large number of query groups (much larger than the number of query groups in the outlier datasets of the public datasets).

\begin{figure}[h]
    \begin{center}
        \includegraphics[width=0.48 \textwidth]{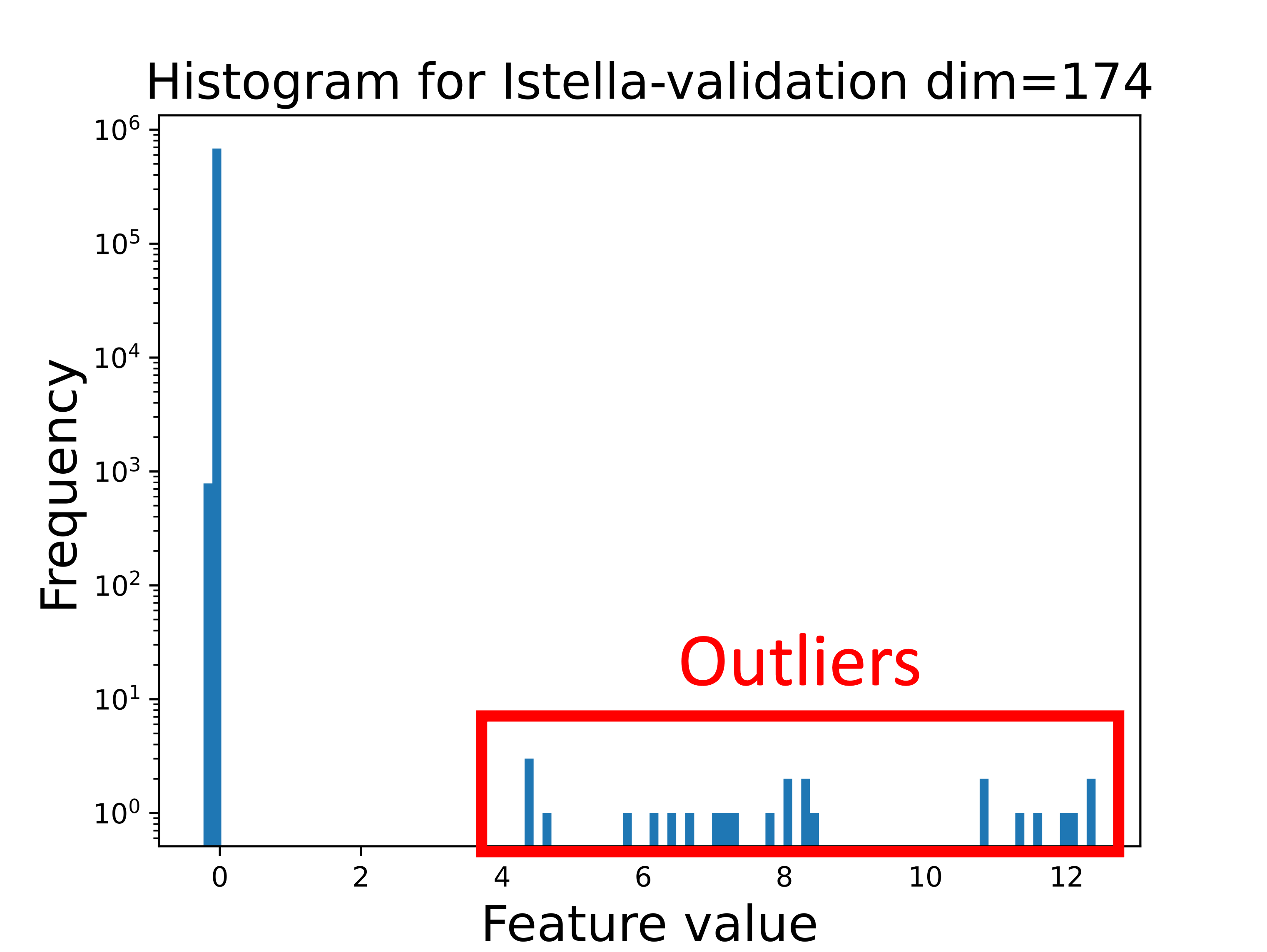}
    \end{center}
    \caption{An example of outlier detection in Istella\_S \citep{lucchese2016post} for our Outlier-NDCG metric.}
    \label{fig:outlieristella}
\end{figure}

For our public datasets, we systematically select outliers as follows: we generate a histogram with 100 bins for each feature across the \textit{validation dataset}. For example, Istella \citep{lamkhede2021recommendations} has 220 features, so we have 220 different histograms.  For each histogram, we scan from left to right on the bins until we have encountered at least $G$ empty bins in a row, and if there is less than 1\% of the validation set above this bin, then all the feature values above this bin are considered outliers.  We also repeat this process right to left.  Any test query group containing items with outlier feature values is labeled an outlier query group, and placed in the test outlier dataset. In \cref{fig:outlieristella}, we present an example histogram of a feature in Istella\_S dataset.

Because different datasets have differently-sized typical gaps, we tune $G$ for each dataset (MSLR, Yahoo, Istella) such that the resulting percentage of outlier queries is as close to 1\% of the test set as we can get.  MSLRWEB30K has $G=5$, with 0.65\% (40/6072) outlier queries, Yahoo has $G=20$, with 1.4\% (30/2147) outlier queries, and Istella has $G=32$, with 0.46\% (34/7202) outlier queries.
1\% is a hyperparameter that can be tuned according to the user's goals.

\subsubsection{Model Settings and Training Details 
}
\label{app:exp-setup-comparing-baselines}
\label{app:exp-details}

\textbf{Pretrained DL model.} 

\textit{Pretraining}:
(1) the encoder used for pretraining is the tabular ResNet \citep{gorishniy2021revisiting} with three ResNet blocks, with the final linear layer removed, (2) pretraining is done on the entire dataset with learning rate 0.0005 using Adam \citep{kingma2014adam} (3) with SimSiam/SimCLR-Rank/SimCLR-Sample, we tuned among four different augmentations for each pretraining method: randomly zeroing out features (``Zeroing'') with probabilities 0.1 or 0.7, and Gaussian noise with scale 1.0 or 2.0, (4) for SCARF we tune the corruption probability in $\{0.3, 0.6, 0.9\}$, (5) for VIME-self we tune the corruption probability in $\{0.3, 0.5, 0.7\}$, (6) for DACL+ we tune the mixup amount in $\{0.3, 0.6, 0.9\}$, (7) for SubTab we divide input features into 4 subsets with 75\% overlap, and tune the masking probability in $\{0.3, 0.6, 0.9\}$, (8) we pretrain for 300 epochs for all methods, and (9) we use a batch size of roughly 200000 items for SimCLR-Rank, SimSiam, VIME-self, and SubTab as they are less GPU memory intensive while using a batch size of roughly 2000 items for SimCLR/DACL+/SCARF which are more GPU memory intensive. 

We also evaluate SAINT \citep{somepalli2021saint}, a transformer-based pretrainining method. For the architecture, we follow \citep{somepalli2021saint}, using $L=4$ layers, dropout of 0.1, attention heads $h=4$, with a self-attention query/key/value dimension of 16 and a intersample attention query/key/value dimension of 64. For augmentation, we follow \citep{somepalli2021saint} and use a CutMix mask parameter of 0.3 and MixUp parameter of 0.2. Again following \citep{somepalli2021saint} we weight the denoising loss 10 times more than the contrastive loss. For all other minor architectural details, we follow the code of \citep{somepalli2021saint}. We pretrain for 5 epochs with a batch size of roughly 100 items due to GPU memory contraints and also the time cost (one epoch of saint took 300x longer than an epoch of SimCLR-Rank in MSLRWEB30K, 6000x longer in Yahoo Set1,  and 1000x longer in Istella). We use a learning rate of 5e-5.

\textit{Finetuning}: (1) finetuning is done on the labeled train set by adding a three-layer MLP to the top of the pretrained model and training only this head for 100 epochs and then fully finetuning for 100 epochs using Adam with a learning rate of 5e-5, (2) we use an average batch size of roughly 1000 items (may vary based on query group size), (3) we use the LambdaRank loss \citep{burges2010ranknet}, (4) we use the validation set to perform early stopping (i.e. using the checkpoint that performed best on the validation set to evaluate on the test set).  

For SAINT we follow the finetuning process of \citep{somepalli2021saint} in using the output of the CLS embedding as the representation for an item, and tune the learning rate in \{5e-5, 1e-5, 5e-6\} as we found the performance to be sensitive to this setting.


\textbf{Implicit binary user feedback setting.} We reuse the methodology of \cref{sec: relevance score}, except we use a scoring head of one linear layer (as opposed to a three layer MLP) for pretrained models. We found that this improved stability and performance in the binary label setting.


\begin{figure}[h]
  \centering
  \begin{subfigure}[b]{0.32\textwidth}
      \centering
      \includegraphics[width=\textwidth]{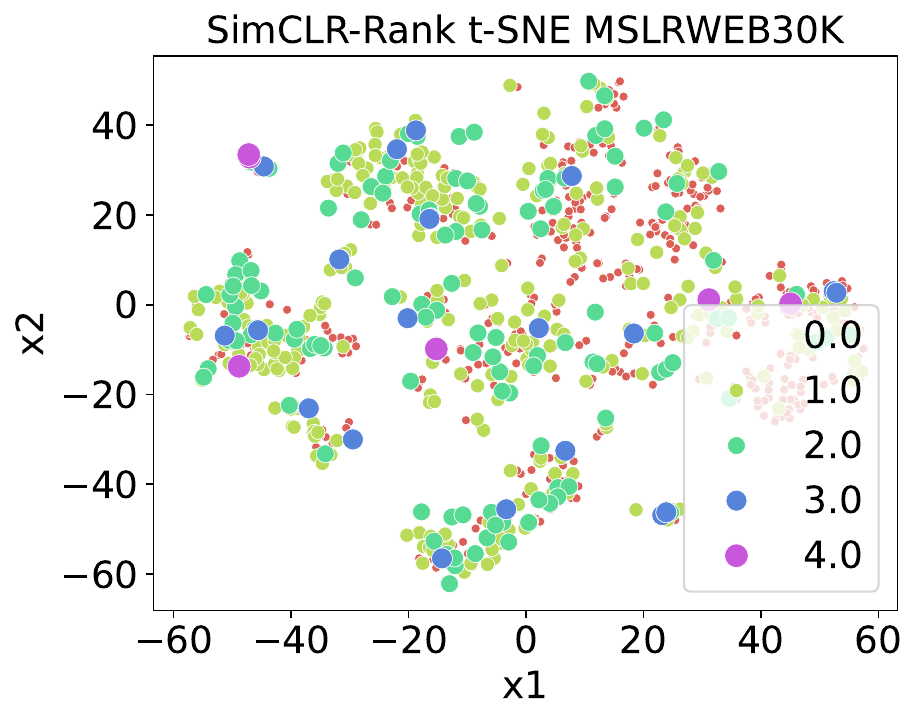}
      \label{fig:tsne mslr simclrrank}
  \end{subfigure}
  \begin{subfigure}[b]{0.32\textwidth}
      \centering
      \includegraphics[width=\textwidth]{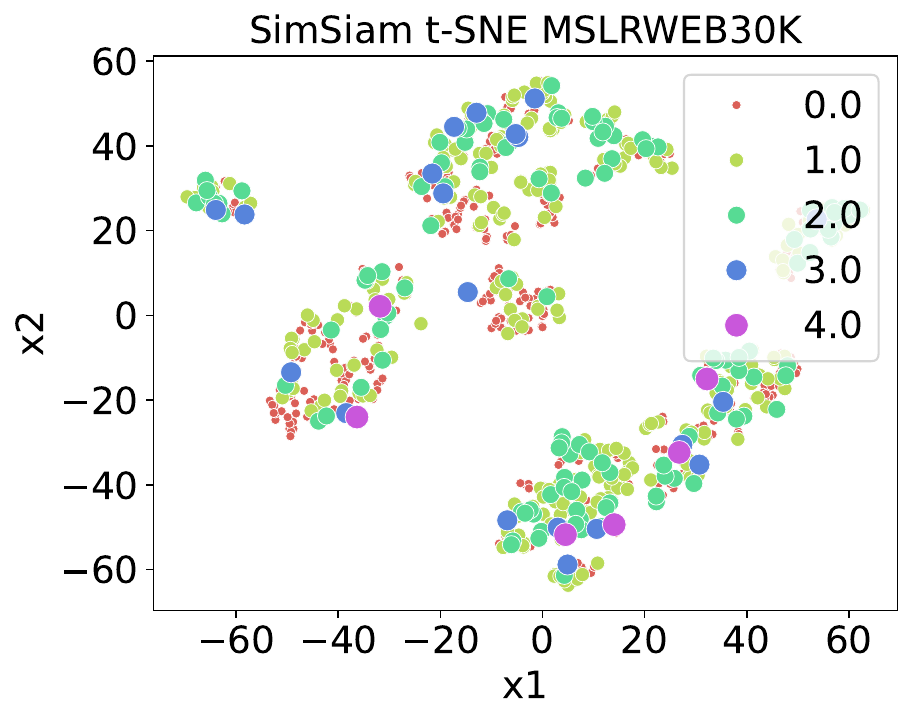}
      \label{fig:tsne mslr simsiam}
  \end{subfigure}
  \begin{subfigure}[b]{0.32\textwidth}
    \centering
    \includegraphics[width=\textwidth]{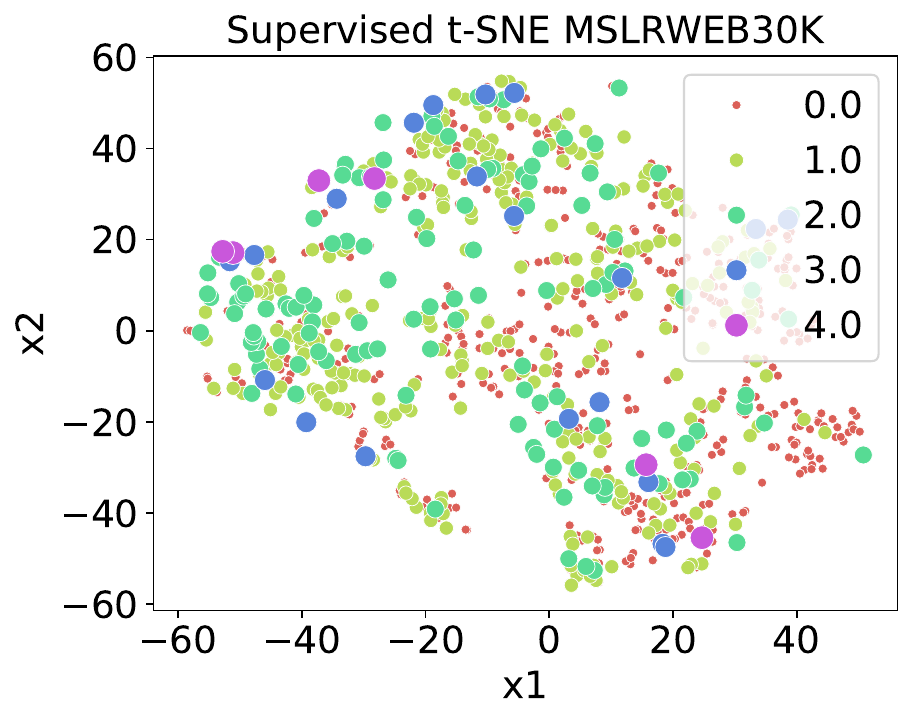}
    \label{fig:tsne mslrsupervised}
\end{subfigure}
     \caption{We plot the t-SNE plots of embeddings produced by three different encoders for the MSLRWEB30K dataset: (1) pretrained by SimCLR-Rank, (2) pretrained by SimSiam, (3) trained via supervised training on the entire training set on roughly 1000 samples from MSLR. 
     Marker size and color indicates relevance. We find (1) SimCLR-Rank/SimSiam cluster different relevances effectively but do not order them as well as the supervised encoder. (2) SimCLR-Rank produces more spread-out embeddings with less defined clusters than SimSiam.}
     \label{fig: tsne plots for mslr}
\end{figure}

\begin{figure}[h]
  \centering
  \begin{subfigure}[b]{0.32\textwidth}
      \centering
      \includegraphics[width=\textwidth]{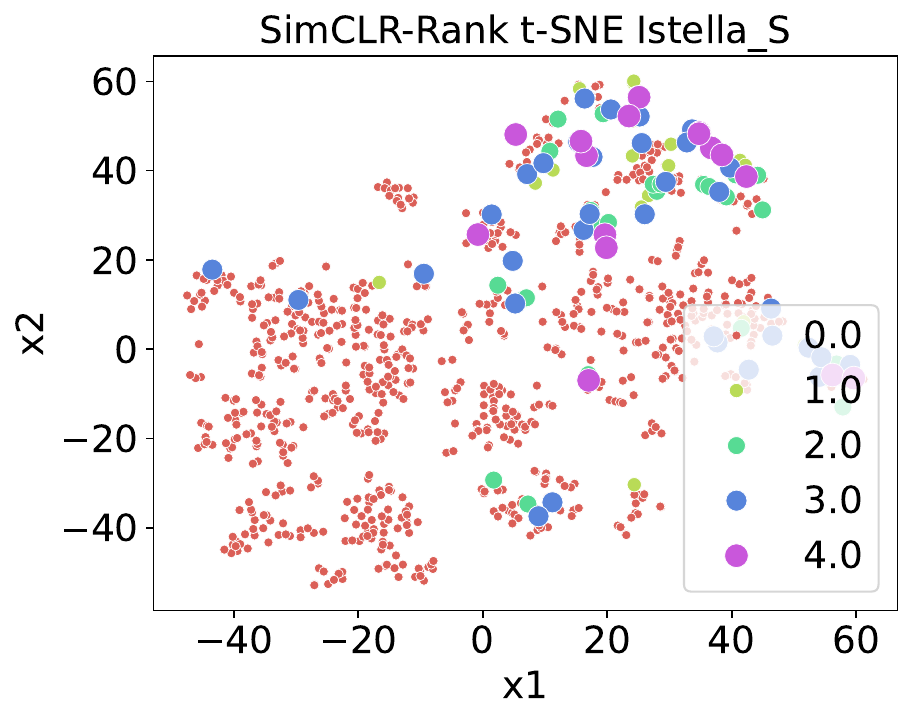}
      \label{fig:tsne istella simclrrank}
  \end{subfigure}
  \begin{subfigure}[b]{0.32\textwidth}
      \centering
      \includegraphics[width=\textwidth]{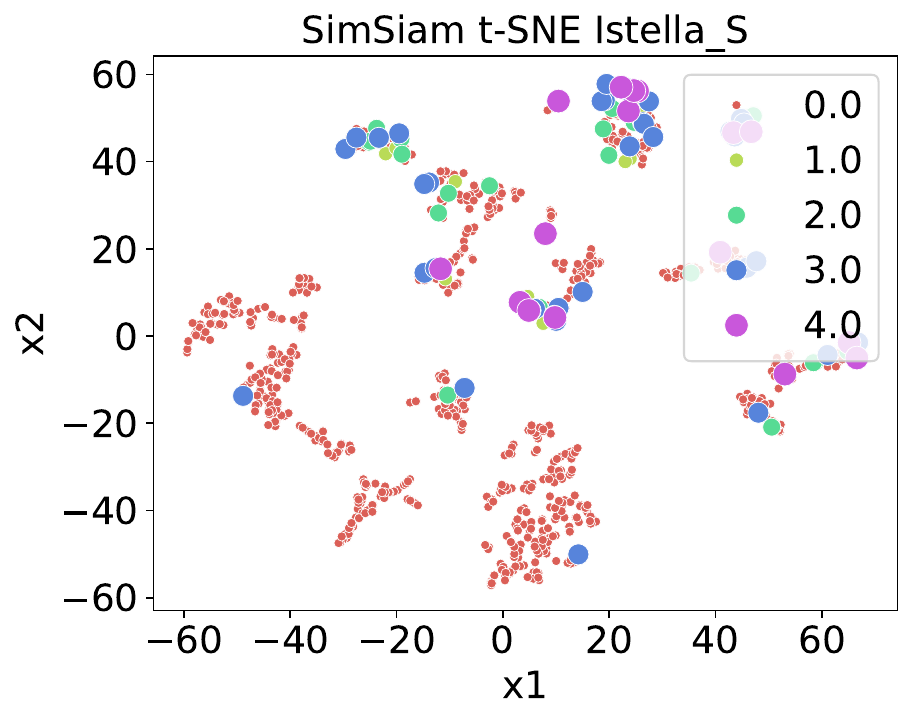}
      \label{fig:tsne istella simsiam}
  \end{subfigure}
  \begin{subfigure}[b]{0.32\textwidth}
    \centering
    \includegraphics[width=\textwidth]{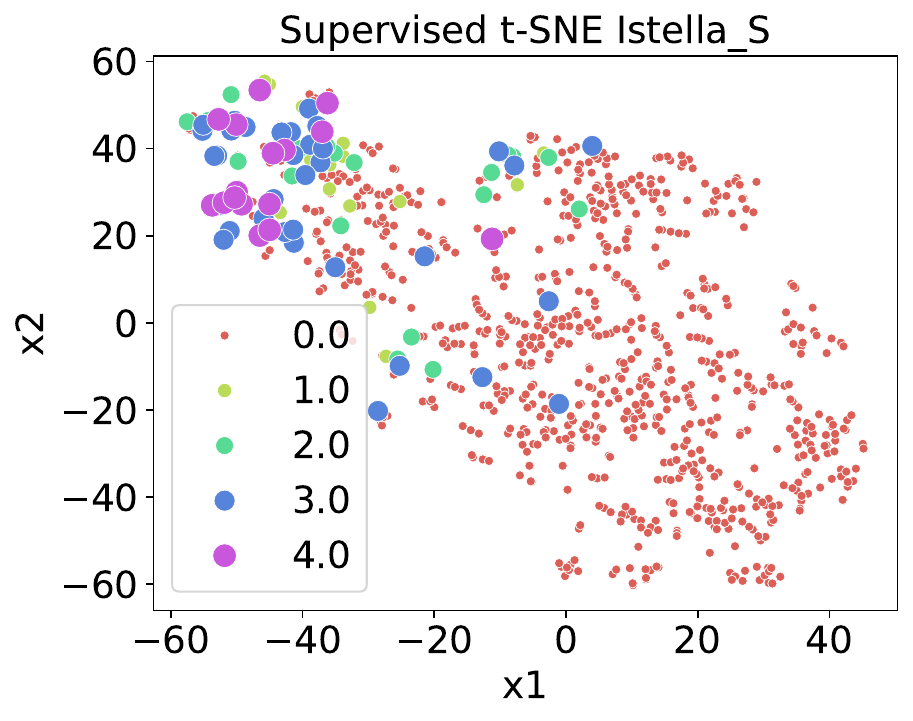}
    \label{fig:tsne istella supervised}
\end{subfigure}
     \caption{We plot the t-SNE plots of embeddings produced by three encoders: (1) pretrained by SimCLR-Rank, (2) pretrained by SimSiam, (3) trained via supervised training on the entire training set on roughly 1000 samples from Istella\_S.  Marker size and color indicates relevance. We find (1) SimCLR-Rank/SimSiam cluster different relevances effectively but do not order them as well as the supervised encoder. (2) SimCLR-Rank produces more spread-out embeddings with less defined clusters than SimSiam. 
     }
     \label{fig: tsne plots for istella}
\end{figure}
\textbf{No Pretrained DL model.}
There are a few rankers we evaluated for the no-pretrain DL baseline: (1) a 3-layer tabular ResNet from \citep{gorishniy2021revisiting} with a three-layer MLP on top of it, and was trained for 300 epochs on the labeled training set with learning rate tuned in \{0.01, 0.001, 0.0001\} using Adam, with early stopping using the validation set, (2) a DeepFM \citep{guo2017deepfm} model following the model architecture in the paper with learning rate tuned in \{0.05, 0.005, 0.0005\}, and (3) a DCNv2 \citep{wang2021dcn} model with learning rate tuned in \{0.05, 0.005, 0.0005\} and cross layers tuned in \{2,3,4\}. DeepFM and DCNv2 were designed to use categorical features, and in MSLRWEB30K/Yahoo Set1/Istella\_S do not document their categorical features (if any). We performed an analysis of the datasets and found 5, 48, 6 categorical features in MSLRWEB30K/Yahoo Set1/Istella\_S respectfully each with cardinality at most 3. We embedded these categorical features into embeddings of dimension 8, following the rule-of-thumb in \citep{wang2021dcn}.

\textbf{GBDT model.}
The GBDT ranker is the one from \texttt{lightgbm} \citep{ke2017lightgbm} and we grid search the number of leaves in \{31, 96, 200\} minimum data in leaf in \{20, 60, 200\}, and column sample in \{0.5, 0.9, 1.0\} individually for each data point (for a total of 9 difference choices), while letting the rest of the parameters be the default in \texttt{lightgbm} (our tuning strategy is similar to \citet{qin2021neural}).  We use the LambdaRank loss to train the models. 

\textbf{GBDT + pseudolabeling.}  We add pseudo-labeled GBDTs as a semi-supervised GBDT baseline. For GBDTs with pseudo-labeling (GBDT-pseudo), we propose the following algorithm as there are no references to our knowledge. First, we train the GBDT on the labeled train set, and then use this GBDT to pseudo-labels for the unlabeled train set. A GBDT ranker outputs real-valued scores. To convert real-valued scores into relevance scores, we take the resulting scores and evenly bin them into 31 buckets. The scores converted to 0 are the lowest bucket, the scores converted to 1 are the second lowest bucket, and so on. We choose 31 because that is the largest number of labels allowed for GBDT ranking in lightgbm (the reason is because the LambdaRank loss requires the calculation of \kkt{$2^\textrm{label}$} which can get very large).

\textbf{GBDT + PCA.} For GBDT+PCA, we take the entire training set's features and learn a PCA on it with \{5, 10, 15\} components, similar to \citep{duh2008learning}, to produce \{5, 10, 15\} new features (i.e. we tune among 5, 10, 15). We then train on the labeled train set, with both the original features and the new PCA features together. 

\textbf{GBDT + PCA + pseudolabeling.} We also introduce a baseline where we perform PCA on the full training dataset, and then perform pseudolabeling as above. We tune the PCA similarly to the GBDT + PCA baseline.

\textbf{SimCLR-Rank + GBDT/SimSiam + GBDT.} We also explore using pretrained neural embeddings as features in the GBDT. To do this, we used the pretrained SimCLR-Rank (or SimSiam) model to produce embeddings over the labeled data, concatenated the original features with the embeddings, and then trained a GBDT on this new dataset. We use the same tuning grid as for the original GBDT model.

\textbf{Comparing finetuning strategies.} When linear probing (LP), we freeze the pretrained model and update only a linear head on top of it for 200 epochs.  In multilayer probing (MP), we freeze the pretrained model and update a 3-layer MLP head on top of it for 200 epochs.  In full finetuning (FF), we use the finetuning strategy from \cref{sec: relevance score}.  We use the validation set for early stopping.  

All reported values for Outlier-NDCG are those achieved by the corresponding ranker with best overall NDCG.
We report all results as averages over 3 trials, and use single V100 GPUs from a shared cluster to run the experiments.

\subsubsection{Details of Binary Label generation used in \cref{sec: sparse}}
\label{app:binary-label}
Following the methodology from \citet{yang2022toward}, we generate stochastic binary labels $y$ from the relevance labels $r$ as
$
y = \mathbbm{1}\{t \cdot r + G_1 > t \cdot \tau_{\text{target}} + G_0\}
$
where $t$ is a temperature parameter, $G_1, G_0$ are standard Gumbels, and $\tau_{\text{target}}$ is a parameter controlling how sparse the binary labels are. Higher $\tau_{\text{target}}$ leads to higher label sparsity. \citet{yang2022toward} show that $y$ is 1 with probability $\sigma(t \cdot (r - \tau_{\text{target}}))$ where $\sigma$ is the sigmoid function. We set $t=4$ as in \citet{yang2022toward}. For a given $\tau_{\text{target}}$, we produce produce a new dataset for each of MSLRWEB30K, Yahoo Set1, and Istella\_S where we convert the relevance labels in the training/validation sets to binary labels and keep the true relevance labels in the test set.

\subsubsection{Large industry-scale dataset experimental details}
\label{sec:large_dataset_expt_details}

For our DL ranker, first we pass the features (numerical and categorical) through some featurizers. Then we use a simple 4 hidden layer MLP, with
and ReLU activation, to predict a single scalar score for each (query, item) pair. We also use  dropout 
and batch normalization in input and hidden layers. 

For the SimCLR pretraining stage, we remove the last linear layer of the above ranker to create the main encoder to produce the pretrained embeddings.
Projector encoder here is a 2 hidden layer MLP, with 
ReLU activation and no bias, which outputs
an input 
dimensional embedding. We also do not use any normalization or dropout in this projector. 

For SimSiam we use the same embedder and a similar projector. The only difference in the projector being batch normalization in the hidden layers and batch normalization without affine learnable parameters in the output layer. Additionaly, SimSiam also employs a predictor MLP, with a single hidden layer 
and ReLU activation, and 
with the same input and output sizes.
It uses batch normalization and no bias in the hidden layer and no input and hidden dropouts.

For pretraining, we used an augmentation which randomly zeroed out some features. We also tried randomly swapping some features between items in the same batch or query group, but they didn't perform as well as zeros augmentation. During the fine-tuning stage we discard the projectors and predictor and add back the linear layer to complete the non-pretrained ranker structure. DL rankers are trained/fine-tuned using PiRank loss \cite{swezey2021pirank}, batchsize of 1000, AdamW optimizer with learning rate of $0.00001$, weight decay of $0.01$ and cosine scheduling with warmup. All training stages and experiments were conducted on machines with 4 
GPUs, where we ran 100k iterations with ddp and picked the best checkpoint based on validation metric.

GBDT rankers were trained with LambdaRank using  the \texttt{lightgbm} package. They were tuned with Bayesian hyperparameter optimization strategy by tuning number of leaves, learning rate, and minimum data in leaf in a reasonably large range with a maximum of 150 jobs. We used default values of the library for the other hyperparameters.

\subsection{Additional results}
\subsubsection{Runtime comparisons of pre-training strategies}
\label{app:runtime-comparison}
In \cref{tab:timecomparison}, we provide the runtime comparisons between SimSiam, SimCLR, and SimCLR-Rank (new method).

\begin{table*}[h]
    \centering
    \caption{Seconds per epoch comparison between pretraining methods. Average over 3 trials.  The encoder we use for pretraining is the tabular ResNet \citep{gorishniy2021revisiting} with the final linear layer taken out.}
    \begin{tabular}{l || c | c | c | c }
      \toprule
      & & \multicolumn{3}{c}{Seconds per epoch}  \\
      Method & Complexity & MSLRWEB30K  & Yahoo Set1   & Istella   \\
      \midrule
      SimCLR & $O(B^2 L^2)$ & 102.81 $\pm$ 1.14 &14.45 $\pm$ 0.04 & 69.26 $\pm$ 0.16  \\
      SimCLR-Rank (new) & $O(B L^2)$ & 9.4700 $\pm$ 0.02  &1.970 $\pm$ 0.01 & 5.950 $\pm$ 0.04  \\
      SimSiam  & $O(BL)$ & 0.8700 $\pm$ 0.01  & 0.130 $\pm$ 0.00 & 0.730 $\pm$ 0.00  \\
    \bottomrule
    \end{tabular}
    \label{tab:timecomparison}
  \end{table*} 

\subsubsection{Additional Results on linear probing vs finetuning}
\label{app:lp-vs-ff}
Here we provide more results on the question of linear probing vs full finetuning, as discussed in \cref{sec:ablations}. In \cref{tab:simsiam linear probing,tab:simclrrank linear probing full} we compare different probing strategies for all datasets when using SimSiam and SimCLR-Rank, respectively, as the pretraining method.
\begin{table*}[t]
  \centering
  \caption{SimSiam under different finetuning strategies on NDCG/Outlier-NDCG, averaged over 3 trials (LP = Linear Probing, MP = Multilayer Probing, FF = Full Finetuning).   We find that linear probing \textit{and} MLP probing perform extremely poorly (except in Yahoo Set1, where MLP probing performs well).}
  \begin{tabular}{c || c | c | c }
    \toprule
    Method&MSLRWEB30K ($\uparrow$)  & Yahoo Set1 ($\uparrow$)  & Istella ($\uparrow$)  \\
    \midrule
    & \multicolumn{3}{c}{NDCG ($\uparrow$)}  \\
    \midrule
    LP & 0.2679 $\pm$ 0.0007 & 0.6089 $\pm$ 0.0032 & 0.3805 $\pm$ 0.0042  \\
    MP & 0.2764 $\pm$ 0.0001  & \textbf{0.6137 $\pm$ 0.0022} & 0.4484 $\pm$ 0.0020 \\
    FF & \textbf{0.3935 $\pm$ 0.0034}  & 0.6107 $\pm$ 0.0035 & \textbf{0.5618 $\pm$ 0.0049} \\
    \midrule
    & \multicolumn{3}{c}{Outlier-NDCG ($\uparrow$)}  \\
    \midrule
    LP & 0.1803 $\pm$ 0.0033 & 0.5088 $\pm$ 0.002 & 0.4407 $\pm$ 0.0388  \\
    MP & 0.1749 $\pm$ 0.0023  & 0.5157 $\pm$ 0.0080 & 0.5324 $\pm$ 0.0002 \\
    FF & \textbf{0.3149 $\pm$ 0.0119}  & \textbf{0.52 $\pm$ 0.0133} & \textbf{0.6348 $\pm$ 0.0164}  \\
  \bottomrule
  \end{tabular}
  \label{tab:simsiam linear probing}
\end{table*}

\begin{table*}[t]
    \centering
    \caption{SimCLR-Rank under different finetuning strategies on NDCG/Outlier-NDCG, averaged over 3 trials (LP = Linear Probing, MP = Multilayer Probing, FF = Full Finetuning).  Multilayer probing performs moderately worse than full finetuning, and linear probing performs much worse than all other strategies.}
    \begin{tabular}{c || c | c | c }
      \toprule
      Method&MSLRWEB30K ($\uparrow$)  & Yahoo Set1 ($\uparrow$)  & Istella ($\uparrow$)  \\
      \midrule
      & \multicolumn{3}{c}{NDCG ($\uparrow$)}  \\
      \midrule
      LP & 0.3219 $\pm$ 0.0224 & 0.5202 $\pm$ 0.0093 & 0.4029 $\pm$ 0.0073  \\
      MP & 0.3890 $\pm$ 0.0011  & 0.5942 $\pm$ 0.0016 & 0.5813 $\pm$ 0.0006 \\
      FF & \textbf{0.3959 $\pm$ 0.0022}  & \textbf{0.6022 $\pm$ 0.0013} & \textbf{0.5839 $\pm$ 0.0013} \\
      \midrule
      & \multicolumn{3}{c}{Outlier-NDCG ($\uparrow$)}  \\
      \midrule
      LP & 0.2304 $\pm$ 0.0332 & 0.3811 $\pm$ 0.0048 & 0.4717 $\pm$ 0.0166  \\
      MP & \textbf{0.2969 $\pm$ 0.0009}  & 0.4888 $\pm$ 0.0046 & \textbf{0.6369 $\pm$ 0.0045} \\
      FF & 0.2892 $\pm$ 0.0025  & \textbf{0.5143 $\pm$ 0.0055} & 0.6352 $\pm$ 0.0140  \\
    \bottomrule
    \end{tabular}
    \label{tab:simclrrank linear probing full}
  \end{table*}


\subsubsection{Additional Results on SimCLR-Rank vs SimSiam}
\label{app:simclr vs simsiam}
In this section we give additional t-SNE plots of the embeddings learned by SimCLR-Rank and SimSiam encoders, and the pre-final layer of the fully supervised ranker (trained with all labels) on MSLRWEB30K dataset (\cref{fig: tsne plots for mslr}), 
and Istella\_S dataset (\cref{fig: tsne plots for istella}). For the Yahoo Set1 dataset, SimSiam performs the best, but for the above two datasets SimCLR-Rank is the best.

\subsubsection{Additional Results: Simulated crowdsourced label setting}
\label{app:public-relevance}
\begin{figure}[h]
    \centering
    \begin{subfigure}[b]{0.32\textwidth}
        \centering
        \includegraphics[width=\textwidth]{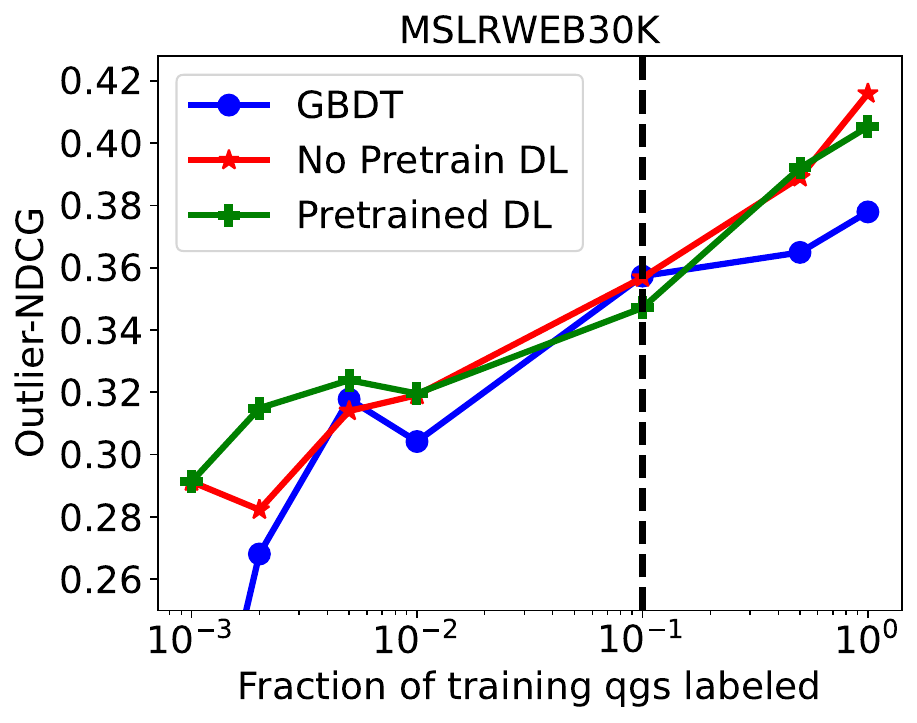}
        \label{fig:fraction outlier mslr}
    \end{subfigure}
    \begin{subfigure}[b]{0.32\textwidth}
        \centering
        \includegraphics[width=\textwidth]{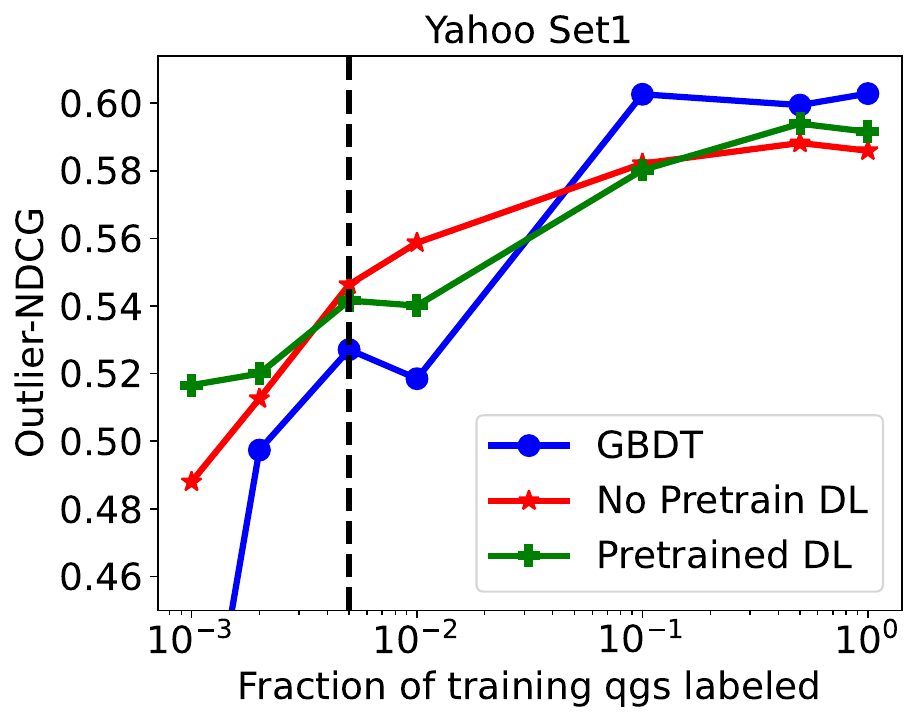}
        \label{fig:fraction outlier yahoo}
    \end{subfigure}
    \begin{subfigure}[b]{0.32\textwidth}
      \centering
      \includegraphics[width=\textwidth]{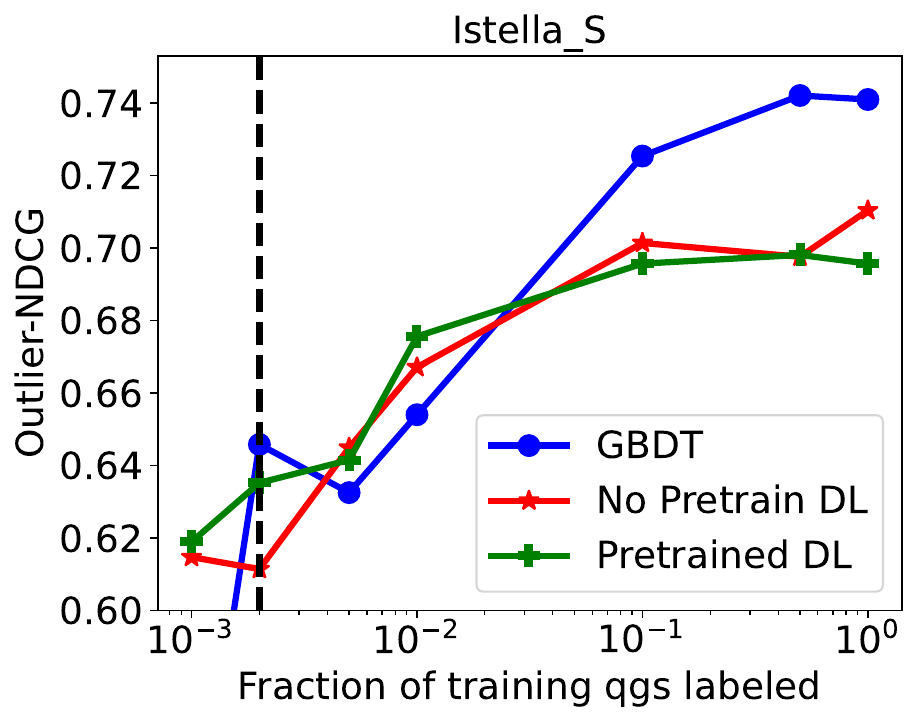}
      \label{fig:fraction outlier istella}
  \end{subfigure}
    \caption{\textbf{Simulated crowdsourcing of labels:} We compare Outlier-NDCG ($\uparrow$) pretrained rankers, non-pretrained DL rankers, and GBDT rankers as we change the percentage of training query groups that are labeled.  To the left of the black dotted line, pretrained rankers perform the best.  Points are averages over three trials.  To the left of the black dotted vertical line, pretrained rankers are (1) significantly better on outliers than GBDTs at the $p=0.05$ level using a one-sided $t$-test, and (2) on average better on outliers than all other non-pretrained methods.}
       \label{fig:outlier fractions}
  \end{figure}

Here we provide the results on label scarcity for Outlier-NDCG in \cref{fig:outlier fractions}.  Result interpretation is given in \cref{sec: relevance score}.

\subsubsection{Additional Results for Simulated Implicit Binary user feedback setting}
\label{sec: sparse appendix}

In this part of the appendix we provide additional results on binary label generation as detailed in \cref{sec: sparse}. In \cref{tab:tau 4.25 stats,tab:tau 5.1 stats} we give the dataset statistics when using $\tau_{\text{target}} = 4.25$ and $\tau_{\text{target}} = 5.1$, respectively. For these two settings, the GBDT baseline does not contain semi-supervised runs as we found that semi-supervision tends to generally degrade GBDT performance; see \cref{app:gbdt comparison}. The ``No Pretrain'' baseline uses the tabular ResNet architecture. 

\begin{table*}[h]
  \centering
  \caption{Train and validation dataset statistics on MSLRWEB30K, Yahoo Set1, and Istella\_S after binary label generation with $\tau_{\text{target}}=4.25$. The test set is left untouched. Labeled QGs are those query groups that have at least one item with $y=1$.}
  \begin{tabular}{c || c | c | c | c }
    \toprule
    Dataset & \# query groups & \# labeled QGs& \# items per QG & 
    \# positives per QG     \\
    & train \ \ \ \ \ val & train \ \ \ \ \ val & train \ \ \ \ \ val & train \ \ \ \ \ val \\
    \midrule
    MSLRWEB30K & 18151 \ \ \ 6072 & 15.9\% \ \ \ 16.2\% & 124.35 \ \ \ 122.38 & 1.4\% \ \ \ 1.4\%  \\
    Yahoo Set1 & 14477 \ \ \ 2147 & 11.1\% \ \ \ 10.8\% & 30.02 \ \ \ 30.32& 6.2\% \ \ \ 6.0\% \\
    Istella\_S & 19200 \ \ \ 7202& 49.0\%  \ \ \ 49.5\% & 106.36 \ \ \ 94.98 & 1.4\% \ \ \ 1.6\%  \\
  \bottomrule
  \end{tabular}
  \label{tab:tau 4.25 stats}
\end{table*}

\begin{table*}[h]
  \centering
  \caption{Train and validation dataset statistics on MSLRWEB30K, Yahoo Set1, and Istella\_S after binary label generation with $\tau_{\text{target}}=5.1$. The test set is left untouched. Labeled QGs are those query groups that have at least one item with $y=1$.}
  \begin{tabular}{c || c | c | c | c }
    \toprule
    Dataset & \# query groups & \# labeled QGs& \# items per QG & 
    \# positives per QG     \\
    & train \ \ \ \ \ val & train \ \ \ \ \ val & train \ \ \ \ \ val & train \ \ \ \ \ val \\
    \midrule
    MSLRWEB30K & 18151 \ \ \ 6072 & 1.1\% \ \ \ 1.3\% & 124.35 \ \ \ 122.38 & 0.8\% \ \ \ 0.7\%  \\
    Yahoo Set1 & 14477 \ \ \ 2147 & 0.6\% \ \ \ 0.6\% & 30.02 \ \ \ 30.32& 5.4\% \ \ \ 5.1\% \\
    Istella\_S & 19200 \ \ \ 7202& 3.0\%  \ \ \ 2.7\% & 106.36 \ \ \ 94.98 & 1.1\% \ \ \ 1.3\%  \\
  \bottomrule
  \end{tabular}
  \label{tab:tau 5.1 stats}
\end{table*}

\begin{table*}[t]
  \centering
  \caption{Train and validation dataset statistics on MSLRWEB30K, Yahoo Set1, and Istella\_S after binary label generation with $\tau_{\text{target}}=4.5$.  The test set is left untouched. Labeled QGs are those query groups that have at least one item with $y=1$.  }
  \begin{tabular}{c || c | c | c | c }
    \toprule
    Dataset & \# query groups & \# labeled QGs& \# items per QG & 
    \# positives per QG     \\
    & train \ \ \ \ \ val & train \ \ \ \ \ val & train \ \ \ \ \ val & train \ \ \ \ \ val \\
    \midrule
    MSLRWEB30K & 18151 \ \ \ 6072 & 8.9\% \ \ \ 8.9\% & 124.35 \ \ \ 122.38 & 1.1\% \ \ \ 1.1\%  \\
    Yahoo Set1 & 14477 \ \ \ 2147 & 5.3\% \ \ \ 5.4\% & 30.02 \ \ \ 30.32& 5.9\% \ \ \ 5.4\% \\
    Istella\_S & 19200 \ \ \ 7202& 25.2\%  \ \ \ 25.2\% & 106.36 \ \ \ 94.98 & 1.3\% \ \ \ 1.3\%  \\
  \bottomrule
  \end{tabular}
  \label{tab:tau 4.5 stats}
\end{table*}

In \cref{tab:tau 4.25,tab:tau 5.1} we give the comparison between pretrained and non-pretrained rankers following the methodology in \cref{sec: sparse} and find that GBDTs perform the best when binary labels are less sparse ($\tau_{\text{target}} = 4.25$) and pretrained DL rankers perform the best when binary labels are sparser ($\tau_{\text{target}} = 5.1$).
This strongly suggests that pretrained DL models outperform GBDTs when the labels arise from binary user feedback that is moderately uncommon or rare (as in online shopping, recommendations, or search).

\begin{table*}[h]
  \centering
  \caption{We compare pretrained models to non-pretrained models in the binary label setting with $\tau_{\text{target}}=4.25$ \pcref{sec: sparse} on NDCG averaged over three trials.  We follow the methodology in \cref{sec: sparse}.}
  \begin{tabular}{c || c | c | c }
    \toprule
    Method&MSLRWEB30K ($\uparrow$)  & Yahoo Set1 ($\uparrow$)  & Istella ($\uparrow$)  \\
    \midrule
    & \multicolumn{3}{c}{NDCG ($\uparrow$)}  \\
    \midrule
    GBDT & 0.3616 $\pm$ 0.0000 & 0.6233 $\pm$ 0.0000 & 0.6251 $\pm$ 0.0000  \\
    No Pretrain DL & 0.3552 $\pm$ 0.0015  & 0.6297 $\pm$ 0.0005 & 0.6147 $\pm$ 0.0006 \\
    Pretrained DL & 0.3602 $\pm$ 0.0007  & 0.6272 $\pm$ 0.0024 & 0.6243 $\pm$ 0.0007 \\
    \midrule
    & \multicolumn{3}{c}{Outlier-NDCG ($\uparrow$)}  \\
    \midrule
    GBDT & 0.2792 $\pm$ 0.0000 & 0.5352 $\pm$ 0.0000 & 0.7393 $\pm$ 0.0000  \\
    No Pretrain DL & 0.3142 $\pm$ 0.0063  & 0.5389 $\pm$ 0.0036 & 0.6631 $\pm$ 0.0102 \\
    Pretrained DL & 0.2923 $\pm$ 0.0079  & 0.5471 $\pm$ 0.0050 & 0.6945 $\pm$ 0.0041 \\
  \bottomrule
  \end{tabular}
  \label{tab:tau 4.25}
\end{table*}

\begin{table*}[h]
  \centering
  \caption{We compare pretrained models to non-pretrained models in the binary label setting with $\tau_{\text{target}}=5.1$ \pcref{sec: sparse} on NDCG averaged over three trials.  We follow the methodology in \cref{sec: sparse}.  ${}^\clubsuit$ indicates metrics on which pretrained rankers outperform non-pretrained rankers significantly via a $p< 0.05$ t-test.}
  \begin{tabular}{c || c | c | c }
    \toprule
    Method & MSLRWEB30K  & Yahoo Set1  & Istella \\
    \midrule 
    & \multicolumn{3}{c}{NDCG ($\uparrow$)}  \\
    \midrule
    GBDT & 0.2844 $\pm$ 0.0000 & 0.5782 $\pm$ 0.0000 & 0.5638 $\pm$ 0.0000  \\
    No Pretrain DL & 0.3432 $\pm$ 0.0065  & 0.5703 $\pm$ 0.0246 & 0.5722 $\pm$ 0.0080 \\
    Pretrained DL & \textbf{0.3564 $\pm$ 0.0054}${}^\clubsuit$ & \textbf{0.6031 $\pm$ 0.0064} ${}^\clubsuit$ & \textbf{0.599 $\pm$ 0.0014} ${}^\clubsuit$ \\
    \midrule
    & \multicolumn{3}{c}{Outlier-NDCG ($\uparrow$)}  \\
    \midrule
    GBDT & 0.2416 $\pm$ 0.0000 & 0.5011 $\pm$ 0.0000 & 0.6640 $\pm$ 0.0000  \\
    No Pretrain DL & 0.2441 $\pm$ 0.0089  & 0.4593 $\pm$ 0.0317 & 0.6334 $\pm$ 0.0054 \\
    Pretrained DL & \textbf{0.2553 $\pm$ 0.0161}  & \textbf{0.54 $\pm$ 0.0055}${}^\clubsuit$  & \textbf{0.6730 $\pm$ 0.0090} ${}^\clubsuit$ \\
  \bottomrule
  \end{tabular}
  \label{tab:tau 5.1}
\end{table*}

\subsubsection{Ablation: Data Augmentation Techniques}
\label{sec: appendix augmentations}
In \cref{tab:appendix SimCLR-Rank augmentation,tab:appendix SimSiam augmentation}, we show the performance of each augmentation choice on SimCLR-Rank and SimSiam, respectively. We use the methodology described in \cref{sec: relevance score}, and have 0.2\% of training QGs labeled.

\begin{table*}[h]
  \centering
  \caption{We compare different augmentation strategies for SimCLR-Rank.  The methodology used is the one in \cref{sec: relevance score}, with 0.2\% of training QGs labeled.}
  \begin{tabular}{c || c | c | c }
    \toprule
    Augmentation & MSLRWEB30K  & Yahoo Set1  & Istella \\
    \midrule 
    & \multicolumn{3}{c}{NDCG ($\uparrow$)}  \\
    \midrule 
    Zeroing (p=0.1) & 0.3830 $\pm$ 0.0007 & \textbf{0.6022 $\pm$ 0.0013} & 0.5820 $\pm$ 0.0024  \\
    Zeroing (p=0.7) & 0.3737 $\pm$ 0.0025  & 0.5998 $\pm$ 0.0053 & 0.5646 $\pm$ 0.0048 \\
    Gaussian noise (scale=1.0) & \textbf{0.3959 $\pm$ 0.0022}  & 0.5998 $\pm$ 0.0026 & \textbf{0.5839 $\pm$ 0.0013} \\
    Gaussian noise (scale=2.0) & 0.3907 $\pm$ 0.0025  & 0.5953 $\pm$ 0.0043 & 0.5809 $\pm$ 0.0010 \\
    \midrule
    & \multicolumn{3}{c}{Outlier-NDCG ($\uparrow$)}  \\
    \midrule 
    Zeroing (p=0.1) & 0.2782 $\pm$ 0.0036 & \textbf{0.5143 $\pm$ 0.0055} & \textbf{0.6408 $\pm$ 0.0063}  \\
    Zeroing (p=0.7) & 0.2730 $\pm$ 0.0054  & 0.5062 $\pm$ 0.0070 & 0.6141 $\pm$ 0.0064 \\
    Gaussian noise (scale=1.0)& \textbf{0.2892 $\pm$ 0.0025} & 0.4963 $\pm$ 0.0024 & 0.6352 $\pm$ 0.0140 \\
    Gaussian noise (scale=2.0)& 0.2886 $\pm$ 0.0096  & 0.4875 $\pm$ 0.0054 & 0.6327 $\pm$ 0.0191 \\
  \bottomrule
  \end{tabular}
  \label{tab:appendix SimCLR-Rank augmentation}
\end{table*}

\begin{table}[h]
  \centering
  \caption{We compare different augmentation strategies for SimSiam. The methodology used is the one in \cref{sec: relevance score}, with 0.2\% of training QGs labeled.}
  \begin{tabular}{c || c | c | c }
    \toprule
    Augmentation & MSLRWEB30K  & Yahoo Set1  & Istella \\
    \midrule 
    & \multicolumn{3}{c}{NDCG ($\uparrow$)}  \\
    \midrule 
    Zeroing (p=0.1) & \textbf{0.3935 $\pm$ 0.0034} & \textbf{0.6107 $\pm$ 0.0035} & 0.5618 $\pm$ 0.0049  \\
    Zeroing (p=0.7) & 0.3911 $\pm$ 0.0003  & 0.6076 $\pm$ 0.0072 & \textbf{0.5660 $\pm$ 0.0047} \\
    Gaussian noise (scale=1.0) & 0.3782 $\pm$ 0.0093  & 0.6010 $\pm$ 0.0026 & 0.5612 $\pm$ 0.0060 \\
    Gaussian noise (scale=2.0) & 0.3860 $\pm$ 0.0011  & 0.6100 $\pm$ 0.0089 & 0.5587 $\pm$ 0.0047 \\
    \midrule
    & \multicolumn{3}{c}{Outlier-NDCG ($\uparrow$)}  \\
    \midrule 
    Zeroing (p=0.1) & \textbf{0.3149 $\pm$ 0.0119} & \textbf{0.5200 $\pm$ 0.0133} & \textbf{0.6348 $\pm$ 0.0164}  \\
    Zeroing (p=0.7) & 0.3002 $\pm$ 0.0060  & 0.5081 $\pm$ 0.0097 & 0.6201 $\pm$ 0.0104 \\
    Gaussian noise (scale=1.0) & 0.2585 $\pm$ 0.0371  & 0.4893 $\pm$ 0.0084 & 0.6145 $\pm$ 0.0160 \\
    Gaussian noise (scale=2.0) & 0.2929 $\pm$ 0.0021  & 0.5163 $\pm$ 0.0101 & 0.5905 $\pm$ 0.0146 \\
  \bottomrule
  \end{tabular}
  \label{tab:appendix SimSiam augmentation}
\end{table}

\subsubsection{Combining SimCLR-Rank and SimSiam}
\label{sec: appendix combine}
Here we describe how we constructively combined the SimCLR-Rank and SimSiam to get the competitive results in \cref{sec:ablations}. Following the setting of \cref{app:exp-setup-comparing-baselines}, we train (1) a SimCLR-Rank model with gaussian augmentation and noise of scale 1.0, and (2) a SimSiam model with zeroing augmentation and corruption probability 0.1. 

After pretraining the SimCLR-Rank and SimSiam models, we finetune each of them on the labeled data and keep the checkpoints with the best validation NDCG. Then we combine these two models by concatenating the final embedding layers together---if each model's penultimate embedding dimension is $h$ ($h = 136$ for us), then the concatenated embedding is of dimension $2 \times h$. Then we pass this concatenated model through a batchnorm (to normalize the scales of each embedding to prevent scale issues), pass it through dropout layer of probability 0.7 (if the dataset is MSLRWEB30K or Istella\_S) or 0.0 (if the dataset is Yahoo Set1), and map this vector (of dimension $2\times h$) to a score of dimension $1$ using a linear layer. We finetune only the final linear layer.

\subsection{Semi-supervised GBDT}
\label{app:gbdt comparison}

\begin{figure}[h]
    \centering
    \begin{subfigure}[b]{0.32\textwidth}
        \centering
        \includegraphics[width=\textwidth]{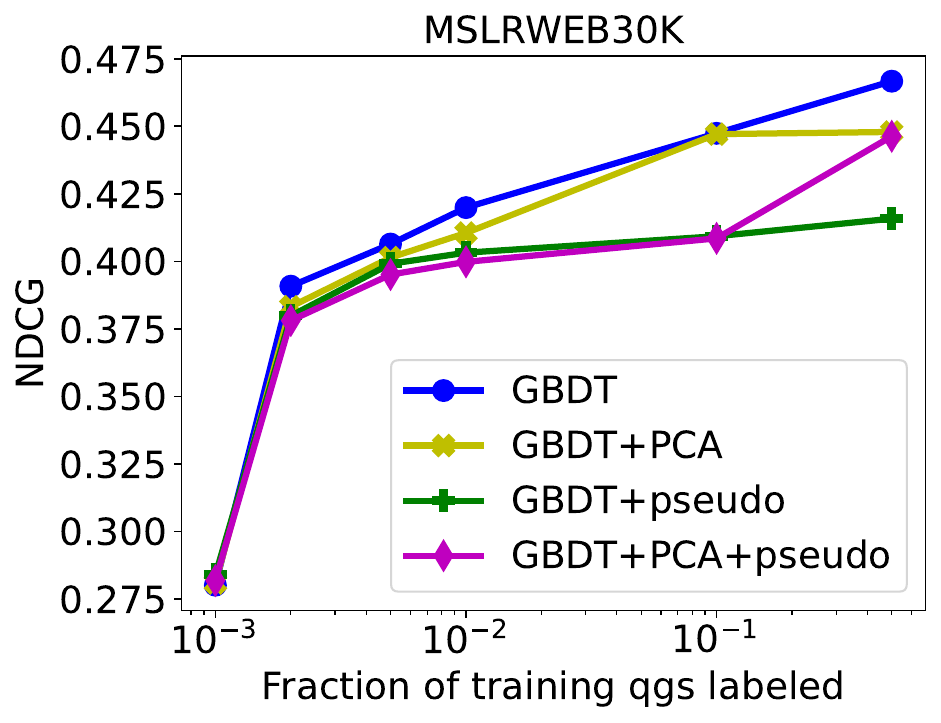}
        \label{fig:fraction gbdt mslr}
    \end{subfigure}
    \begin{subfigure}[b]{0.32\textwidth}
        \centering
        \includegraphics[width=\textwidth]{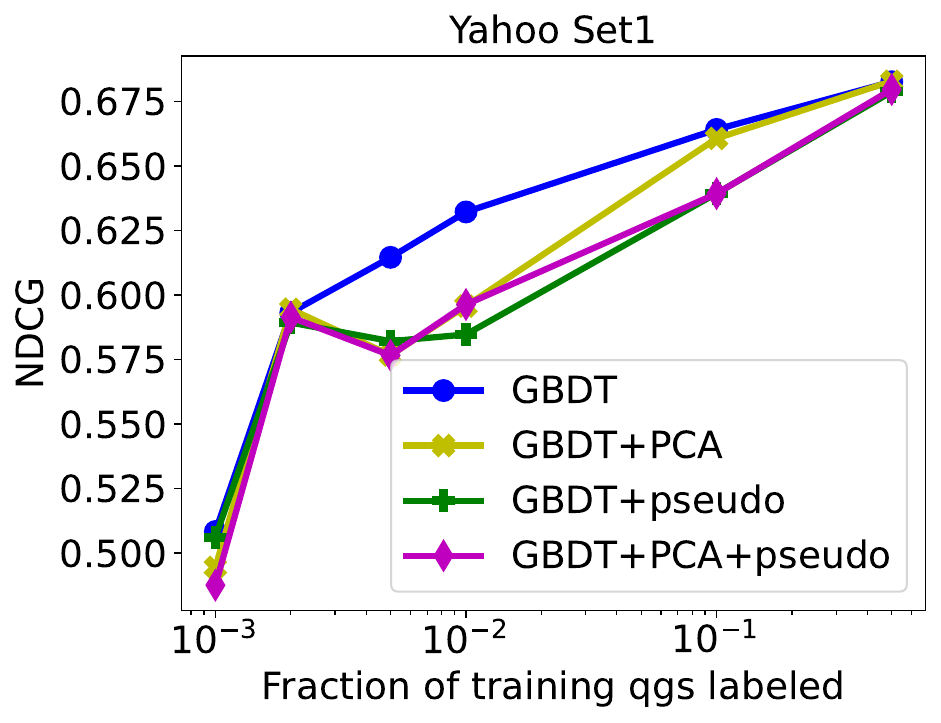}
        \label{fig:fraction gbdt yahoo}
    \end{subfigure}
    \begin{subfigure}[b]{0.32\textwidth}
      \centering
      \includegraphics[width=\textwidth]{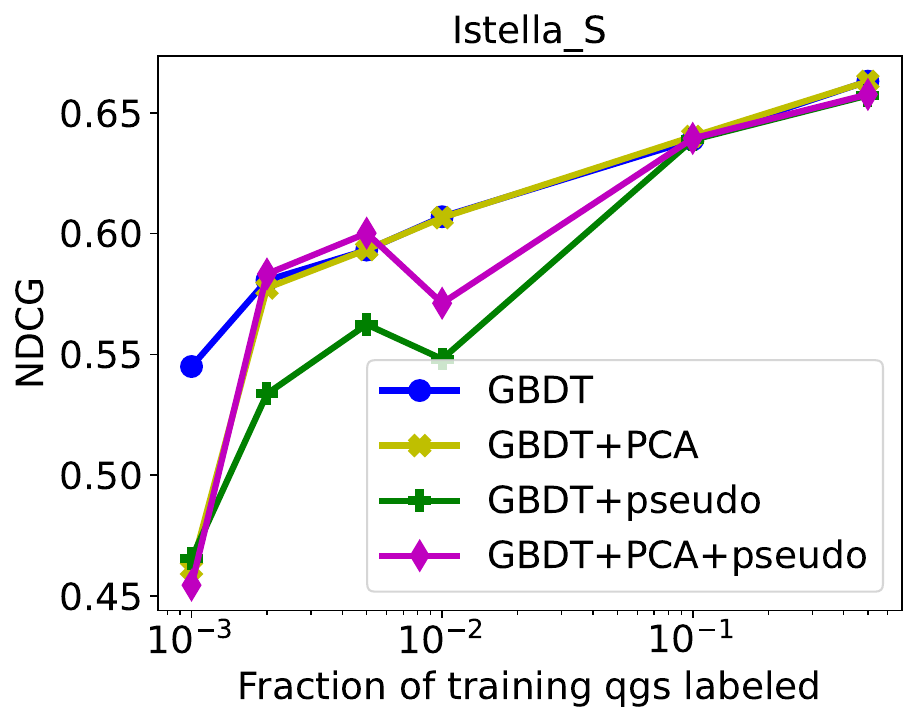}
      \label{fig:fraction gbdt istella}
  \end{subfigure}
    \caption{A comparison of Supervised GBDT, GBDT + PCA, GBDT + pseudolabeling, and GBDT + PCA + pseudolabeling. We find that semi-supervised learning may sometimes benefit GBDT rankers, though it is not very consistently beneficial.}
       \label{fig:gbdt fractions}
  \end{figure}

Here we provide a more detailed comparison between Supervised GBDTs, GBDT + pseudolabeling, and GBDT + PCA + pseudolabeling in \cref{fig:gbdt fractions} for the simulated crowdsourcing of labels setting and in \cref{tab:tau 4.5 gbdt} for the simulated implicit binary user feedback setting. We follow the experimental methodology in \cref{app:exp-details}. We find that for the most part, semi-supervised learning does not benefit GBDT rankers.

\subsection{Quantitative measures of embedding quality}
\label{app:embedding-quality}

We provide a quantitative measure of embedding quality in this section. We attempt to verify that the focus on in-query group negatives helps improve embedding quality. To do this, we compare the embeddings produced by a SimCLR-Rank model and a SimCLR-Sample model (SimCLR-Sample is the same as SimCLR-Rank except the negatives are randomly taken from the batch rather than being in-query group), both under zeroes augmentation with $p = 0.1$, and measure how spread out the embeddings are within a query group for each method. In \cref{tab:in-qg deviation}, we find that SimCLR-Rank embeddings are consistently more spread out within a query group than SimCLR-Sample, showing that in-query group focus leads to a measurable impact on the distinguish-ability of embeddings within a query group. This may help explain why SimCLR-Rank performs better than SimCLR-Sample. 


\begin{table*}[h]
  \centering
  \caption{ The ratio of the in-query group mean absolute deviation over the overall mean absolute deviation of embeddings from SimCLR-Rank and SimCLR-Sample. These embeddings were calculated over the validation set of each dataset. We find that SimCLR-Rank has a higher ratio, meaning the embeddings are more spread out within a query group.}
  \begin{tabular}{c || c | c | c }
    \toprule
    Method & MSLRWEB30K  & Yahoo Set1  & Istella \\
    \midrule
    SimCLR-Rank & 0.9953 & 0.9904 & 0.9915  \\
    SimCLR-Sample & 0.9945 & 	0.9881 & 0.9797 \\
  \bottomrule
  \end{tabular}
  \label{tab:in-qg deviation}
\end{table*}

\begin{table*}[h]
  \centering
  \caption{We compare GBDT models in the binary label setting with $\tau_{\text{target}}=4.5$ \pcref{sec: sparse} on NDCG averaged over three trials.  We follow the methodology in \cref{sec: sparse}.  }
  \begin{tabular}{c || c | c | c }
    \toprule
    Method & MSLRWEB30K  & Yahoo Set1  & Istella \\
    \midrule 
    & \multicolumn{3}{c}{NDCG ($\uparrow$)}  \\
    \midrule
    Supervised GBDT & 0.3335 $\pm$ 0.0000 & 0.6168 $\pm$ 0.0000 & 0.6024 $\pm$ 0.0000  \\
    GBDT + PCA & 0.3105 $\pm$	0.0236 & 	0.6141 $\pm$	0.0022 &	0.5508 $\pm$	0.0410 \\
    GBDT + pseudolabeling & 0.3467 $\pm$	0.0011 &	0.5868 $\pm$	0.0012 &	0.5178 $\pm$	0.0068 \\
    GBDT + PCA + pseudolabeling & 0.3241 $\pm$	0.0270 &	0.5944 $\pm$	0.0012 &	0.5315 $\pm$	0.0120 \\
  \bottomrule
  \end{tabular}
  \label{tab:tau 4.5 gbdt}
\end{table*}

\end{document}